%% file: eacl2024.tex
\pdfoutput=1

\documentclass[11pt]{article}

\usepackage{acl}

\usepackage{arabtex}
\usepackage{utf8}
\setcode{utf8}
\usepackage{subcaption}
\usepackage{times}
\usepackage{latexsym}

\usepackage[T1]{fontenc}

\usepackage[utf8]{inputenc}

\usepackage{microtype}

\usepackage{inconsolata}
\usepackage{microtype}
\usepackage{graphicx}
\usepackage{import}
\usepackage{layout}
\usepackage{tabularx, makecell}
\usepackage{booktabs}
\usepackage{mathrsfs}
\usepackage{amssymb} 
\usepackage{url}
\usepackage{graphicx}
\usepackage{xspace,paralist}
\usepackage{times,latexsym}
\usepackage{amsmath}
\usepackage{appendix}
\usepackage{comment} 
\usepackage{enumitem}
\usepackage{makecell}
\usepackage{multirow}
\usepackage{xcolor}
\usepackage{cleveref}
\usepackage{todonotes}
\usepackage{longtable,supertabular}
\usepackage[T2A,T1]{fontenc}
\usepackage[russian,english]{babel}
\usepackage{array}
\usepackage{CJKutf8}

\newcolumntype{H}{>{\setbox0=\hbox\bgroup}c<{\egroup}@{}}

\newcommand{\tabref}[2][]{Table#1~\ref{#2}\xspace}
\newcommand{\figref}[1]{Figure~\ref{#1}\xspace}
\newcommand{\secref}[1]{Section~\ref{#1}\xspace}
\newcommand{\appref}[1]{Appendix~\ref{#1}\xspace}

\newcommand{\dataset}[1]{\text{#1}\xspace}
\newcommand{\arxiv}{\dataset{arXiv}}
\newcommand{\wikipedia}{\dataset{Wikipedia}}
\newcommand{\wikihow}{\dataset{WikiHow}}
\newcommand{\peerread}{\dataset{PeerRead}}
\newcommand{\reddit}{\dataset{Reddit}}

\newcommand{\model}[1]{\text{#1}\xspace}

\newcommand{\chatgpt}{\model{ChatGPT}}
\newcommand{\gptfour}{\model{GPT-4}}

\newcommand{\davinci}{\model{davinci-003}}

\newcommand{\cohere}{\model{Cohere}}
\newcommand{\dolly}{\model{Dolly-v2}}
\newcommand{\bloomz}{\model{BLOOMz}}

\newcommand{\multirowcellii}[1]{\begin{tabular}[c]{@{}c@{}}#1\end{tabular}}

\title{M4: Multi-Generator, Multi-Domain, and Multi-Lingual\\ Black-Box Machine-Generated Text Detection}

\author{Yuxia Wang,\textsuperscript{\textdagger} Jonibek Mansurov,\textsuperscript{\textdagger}\thanks{$^*$Equal contribution.}~~Petar Ivanov,\textsuperscript{\textdagger}$^*$ Jinyan Su,\textsuperscript{\textdagger}$^*$ Artem Shelmanov,\textsuperscript{\textdagger}$^*$ \\ 
{\bf Akim Tsvigun,\textsuperscript{\textdagger} Chenxi Whitehouse,\textsuperscript{\S}\textsuperscript{\textdagger} Osama Mohammed Afzal,\textsuperscript{\textdagger}}\\
{\bf Tarek Mahmoud,\textsuperscript{\textdagger} Toru Sasaki,\textsuperscript{\textparagraph} Thomas Arnold,\textsuperscript{\textparagraph}}\\
{\bf Alham Fikri Aji,\textsuperscript{\textdagger} Nizar Habash,\textsuperscript{\textdagger}\textsuperscript{\textdaggerdbl} Iryna Gurevych,\textsuperscript{\textdagger} Preslav Nakov\textsuperscript{\textdagger}}\\
  \textsuperscript{\textdagger}Mohamed bin Zayed University of Artificial Intelligence, UAE \quad \textsuperscript{\textparagraph}TU Darmstadt, Germany \\
      \textsuperscript{\S}University of Cambridge, UK \quad\textsuperscript{\textdaggerdbl}New York University Abu Dhabi, Abu Dhabi, UAE\\
  \texttt{\{yuxia.wang, jonibek.mansurov, preslav.nakov\}@mbzuai.ac.ae} \\
  }

\begin{document}

\maketitle

\begin{abstract}

Large language models (LLMs) have demonstrated remarkable capability to generate fluent responses to a wide variety of user queries. However, this has also raised concerns about the potential misuse of such texts in journalism, education, and academia.
In this study, we strive to create automated systems that can detect machine-generated texts and pinpoint potential misuse.
We first introduce a large-scale benchmark \textbf{M4}, which is a \textbf{m}ulti-generator, \textbf{m}ulti-domain, and \textbf{m}ulti-lingual corpus for \textbf{m}achine-generated text detection. Through an extensive empirical study of this dataset, we show that it is challenging for detectors to generalize well on instances from unseen domains or LLMs. 
In such cases, detectors tend to misclassify machine-generated text as human-written. These results show that the problem is far from solved and that there is a lot of room for improvement. We believe that our dataset
will enable future research towards more robust approaches to this pressing societal problem. The dataset is available at \url{https://github.com/mbzuai-nlp/M4}.

\end{list} 
\end{abstract}

\input{section/1_intro}
\input{section/2_relatedwork}
\input{section/3_data}
\input{section/4_methods}
\input{section/5_experiment}

\input{section/6_conclusion}

\input{section/7_limitations}

\bibliography{custom,ref}
\bibliographystyle{acl_natbib}

\input{section/8_appendix}

\end{document}

%% file: section/1_intro.tex
\section{Introduction}

Large language models (LLMs) are becoming mainstream and easily accessible, ushering in an explosion of machine-generated content over various channels, such as news, social media, question-answering forums, educational, and even academic contexts. 
Recently introduced LLMs, such as  ChatGPT, GPT-4, LLaMA 2 \cite{touvron2023llama2}, and Jais \cite{sengupta2023jais}, generate remarkably fluent responses to a wide variety of user queries. 
The high quality of the generated texts makes them attractive for replacing human labor in many scenarios. 
However, this raises concerns regarding their potential misuse, e.g.,~to spread disinformation or to cause disruptions in the education system \citep{tang2023science}.

Since humans perform only slightly better than chance when classifying machine-generated vs. human-written texts~\cite{mitchell2023detectgpt}, we aim to facilitate the development of automatic detectors to mitigate the potential misuse of LLMs. In particular, we construct a diverse resource that could be used for training and testing various models for detecting machine-generated text (MGT).

Previous efforts in detecting MGT (\emph{i})~focused on only one or two particular languages, typically only on English, (\emph{ii})~used a single generator, e.g., just ChatGPT~\citep{guo2023close, shijakuchatgpt}, (\emph{iii})~leveraged fine-tuned LLMs for specific tasks, e.g.,~machine translation or text summarization~\citep{shamardina2022ruatd}, or (\emph{iv})~considered only one specific domain e.g.,~news \citep{zellers2019defending,macko-etal-2023-multitude}. 
In contrast, here we encompass multiple languages,  various LLMs, and several diverse domains, aiming to enable more general machine-generated text detection. Our dataset serves as the basis for SemEval-2024 Task~8 \citep{semeval2024task8}.

Our contributions are as follows:
\begin{itemize} 
    \item We construct \textbf{M4}: a large-scale \textbf{m}ulti-generator, \textbf{m}ulti-domain, and \textbf{m}ulti-lingual corpus for detecting \textbf{m}achine-generated texts in a black-box scenario where there is no access to a potential generator or its outputs except for plain text.
    \item We study the performance of automatic detectors from various perspectives: (a)~different detectors across different domains for a specific LLM generator, (b)~different detectors across different generators for a specific domain, (c)~interactions of domains and generators in a multilingual setting, and (d)~the performance of the detector on data generated from different time periods. From these experiments, we draw a number of  observations, which can inform future research.
    \item We release our data and code freely, and we plan to keep our repository constantly growing, adding new generators, domains, and languages over time.
\end{itemize}

The remainder of the paper is organized as follows: \secref{sec:related} discusses related work. 
\secref{sec:m4-dataset} describes the process of collecting the corpus from multiple generators (including davinci-text-003, ChatGPT, GPT4, Cohere, Dolly2, and BLOOMz), multiple domains (including Wikipedia, WikiHow, Reddit, QA, news, paper abstracts, and peer reviews), and multiple languages (Arabic, Bulgarian, Chinese, English, Indonesian, Russian, and Urdu) for machine-generated text detection. 
\secref{sec:methods} presents the seven detectors we experiment with.
\secref{sec:results} evaluates their performance across domains given a generator (ChatGPT or davinci) and across generators given a domain (arXiv or Wikipedia), as well as across different languages. Finally, \secref{sec:conclusion} concludes and points to possible directions for future work.

%% file: section/2_relatedwork.tex
\section{Related Work}
\label{sec:related}

\paragraph{White-Box vs. Black-Box Detection}
We categorize the detection strategies into black-box and white-box, contingent on the level of access to the LLM that is suspected to have generated the target text.
White-box methods focus on zero-shot detection without any additional training overhead \citep{sankar2023canai}. Some use watermarking techniques \citep{szyller2021dawn, he2022protect, kirch2023watermark, zhao2023protecting} and others rely on the expected per-token log probability of texts~\citep{krishna-etal-2022-rankgen,mitchell2023detectgpt}.
Black-box detectors only need API-level access to the LLM (i.e., when only the generated text is available) and typically extract and select features based on training text samples originating from both human and machine-generated sources. 

In this study, we focus on black-box techniques because they aim to solve the task for the more practical and general use case.
However, we note that their effectiveness heavily depends on the quality and the diversity of the training corpus. 

\paragraph{Related Corpora}
Recently, a growing body of research has concentrated on amassing responses generated by LLMs.
TuringBench~\citep{uchendu2021turingbench} comprises 200K human- and machine-generated pieces of text from 19 generative models. However, it is outdated, as the most advanced model used in this research is GPT-3.

\citet{guo2023close} collected the HC3 dataset, which consists of nearly 40K questions and their corresponding answers from human experts and ChatGPT (English and Chinese), covering a wide range of domains (computer science, finance, medicine, law, psychology, and open-domain). 

\citet{shijakuchatgpt} gathered TOEFL essays written by examined people and such generated by ChatGPT (126 essays for each).

The RuATD Shared Task 2022 involved artificial texts in Russian generated by various language models fine-tuned for specific domains or tasks such as machine translation, paraphrase generation, text summarization, and text simplification~\citep{shamardina2022ruatd}.
We pay more attention to zero-shot generations of LLMs, such as the subset of RuATD generated by ruGPT-3.

In general, previous studies have concentrated on detecting machine-generated texts in one or two languages, for a specific LLM such as ChatGPT, or within a single domain such as news \cite{zellers2019defending,macko-etal-2023-multitude}. Our work broadens this scope to include multiple languages and a variety of widely-used LLMs across different domains.

\paragraph{Black-box Detectors} are usually binary classifiers based on three types of features: statistical distributions \cite{guo2023close,shijakuchatgpt}, e.g.,~GLTR-like word rankings~\citep{gehrmann-etal-2019-gltr}, linguistic patterns (such as vocabulary, part-of-speech tags, dependency parsing, sentiment analysis, and stylistic features), and fact-verification features~\citep{tang2023science}.
Classification models involve deep neural networks, such as RoBERTa \cite{guo2023close}, or more traditional algorithms, such as logistic regression, support vector machines, Na\"{i}ve Bayes, and decision trees.

There are also widely-used off-the-shelf MGT detectors, such as the OpenAI detector,\footnote{\url{platform.openai.com/ai-text-classifier}} 
GPTZero,\footnote{\url{https://gptzero.me/}} and ZeroGPT.\footnote{\url{https://www.zerogpt.com/}}
According to the limited public information about them, these detectors are trained on collections of human-written texts and texts generated by various LLMs. For example, the training data of the OpenAI detector contains generations from 34 LLMs from various organizations, including OpenAI itself.
For our M4 dataset, we selected a diverse set of state-of-the-art black-box methods and features, including one off-the-shelf detector.

%% file: section/3_data.tex
\section{The M4 Dataset}
\label{sec:m4-dataset}

\begin{table*}[t!]
    \centering
    \resizebox{\textwidth}{!}{
    \begin{tabular}{lp{3.2cm}lr|cccccccc}
    \toprule
        \textbf{Source/} & \textbf{Data} & \textbf{Language} & \textbf{Total} & \multicolumn{8}{c}{\textbf{Parallel Data}}  \\
        \textbf{Domain} & \textbf{License} && \textbf{Human} & \textbf{Human} & \textbf{Davinci003} & \textbf{ChatGPT} & \textbf{GPT4} & \textbf{Cohere} & \textbf{Dolly-v2} & \textbf{BLOOMz} & \textbf{Total}  \\
    \midrule
        Wikipedia & CC BY-SA-3.0 & English & 6,458,670 & 3,000 & 3,000 & 2,995 & 3,000 & 2,336 & 2,702 & 3,000 & 20,033 \\
        Reddit ELI5 & Huggingface & English & 558,669 & 3,000 & 3,000 & 3,000 & 3,000 & 3,000 & 3,000 & 3,000 & 21,000 \\
        WikiHow &  CC-BY-NC-SA & English & 31,102 &  3,000 & 3,000 & 3,000 & 3,000 & 3,000 & 3,000 & 3,000 & 21,000 \\
        PeerRead & Apache license & English & 5,798 &  5,798 & 2,344 & 2,344 & 2,344 & 2,344 & 2,344 & 2,344  & 19,862 \\
        \arxiv abstract & CC0-public domain & English & 2,219,423 & 3,000 & 3,000 & 3,000 & 3,000 & 3,000 & 3,000 & 3,000 & 21,000 \\
    \midrule
        Arabic-Wikipedia & CC BY-SA-3.0 & Arabic & 1,209,042 &  3,000 & -- &  3,000 & -- & -- & --  & -- & 6,000\\
    \midrule
        True \& Fake News & MIT License & Bulgarian &   94,000 &  3,000 & 3,000 & 3,000 & -- & -- & -- & -- & 9,000 \\
    \midrule
        Baike/Web QA & MIT license & Chinese &  113,313 & 3,000 & 3,000 & 3,000 & -- & -- & -- & -- & 9,000 \\
    \midrule
        id\_newspapers\_2018 & CC BY-NC-SA-4.0 & Indonesian &   499,164 &  3,000 & -- &  3,000 & -- &  -- & -- & -- & 6,000 \\
    \midrule
        RuATD & Apache 2.0 license & Russian & 75,291 & 3,000 & 3,000 & 3,000 & -- & -- & -- & -- & 9,000 \\
    \midrule
        Urdu-news & CC BY 4.0 & Urdu & 107,881 & 3,000 & -- & 3,000  & -- & -- & -- & -- & 6,000 \\
    \midrule
        \bf Total & & & & 35,798 & 23,344 & 32,339 & 14,344 & 13,680 & 14,046 & 14,344  
        & 147,895 
        \\
    \bottomrule
    \end{tabular}
    }
    \caption{Statistics about our M4 dataset, which includes non-parallel human data and parallel human and machine-generated texts.}
    \label{tab:dataset}
\end{table*}

We gather human-written texts from a diverse range of sources across various domains and languages. For English we have Wikipedia (the March 2022 version), WikiHow~\citep{DBLP:journals/corr/abs-1810-09305}, \reddit (ELI5), \arxiv, and \peerread~\citep{kang-etal-2018-dataset}, for Chinese we have~Baike/Web QA question answering (QA), for Russian we have RuATD~\citep{shamardina2022ruatd}, for Arabic \wikipedia, and we use news for Urdu, Indonesian, and Bulgarian. Details about the data sources are provided in \appref{sec:englishsource} and \ref{sec:otherlanguagesource}.
 
For machine generation, we prompt the following multilingual LLMs: GPT-4, ChatGPT, GPT-3.5 (\textit{text-davinci-003}), Cohere, Dolly-v2 \cite{DatabricksBlog2023DollyV2}, and BLOOMz 176B~\citep{niklas2022bloomz}.
The models are asked to write articles given a title (\wikipedia), abstracts given a paper title (\arxiv), peer reviews based on the title and the abstract of a paper (\peerread), news briefs based on a title (news), also to summarize \wikipedia articles (Arabic), and to answer questions (e.g., \reddit and Baike/Web QA).\footnote{The OpenAI detector states that texts with less than 1,000 English characters are difficult, and thus we set the minimum length as 1,000 for English, and a length equal to 1,000 English characters for other languages when selecting human texts and prompting LLMs.}

\subsection{Collection}
\paragraph{Prompt Diversity}
For each generator, we carefully designed multiple (2-8) prompts in various styles, aiming to produce diverse outputs that are more aligned to divergent generations in real-world application scenarios. 
For example, on simple domains of \wikipedia and \wikihow, two prompts are applied. For \arxiv and \reddit, as well as for \chatgpt, we use five prompts and four prompts for \peerread.
We generate varying tones of responses with prompts such as \textit{answer the question} (1) ``like I am five years old''; (2) ``in an expert confident voice''; (3) ``in a formal academic and scientific writing voice''; etc. 
\tabref{tab:num-prompts} in \appref{sec:dataset} gives some statistics about the prompts used to generate the data collection, and \tabref{tab:hyperparameters} shows the hyper-parameters for the various generators.

\paragraph{Data Cleaning}
Simple artifacts in MGTs, such as multiple newlines and bullet points, could assist detectors, as their presence in the training data may discourage detectors from learning more generalized signals.

Therefore, we performed minimal cleaning of the human-written and the machine-generated texts: (\emph{i})~in a human-written \wikihow text, we removed multiple commas at the beginning of a new line (like ``,,,,,,,,,,, we believe that ...'') and repeating newlines (``$\backslash$n$\backslash$n$\backslash$n$\backslash$n$\backslash$n text begin $\backslash$n$\backslash$n$\backslash$n$\backslash$n$\backslash$n”);
(\emph{ii})~in machine-generated \wikihow texts, we removed bullet points (as there were no bullet points in human-written texts);
(\emph{iii})~in human-written \wikipedia articles, we removed references (e.g., [1], [2]), URLs, multiple newlines, as well as paragraphs whose length was less than 50 characters; and 
(\emph{iv})~in human-written \arxiv abstracts, we removed newlines stemming from PDF conversion.

\paragraph{Quality Control}
Unlike other tasks, where the data quality can be evaluated through the agreement between annotators over gold labels, we naturally obtain gold labels along with the collection of machine-generated texts. Therefore, we checked the data quality by randomly sampling 10-20 cases for each domain/generator and manually assessing the plausibility of generated texts. This can effectively circumvent incoherent, disorganized, and illogical generations that are easy to distinguish from human-written ones due to improper prompts or hyper-parameter settings of the generators (e.g.,~some generators repeat newly generated snippets to satisfy the minimum setup of new tokens). Moreover, in order to mimic human-written texts, we control the length of MGTs. 

It should be highlighted that we did not pick examples. The quality control we exercised was model-level rather than example-level. We checked for cases where a model fundamentally failed, e.g.,~by generating visibly very bad output (e.g.,~very repetitive, English instead of foreign language output, etc.). This was very high-level checking (whether to keep a certain model in M4 or not); at the individual example level, we just checked whether the output had at least 1000 characters in length. Thus, we believe any biases that we might have introduced are minimal.

\paragraph{Statistics}
The overall statistics about our M4 dataset for different tasks and languages are given in \tabref{tab:dataset}. 
We collected $\sim$ 147k human--machine parallel data in total, with 102k for English and 45k for other languages: 9k for Chinese, Russian, and Bulgarian; and 6k for Urdu, Indonesian, and Arabic respectively, in addition to over 10M non-parallel human-written texts.

\paragraph{Train, Dev, and Test Splits:} For all languages and domains, given a generator (e.g., ChatGPT), we keep 500$\times$2 (500 human-written examples and 500 machine-generated texts) for development, 500$\times$2 for testing, and the rest for training (typically, 2000$\times$2, but in some cases a bit less).

\subsection{Data Analysis}

We performed analysis of our dataset in terms of vocabulary richness at the n-gram level, as well as in terms of human performance on the task of detecting machine-generated content.

\subsubsection{N-gram Analysis}

We compared the uni-gram and the bi-gram distributions of human-written vs. machine-generated texts and found that the former had a richer vocabulary than each of the six generators; see \tabref{tab:ngramanalysis} in \appref{sec:ngram} for detail. 
Dolly-v2 had the largest number of unique uni- and bi-grams, followed by davinci, ChatGPT, and BLOOMz, and Cohere had the least.
The combination of all generators had comparable vocabulary to humans.

When comparing across domains, we observed that \wikipedia, which covers a wide range of topics, contains the highest number of unique unigrams, followed by \wikihow and \reddit. In contrast, \arxiv and \peerread, which are specific to academic papers and peer reviews, exhibited fewer unique uni-grams and bi-grams.
Within the same domain, we calculated the overlap of unique unigrams and bi-grams between human and machine-generated texts. This overlap ranges in 20--35\% for unigrams and in 10--20\% for bi-grams. These variations can provide distinctive signals for black-box machine-generated text detection approaches.

\subsubsection{Human Evaluation}

From the \reddit and the \arxiv (\chatgpt) test sets, for each domain, we sampled the first 50 (human, machine) pairs of texts and shuffled them into two groups, where two texts from the same pair would go in different groups. The annotators were then asked to focus on one group, which meant that they had to make a decision looking at each example individually, rather than having a pair of examples and deciding which one in the pair was human-written and which one was machine-generated (as some previous work did). This ensures a realistic scenario. For \reddit, we had 29 examples by humans and 21 by machines for group 1, and (21 human, 29 machine) for group 2; and (human:26, machine:24) for \arxiv group 1, (human:24, machine:26) for \arxiv group 2. 

We had a total of six human annotators, who came from different countries and were native speakers of different languages. They were all proficient in English and all had NLP background: three PhD students, two MSc students, and 1 postdoc. Annotator 3 was an English native speaker who is also proficient in Arabic. Annotators 1 and 4 were Chinese native speakers, annotators 2 and 6 ware Russian native speakers, and annotator 5 was a Bulgarian native speaker.

\begin{table}[t]
\centering
\scalebox{0.66}{
\addtolength{\tabcolsep}{-2pt}
    \begin{tabular}{l c c c c | c c c c}
        \toprule
        Domain$\rightarrow$ & \multicolumn{4}{c}{\textbf{\reddit}} & \multicolumn{4}{c}{\textbf{\arxiv}} \\
        Group$\downarrow$ & Acc. & Prec. & Recall & F1 & Acc. & Prec. & Recall & F1 \\
        \midrule
        XLM-R &  0.996 & 0.992 & 1.000 & 0.996 &  1.000 &  1.000 &  1.000 &  1.000 \\
        \midrule
        All & 0.770 & 0.770 & 0.770 & 0.770 & 0.720 & 0.739 & 0.720 & 0.714 \\
        Group1 & 0.780 & 0.775 & 0.771 & 0.773 & 0.720 & 0.744 & 0.713 & 0.708\\
        Group2 & 0.760 & 0.754 & 0.754 & 0.754 & 0.720 & 0.733 & 0.724 & 0.718 \\
        \midrule
        Annotator1 & 0.765 & 0.846 & 0.750 & 0.742 & 0.600 & 0.675 & 0.612 & 0.566 \\
        Annotator2 & 0.882 & 0.917 & 0.857 & 0.871 & 0.840 & 0.838 & 0.838 & 0.838 \\
        Annotator3 & 0.688 & 0.773 & 0.75 & 0.686 & 0.640 & 0.640 & 0.638 & 0.638  \\
        Annotator4 & 0.938 & 0.929 & 0.950 & 0.935 & 0.800 & 0.844 & 0.821 & 0.799 \\
        Annotator5 & 0.412 & 0.410 & 0.410 & 0.410 & -- & -- & -- & -- \\
        Annotator6 & 0.941 & 0.955 & 0.929 & 0.938 & -- & -- & -- & --\\
        
        \bottomrule
    \end{tabular}
    }
    \caption{Human evaluation on 100 examples from \reddit and \arxiv (human, \chatgpt). The XLM-R detector fine-tuned on in-domain data demonstrated much better results than human annotators.}
    \label{tab:human-eval}
\end{table}

Each annotator made a guess about 17 unique examples for \reddit (finished by six annotators) and 25 examples for \arxiv (finished by four).\footnote{The best and the worst raters were not invited to annotate for \arxiv, to avoid the bias of representing the average ability of human detection.} The results are shown in Table~\ref{tab:human-eval}.
Interestingly, the English native speaker did not perform as well as some other annotators. 

We can further see in Table~\ref{tab:human-eval} that annotator 4 performed much better than annotator 1, even though they were both Chinese native speakers; this may be because annotator 4 had better understanding of how LLM generations work. Moreover, annotator 6 was the best rater, and he was also the one who was very familiar with LLM generation mechanisms, achieving higher guessing accuracy than annotator 2.

Thus, the annotators' proficiency in English may affect the evaluation, but for equal language proficiency, the degree of understanding of the LLM generation styles or patterns will also impact the quality of the annotator's guess.

On average, the accuracy of the human guesses was 0.77 for \reddit and 0.72 for \arxiv. This indicates that it is not easy for humans to detect machine-generated text, especially for non-native English speakers who are not familiar with the \chatgpt generation patterns (e.g.,~annotators 1,3,5).
Besides, it is harder to classify the texts from \arxiv than from \reddit.

This is consistent with the findings in \citet{clark-etal-2021-thats}.
Without training, evaluators distinguished between GPT3-written and human-authored text at the chance level, and training by detailed instructions, annotated examples, and paired examples will improve the accuracy while the improvement across domains differs.

We hypothesize that our human annotators depended less on content signals and more on stylistic cues when identifying MGT for the \arxiv domain, which results in the accuracy disparity between the two domains.
Overall, it is challenging for general readers to understand and to follow abstracts of academic papers, but it is much easier to read \reddit answers.

We further compared the human performance to an XLM-R detector fine-tuned on in-domain training data. The classifier achieved near-perfect accuracy across the two domains, outperforming all human annotators. These findings strongly indicate the potential for automated in-domain black-box detection.

%% file: section/4_methods.tex
\section{Detectors}
\label{sec:methods}
We evaluated seven detectors; see \tabref{tab:detector-parameters} for their hyper-parameter settings.

\paragraph{RoBERTa}
This detector is based on the pre-trained RoBERTa model \cite{liu2019roberta}, which we fine-tuned to detect machine-generated texts. 

\paragraph{ELECTRA}
We further fine-tuned ELECTRA~\citep{clark2020electra}. Its pre-training objective is more aligned with our MGT task: it was pre-trained to predict whether a token in a corrupted input was replaced by a plausible alternative sampled from a small generator network.

\begin{figure*}[t!]
	\centering
	\includegraphics[scale=0.28]{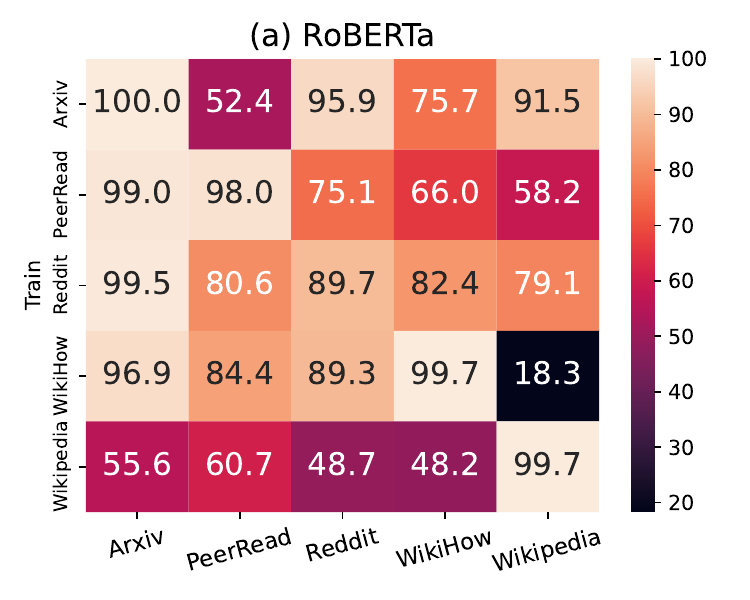} 
	\includegraphics[scale=0.28]{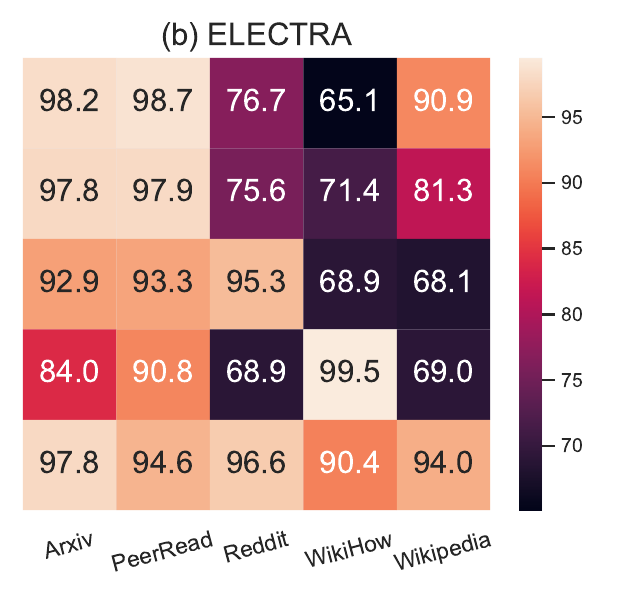}
        \includegraphics[scale=0.28]{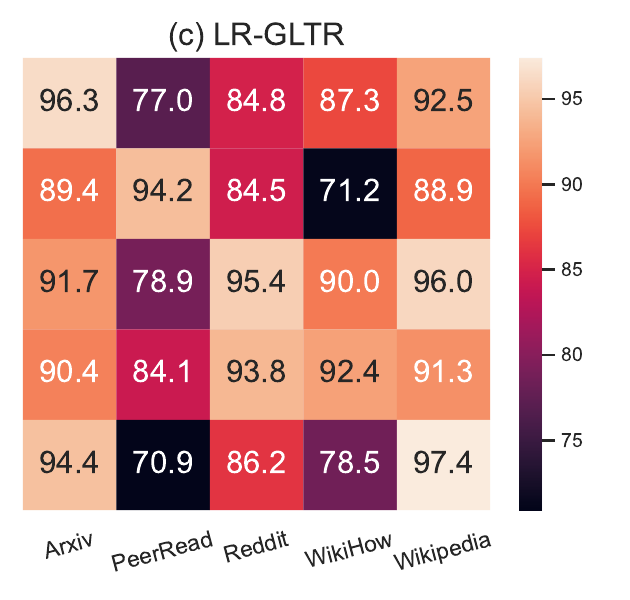}
        \includegraphics[scale=0.28]{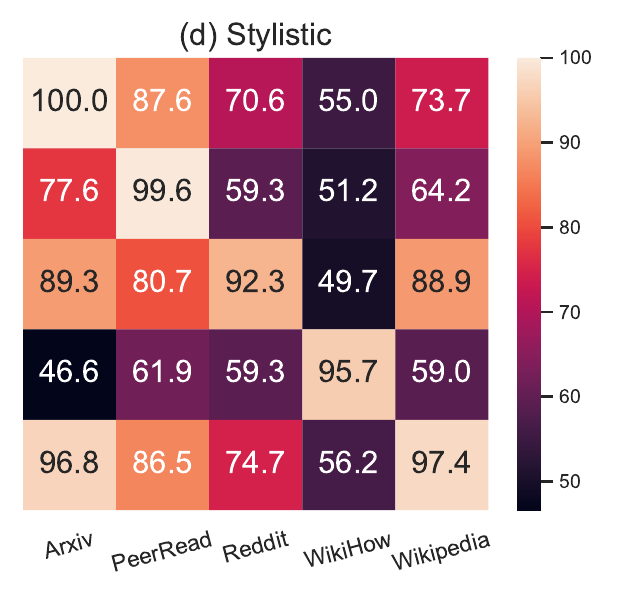}
        \includegraphics[scale=0.28]{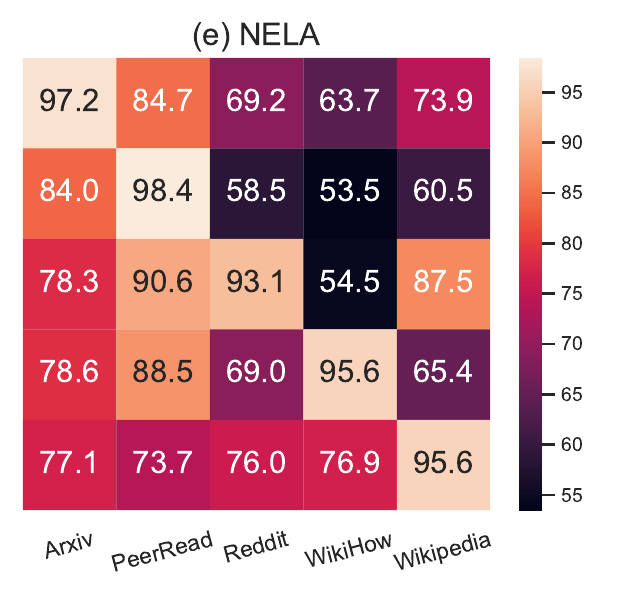} \\
        \includegraphics[scale=0.28]{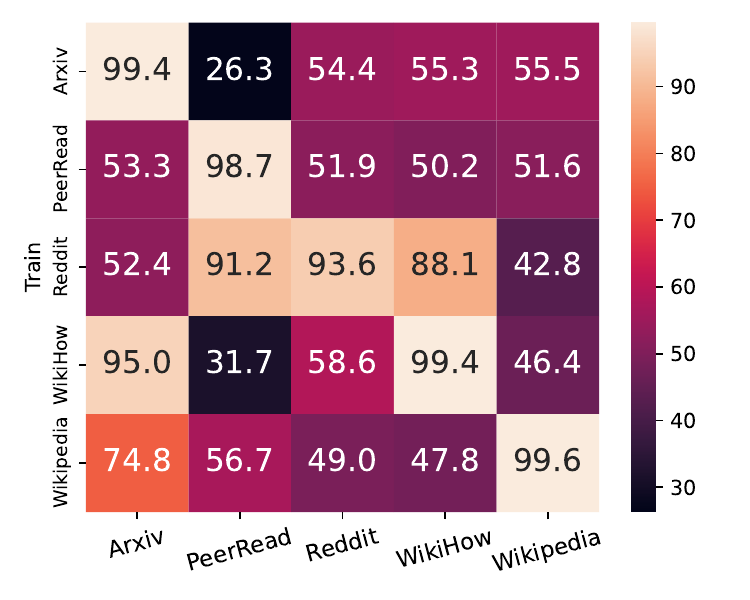} 
	\includegraphics[scale=0.28]{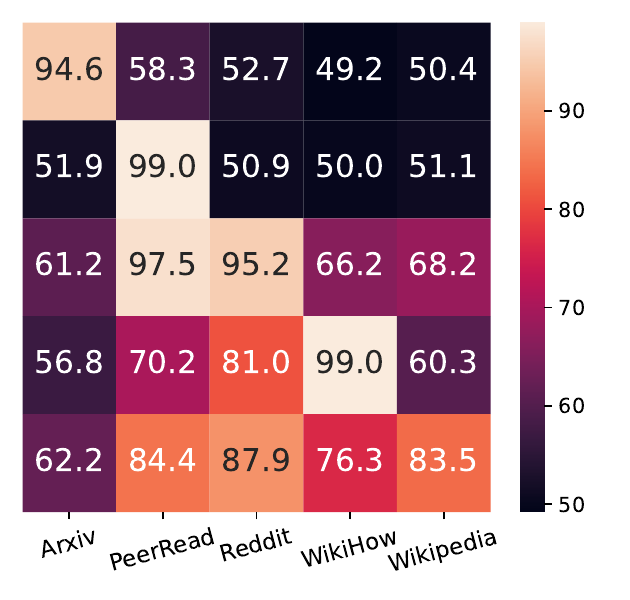}
        \includegraphics[scale=0.28]{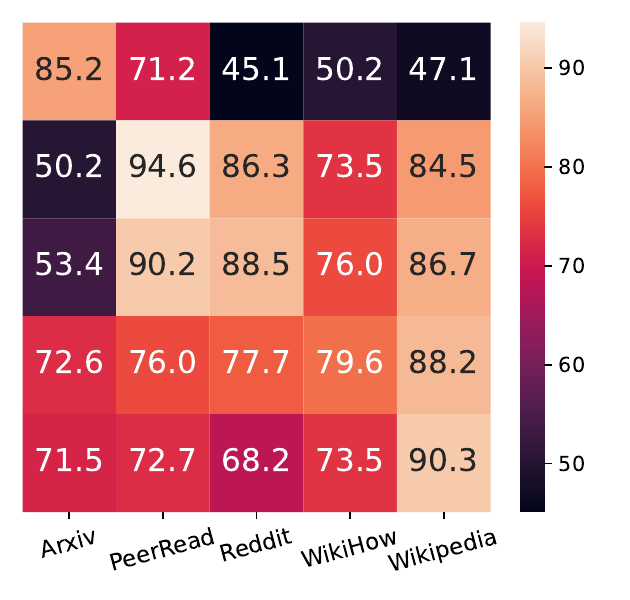}
        \includegraphics[scale=0.28]{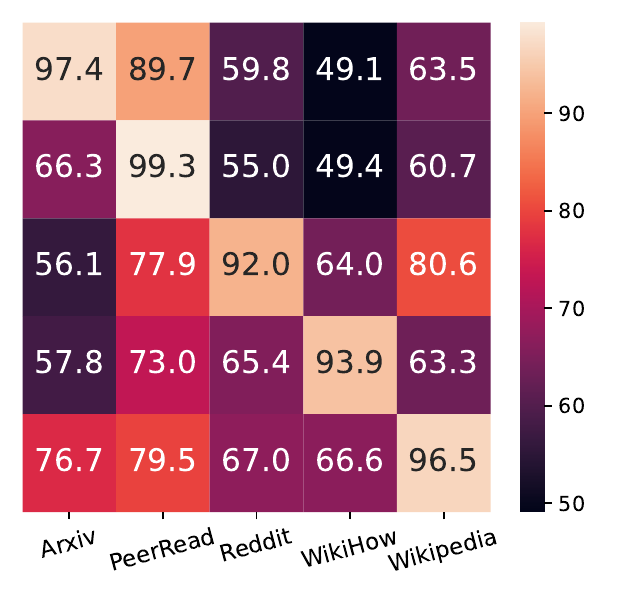}
        \includegraphics[scale=0.28]{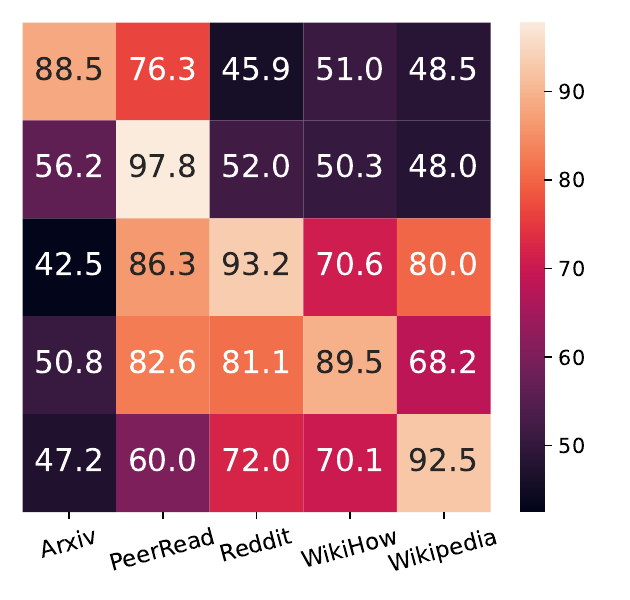} 
	\caption{\textbf{Accuracy of cross-domain experiments}: given generations from ChatGPT (top) or \textit{davinci} (bottom), train on a single domain and test across domains across five detectors. (see more detail in Tables \ref{tab:chatgpt} and \ref{tab:davinci})}
	\label{fig:samegenerator}
\end{figure*}

\paragraph{XLM-R}
We fine-tuned XLM-RoBERTa, a multilingual variant of RoBERTa~\citep{DBLP:journals/corr/abs-1911-02116}.

\paragraph{Logistic Regression with GLTR Features} 
We trained a logistic regression model based on 14 GLTR features from \citep{gehrmann-etal-2019-gltr}, which are based on the observation that most LLM decoding strategies sample high-probability tokens from the head of the distribution. Thus, word ranking information about an LLM can be used to distinguish machine-generated texts from human-written ones. We selected two categories of these features: (\emph{i})~the number of tokens in the top-10, top-100, top-1000, and 1000+ ranks from the LM predicted probability distributions (4 features), and (\emph{ii})~the Frac($p$) distribution over 10 bins ranging from 0.0 to 1.0 (10 features). Frac($p$) describes the fraction of probability for the actual word divided by the maximum probability of any word at this position.

\paragraph{Stylistic Features}
We trained an SVM classifier based on stylistic features from \citep{stylometry2014features}: (\emph{i})~character-based features, e.g.,~number of characters, letters, special characters, etc., (\emph{ii})~syntactic features, e.g.,~number of punctuation and function words, (\emph{iii})~structural features, e.g.,~total number of sentences, and (\emph{iv})~word-based features, e.g.,~total number of words, average word length, average sentence length, etc.

\paragraph{NEws LAndscape (NELA)} 
We trained an SVM classifier using the NELA features~\citep{horne2019robust}, which cover six aspects: (\emph{i})~\emph{style}: the style and the structure of the article; (\emph{ii})~\emph{complexity}: how complex the writing is; (\emph{iii})~\emph{bias}: overall bias and subjectivity; (\emph{iv})~\emph{affect}: sentiment and emotional patterns; (\emph{v})~\emph{moral}: based on the Moral Foundation Theory~\citep{grahm2012hadit}; and (\emph{vi})~\emph{event}: time and location.

\paragraph{GPTZero} 
Finally, we used the GPTZero system without any adaptation. It was trained on a large diverse corpus of human-written and AI-generated texts, focussing on English. 
The system can analyze texts ranging from individual sentences to entire documents.

%% file: section/5_experiment.tex
\section{Experiments and Results}
\label{sec:results}

In this section, we first describe our experiments, which come in three settings: (\emph{i})~same generator, cross-domain evaluation, (\emph{ii})~same domain, cross-generator evaluation, and (\emph{iii})~cross-lingual, cross-generator evaluation. As mentioned in the previous section, we also experiment with GPTZero in a zero-shot setting, as it has not seen our data (even though it might have been trained on some domains involved in our data). We further discuss the evaluation results of these experiments.

\begin{figure*}[t!]
	\centering
	\includegraphics[scale=0.28]{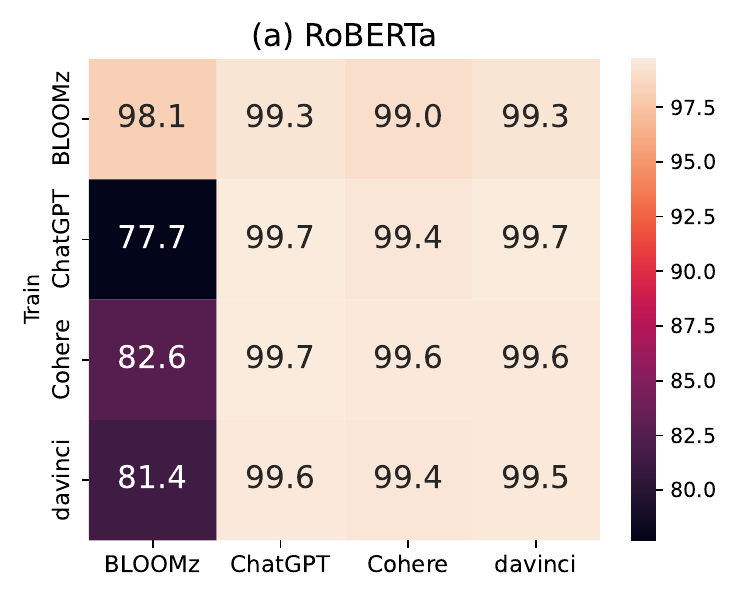} 
	\includegraphics[scale=0.28]{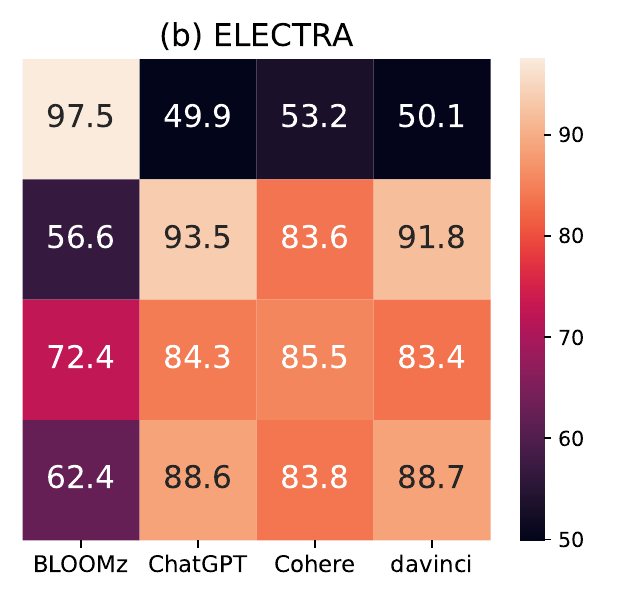}
        \includegraphics[scale=0.28]{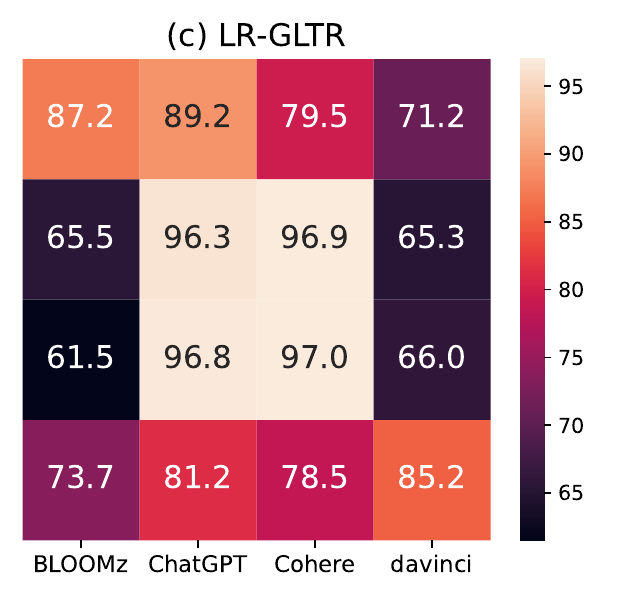}
        \includegraphics[scale=0.28]{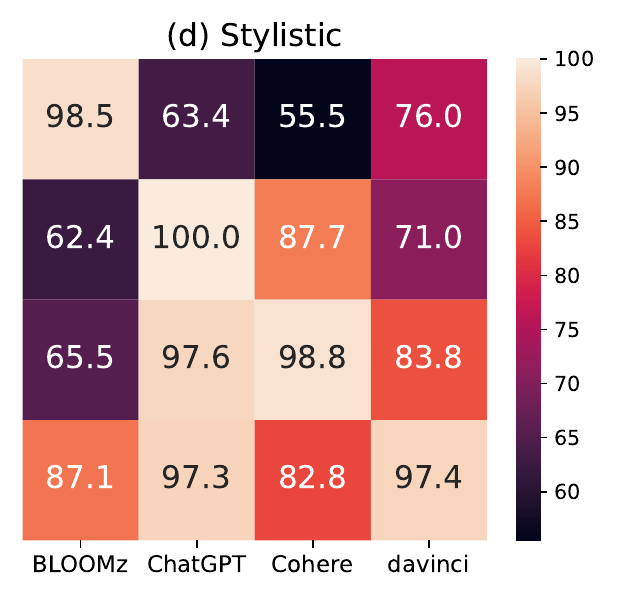}
        \includegraphics[scale=0.28]{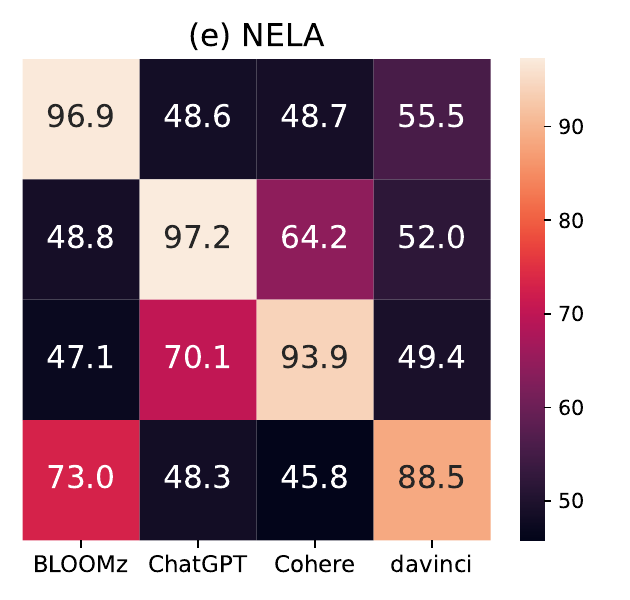} \\
        \includegraphics[scale=0.28]{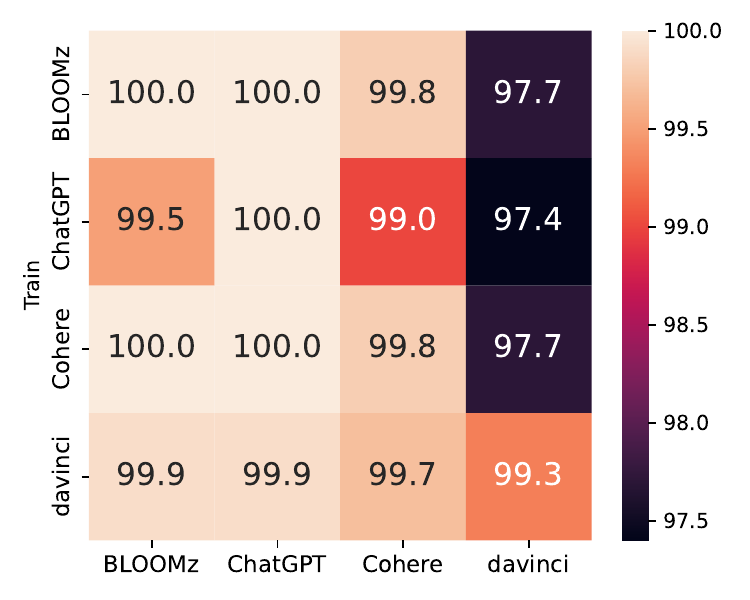} 
	\includegraphics[scale=0.28]{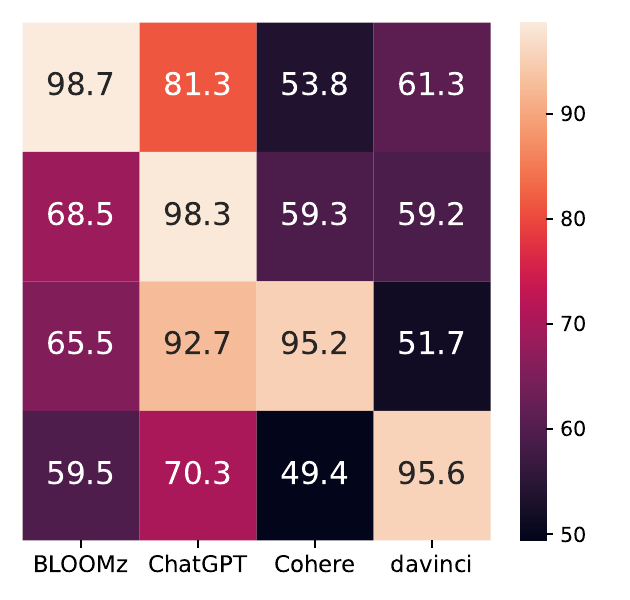}
        \includegraphics[scale=0.28]{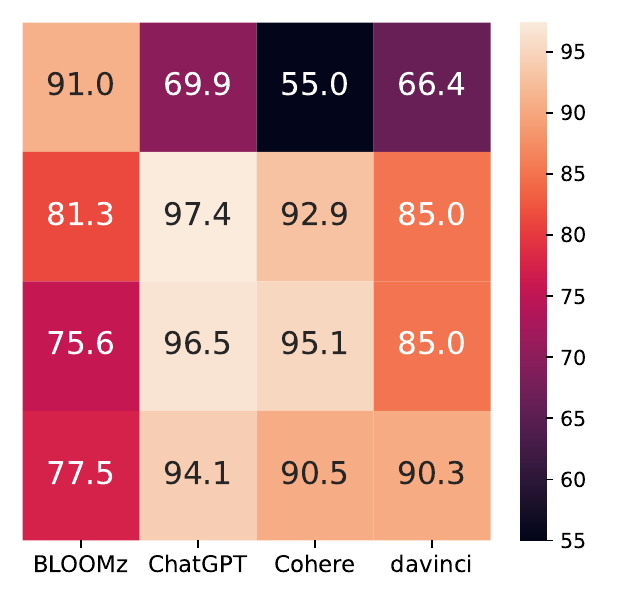}
        \includegraphics[scale=0.28]{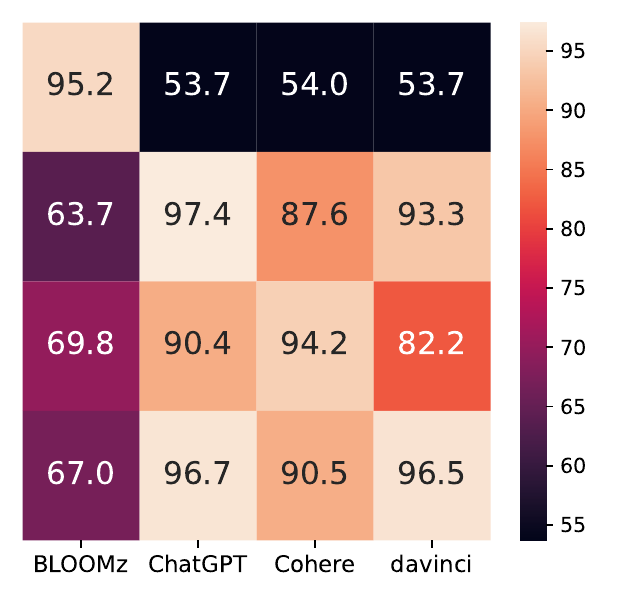}
        \includegraphics[scale=0.28]{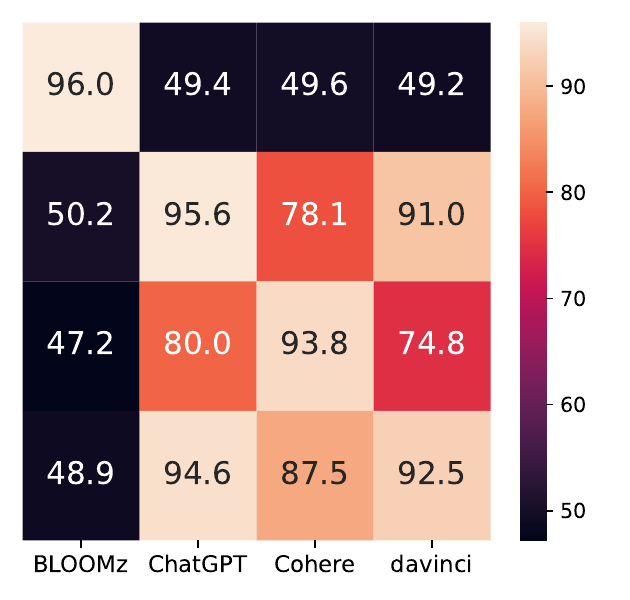} 
	\caption{\textbf{Accuracy of cross-generator experiments}: train and test on \textit{\arxiv} (top) and \textit{\wikipedia} (bottom) across five detectors, over single machine-text generator vs. human. (see detail in Tables \ref{tab:arxiv-1} and \ref{tab:wikipedia-1})}
	\label{fig:samedomain}
\end{figure*}

\subsection{Same-Generator, Cross-Domain}
\label{sec:same-generator}
Given a specific text generator, such as ChatGPT and \textit{davinci-003}, we train a detector using data from one domain and evaluate it on the test set from the same domain (in-domain evaluation) and other domains (out-of-domain evaluation). The results are shown in \figref{fig:samegenerator} and Tables \ref{tab:chatgpt} and \ref{tab:davinci}.

\textbf{In-domain detection is easy} and can be done with very high accuracy, sometimes very close to a perfect score of 100\%. This is especially the case for the RoBERTa detector, which reaches 100\% accuracy for detecting ChatGPT-generated text on \arxiv, 99.7\% on Wikipedia, 99.7\% on \wikihow, and 98.0\% on PeerRead. The only dataset where the best score for the RoBERTa detector is achieved when training on a different domain is \reddit. We can further see that the results with \textit{davinci-003} show the same pattern: all in-domain evaluation scores are usually very high, approaching 100\%. Other detectors also show high performance in the in-domain evaluation setting, but they usually overfit less to a particular domain. For example, the LR-GLTR detector shows only 79.6\% accuracy on \wikihow when the \textit{davinci-003} generator was used, while the score for the RoBERTa-based detector exceeds 99\%. 

\textbf{The best performance in the out-of-domain evaluation} is often achieved by fine-tuning ELECTRA for the task. We attribute this to the specific pre-training objective of this model, which is based on the detection of replaced tokens. ELECTRA shows slightly lower performance than RoBERTa for the in-domain evaluation, but achieves huge improvements in the out-domain evaluation setting. For example, in the case of training on \wikipedia to detect \textit{davinci-003} on Reddit, the RoBERTa's performance is close to random guessing, while ELECTRA achieves 87.9\% accuracy. Another strong approach for out-of-domain detection is LR-GLTR, which outperforms ELECTRA in some scenarios, such as detecting ChatGPT on the Wikipedia domain.

\textbf{Out-of-domain detection might be hard.} This is especially noticeable when training on \arxiv and detecting artificial texts for Reddit or training on \arxiv and detecting for Wikipedia. This is expected as these pairs of domains are very different.  There are some domains that offer better generalization than others. The RoBERTa-based detector and the detector based on NELA features are the most vulnerable in this regard. RoBERTa overfits to the training domain, while the NELA features are not tailored to machine-generated text detection, but rather initiated for fake news detection.

\textbf{The best training domain for out-of-domain generalization} is Reddit. Training on Reddit ELI5 usually yields the best out-of-domain performance. Wikipedia is also often a good domain for training. Training on \arxiv and PeerRead yields the worst generalization across other domains because the writing style of academic papers is very specific.

\textbf{The most challenging domain for machine-generated text detection} is WikiHow, while PeerRead is the easiest one.
 
\textbf{The GPT-3.5 (\textit{davinci-003}) generator is harder to detect than ChatGPT.}
Aggregating the results across all domains and both generators, we can see that the accuracy for ChatGPT is usually higher than that for \textit{davinci-003}.
This indicates that ChatGPT may leave more distinctive signals in generated texts than \textit{davinci-003}.

\textbf{Feature Analysis.} We conducted feature analysis of in-domain detectors using LIME \cite{ribeiro2016should}, and we found that detectors did not overfit to MGT artifacts and leveraged word distribution for classification. See Figure \ref{img:lime} in Appendix \ref{app:lime} for more detail.

\subsection{Same-Domain, Cross-Generator}
\label{sec:same-domain}
Given a specific domain, we train the detector using the training data from one generator and we evaluate it on the test data from the same and also from other generators. 
The accuracy on \arxiv and \wikipedia is shown in \figref{fig:samedomain} (see \tabref{tab:arxiv-1} and \ref{tab:wikipedia-1} in \secref{sec:samedomain} for precision, recall, and F1).

\textbf{RoBERTa performs the best among five detectors.}
It is the best on both \arxiv (95.9\%: average accuracy) and \wikipedia (99.4\%), followed by LR-GLTR (84.0/80.7\%), stylistic features (80.4/82.8\%), and ELECTRA (72.5/76.6\%); NELA features are the worst (73.7/64.3\%). We can see that apart from the main diagonal, most scores for the detector using NELA features are around or lower than 50.0\%, particularly on \arxiv.
This indicates that they are not suitable for distinguishing machine-generated and human-written texts. Moreover, the accuracy for \wikipedia is higher than for \arxiv, especially for RoBERTa pre-trained using \wikipedia data. This suggests that \arxiv is somewhat harder to detect than \wikipedia, and exposure bias on pre-training can impact a detectors' domain-specific performance.

\textbf{The highest accuracy is for the same generator.}
Akin to the trend of cross-domain evaluation, training and testing using the same generator always yields the best accuracy for both \arxiv and \wikipedia across the five detectors.
Even for NELA, and detection over generations by BLOOMz, the accuracy mostly remains over 90.0.
Performance drops substantially when the training and the test data are generated from different LLMs because of different distributions between the outputs of different generators.

\textbf{BLOOMz-generated text is much different from ChatGPT, \textit{davinci}, and Cohere.}
For all detectors in both \arxiv and \wikipedia, BLOOMz shows the lowest cross-generator accuracy.
Specifically, when training on BLOOMz and testing on other generators, or when training on other generators and testing on BLOOMz, it shows low recall (<0.5) for machine-generated texts.
This means that there are many false negative examples, namely, many machine-generated texts are mis\-classified as human-written ones. 
Most accuracy scores are $\leq$50.0\%, i.e.,~similar or even worse than a random guess. 
This indicates that the distribution of BLOOMz outputs is very different from the other three generators. We assume that this is because BLOOMz is primarily fine-tuned for NLP downstream data.

Moreover, we found that, for all detectors, when training on Cohere, the accuracy for ChatGPT is comparable to the accuracy on Cohere itself, and similarly high accuracy occurs when training on ChatGPT and testing on Cohere. This suggests that ChatGPT and Cohere share some generative patterns.

\input{section/tables/gptzero-table}

\input{section/tables/cross-language-table}
\subsection{Zero-shot Evaluation: GPTZero}
\tabref{tab:results_gptzero} shows that, from the perspective of the domain, GPTZero performs the best on \wikipedia, while the worst results are on \arxiv where, for all generators, the F1 score is below 50\%. From the perspective of generators, GPTZero shows the best performance on ChatGPT and the worst performance on BLOOMz.
The recall for BLOOMz is close to 0\% across all domains, which is consistent with the results for other detectors. GPTZero also demosntrated low performance for Dolly v2.
GPTZero may have been trained on generations of ChatGPT and on data from domains such as \wikipedia and \reddit, thus showing remarkable scores for them. At the same time, zero-shot detection for unseen domains and generators poses a major challenge for GPTZero.

\subsection{Multilingual Evaluation}
\label{sec:crosslingual}

In this section, we discuss the results for our multilingual experiments with the XLM-R detector across seven languages. For multilingual evaluation, we used ChatGPT and \textit{davinci-003} as generators.
The results are shown in \tabref{tab:multilingualtest_chatGPT} (see \secref{sec:mult_lingual} in the Appendix for more detail).

We constructed the English training, development, and test sets by combining English texts across all domains: \wikipedia, \wikihow, \reddit ELI5, \arxiv, and \peerread. 
Then, the \textbf{All} row refers to the combination of all training data in Arabic, Bulgarian, Chinese, English, Indonesian, Russian, and Urdu from the same generator.
We aim to evaluate the performance of a detector over each monolingual test set from a single domain when fully leveraging the available training data, thus observing the benefits brought by the interaction of multiple languages and domains.

We can see in \tabref{tab:multilingualtest_chatGPT} that the best accuracy is achieved when training and testing on the same language and using the same generator, while when training on one generator and testing on another one, the highest scores tend to appear in the row of \textbf{All}, i.e., when using the training data for all languages, except for Bulgarian (training on Bulgarian is best, if we want to test on Bulgarian).

We can also see that it is difficult for XLM-R to detect machine-generated text in a language that it has never seen during training. For example, it struggles to detect Russian, Urdu, and Indonesian machine/human-generated text when it was not trained on them. 
Interestingly, XLM-R still demonstrates good performance for Arabic even when trained on English data only.

\subsection{Time Domain Evaluation}

\begin{table}[t!]
    \centering
\scalebox{0.75}{
    \setlength{\tabcolsep}{1.5pt}

    \begin{tabular}{ l | cccc | cccc}
    \toprule
     Test  $\rightarrow$ & \multicolumn{4}{c|}{March 2023} & \multicolumn{4}{c}{September 2023} \\
     Train  $\downarrow$ & Acc & Precision & Recall & F1 & Acc & Precision & Recall & F1  \\
    \midrule 
         March & 99.5 & 99.0 & 100 & 99.5 & 99.4 & 99.0 & 99.8 & 99.4 \\
         September & 96.0 & 100 & 92.0 & 95.8 & 99.5 &  99.0 & 100 & 99.5\\
    \bottomrule
    \end{tabular} 
   }
    \caption{\textbf{Impact of ChatGPT update over time.} Accuracy (Acc), Precision, Recall, and F1 scores(\%) with respect to machine generations for \reddit from March 2023 and September 2023 ChatGPT generations based on XLM-R as a detector.}
    \label{tab:timedomain}
\end{table}

LLMs are constantly improving over time. This raises the question of the robustness of detectors for the same generator across different time points. 
With this in mind, we compared ChatGPT output generated in March 2023 (from our M4 dataset) vs. September 2023 on the \reddit-ELI5 domain and using XLM-R as a detector, and the same prompts and questions as for the M4 dataset. The results are shown in \tabref{tab:timedomain}, where we can see that the detector trained on the earlier version can effectively classify generations produced by the September 2023 version. This implies that a detector may remain effective even when applied to a newer generator trained using fresh data.

\begin{table}[t]
    \centering
    \small
    \resizebox{\columnwidth}{!}{
    \begin{tabular}{ l | ccccc }
        \toprule
         Length  $\rightarrow$ & Full Length & 1,000 & 500 & 250 & 125 \\
         \midrule
         Accuracy &  99.0 & 98.9 & 96.8 & 96.4 & 94.5 \\
         Precision &  98.2 & 97.8 & 94.2 & 94.4 & 92.5 \\
         Recall & 99.8 & 100.0 & 99.8 & 98.6 & 96.8 \\
         F1 & 99.0 & 98.9 & 96.9 & 96.5 & 94.6 \\
        \bottomrule
        \end{tabular}}
        \caption{\textbf{Impact of text length on detection accuracy} on \arxiv using XLM-R.}
    \label{tab:lengthdomain}
\end{table}

\subsection{Impact of Text Length}

Finally, we investigated the impact of text length on detection accuracy. 
We truncated \arxiv articles at the first 1,000, 500, 250, and 125 characters and compared the accuracy of XLM-R detectors trained and tested on such truncated articles for machine-generated content produced by ChatGPT. The results are shown in \tabref{tab:lengthdomain}. We can see that as the length decreases from 1,000 to 125, the accuracy drops by 4.5 points.
This illustrates the negative impact of smaller text length on detection performance; more experiments on the \arxiv and the \reddit datasets are presented in \figref{fig:lengthimpact} in the Appendix.


%% file: section/tables/gptzero-table.tex
\begin{table}
\scriptsize
    \centering
    \setlength{\tabcolsep}{3pt}
\begin{tabular}{l|rr|rr|rr|rr|rr}
\toprule
& \multicolumn{2}{c|}{\arxiv} & \multicolumn{2}{c|}{\reddit} & \multicolumn{2}{c|}{\wikihow} & \multicolumn{2}{c|}{Wikipedia} & \multicolumn{2}{c}{\peerread} \\
{} & Rec & F1 & Rec & F1 & Rec & F1 & Rec & F1 & Rec & F1 \\
\midrule
BLOOMz  & 0.4 & 0.8 & 7.6 & 13.8 & 0.0 & 0.0 & 2.0 & 3.9 & 5.8 & 10.9 \\
ChatGPT &        26.2 &   41.5 &        86.4 &    91.6 &          49.4 &     62.1 &            87.2 &       93.1 & 70.8 & 82.7 \\
\textit{davinci} &        0.2 &    0.4 &          60.4 &    74.3 &                 45.2 & 59.4 & 53.8 &       70.0 & 96.2 & 97.9 \\
Cohere  &18.6 &   31.4 &          30.2 &       44.5 &         68.0 &        77.9 &          69.0 &       81.7 & 84.4 & 91.3 \\
Dolly v.2 & 5.4 & 10.3 &              52.8 &           66.7 &       13.6 &  21.1  &               29.4 &       45.4 & 18.6 & 31.3 \\
    \bottomrule
    \end{tabular}
\caption{Zero-shot detection with GPTZero: recall (Rec) and F1-score with respect to generators and domains.}
\label{tab:results_gptzero}
\end{table}

%% file: section/tables/cross-language-table.tex
\begin{table*}[t!]
    \centering
\scalebox{0.8}{
            \setlength{\tabcolsep}{1pt}
    \begin{tabular}{ll | c c c c c c c| c c c c}
    \toprule
        Generator  &$\rightarrow$& \multicolumn{7}{c|}{ChatGPT} & \multicolumn{4}{c}{davinci-003}\\
         $\downarrow$&Test Domain $\rightarrow$ & All  & Baike/  & Ru  & Bulgarian  & IDN  & Urdu & Arabic & All  & Baike/  & Ru  & Bulgarian \\
        &Train Domain $\downarrow$ & domain & Web QA &ATD & News & & -News & Wikipedia & domain & Web QA  & ATD & News\\
        & &  (en) &  (zh) & (ru) &  (bg) & (id) & (ur) &  (ar) & (en)  &  (zh)  & (ru)  &  (bg)\\
    \midrule 

& All domains (en) & \textbf{98.6} & 97.5 & 76.6 & 80.8 & 76.9 & 57.7  & 96.5 & 90.2 & 93.0 & 54.1 & 66.0 \\
 & Baike/Web QA (zh) & 61.8 & \textbf{99.4} & 63.1 & 65.0 & 64.1 & 81.8 & 62.7 & 61.6 & 93.5 & 58.8 & 57.7\\
& RuATD (ru) & 59.1 & 92.6 & \textbf{97.5} & 81.7 & 76.9 & 55.5 & 86.2 & 56.7 & 75.7 & 84.7 & 82.2\\
ChatGPT & Bulgarian News (bg) & 83.8 & 87.8 & 83.7  & \textbf{96.9}  & 92.6 & 64.9 & 88.3 & 74.2 & 78.3 & 53.8 & \textbf{95.4}\\
& IDN (id) & 65.9 & 59.9 & 62.6 & 67.6 & \textbf{98.4} & 50.6 & 54.6 & 61.0 & 55.6 & 50.6 & 58.7\\
& Urdu-News (ur) & 50.0 & 51.0 & 50.0 & 50.3  & 50.1 & \textbf{99.9} & 50.5 & 50.0 & 50.8 & 50.0 & 50.2\\
& Arabic Wikipedia (ar) & 76.4 & 87.0 & 66.0 & 65.5 & 68.9 & 67.7  & \textbf{96.8} & 72.8 & 83.9 & 62.0 & 64.6\\
& \textbf{All} & 98.3 & 99.1 & 95.4 & 83.4 & 97.3 & \textbf{99.9} & 96.7 & \textbf{91.3} & \textbf{94.5} & \textbf{86.1} & 82.6\\
\midrule
& All domains (en) & 95.9 & 79.7 & 70.4 & 72.4 & 67.2 & 61.1 & 93.1& 95.8 & 79.5 & 60.5 & 65.8\\
 & Baike/Web QA (zh) & 66.8 & \textbf{98.0} & 62.0 & 57.1 & 57.3  & \textbf{83.0} & 76.1 & 66.4 & \textbf{98.9} & 59.5 & 48.6\\
davinci-003 & RuATD (ru) & 61.4 & 60.5 & 88.6 & 72.4 & 58.6 & 49.7 & 68.9 & 62.8 & 49.6 & \textbf{95.3} & 86.5\\
& Bulgarian News (bg) & 64.9 & 69.3  & 61.5 & \textbf{84.9} & 64.7 & 66.4 &  73.8 & 64.8 & 59.0  & 59.0  & \textbf{99.6}\\
& \textbf{All} & \textbf{96.4} & 95.5 & \textbf{94.3} & 83.3 & \textbf{74.5} &  76.1  & \textbf{93.3}& \textbf{96.3}  & 98.7 & 92.8  & 85.2\\
    \bottomrule
    \end{tabular}
    }
    \caption{Accuracy (\%) based on XLM-R on test sets across different languages over ChatGPT and \textit{davinci-003}.}
    \label{tab:multilingualtest_chatGPT}
\end{table*}

%% file: section/6_conclusion.tex
\section{Conclusion and Future Work}
\label{sec:conclusion}

We presented M4, a large-scale multi-generator, multi-domain, and multi-lingual dataset for machine-generated text detection.
We further experimented with this dataset performing a number of cross-domain, cross-generator, cross-lingual, and zero-shot experiments using seven detectors. 
We found that detectors struggle to differentiate between machine-generated and human-written texts if the texts come from a domain, a generator, or a language that the model has not seen during training.
Our results show that the problem is far from solved and that there is a lot of room for improvement.
We hope that our release of M4, which we make freely available to the community, will enable future research towards more robust approaches to the pressing societal problem of fighting malicious machine-generated text. We have already created an extension of M4 for SemEval-2024 Task~8 \citep{semeval2024task8},\footnote{\url{https://github.com/mbzuai-nlp/SemEval2024-task8}} which features additional languages, domains, and three new task (re)formulations.

In future work, we plan to expand our M4 dataset continuously by introducing new LLM generators, by exploring different domains, by incorporating new languages, and by diversifying the range of tasks and prompts used. We believe that this is a good, practical way to keep the dataset up-to-date in response to the ongoing progress in LLMs. Our aim is to maintain a dataset that remains relevant as LLMs continue to evolve.

%% file: section/7_limitations.tex
\section*{Ethics and Broader Impact}

Below, we discuss some potential ethical concerns about the present work.

\paragraph{Data Collection, Licenses, and User Privacy.}

Creating the M4 dataset did not involve scraping raw data from websites. Instead, we used pre-existing corpora that have been publicly released and approved for research purposes, with clear dataset licenses, which are listed in \tabref{tab:dataset}. To the best of our knowledge, all included datasets adhere to ethical guidelines and minimize privacy concerns. Since the human-written data has already been published and made publicly available for research purposes, we see no additional privacy risks in releasing that data as part of our M4 dataset. 

The human text components of M4 are publicly available and can be freely accessed and used for research purposes. However, researchers must acknowledge the original sources of the text and comply with the respective licensing terms.

The machine-generated text components of our M4 dataset are subject to the licensing terms of the underlying LLMs. For text generated using LLMs, researchers must comply with the respective licensing terms of those LLMs:

\begin{itemize}
    \item \davinci, \chatgpt, \gptfour: no specific license. They welcome research publications related to the OpenAI API.\footnote{\url{https://openai.com/policies/sharing-publication-policy}}
    \item \dolly: Apache 2.0 \footnote{\url{https://github.com/databrickslabs/dolly}}
    \item \cohere: no specific license. They point out that CUSTOMER RETAINS ALL OWNERSHIP AND INTELLECTUAL PROPERTY RIGHTS IN AND TO CUSTOMER DATA.\footnote{\url{https://cohere.com/saas-agreement}}
    \item \bloomz: Apache 2.0 \footnote{\url{ https://github.com/bigscience-workshop/xmtf}}
\end{itemize}

\paragraph{Potential Biases}
We recognize the potential for biases in our M4 dataset, stemming from both the original human-written corpora and the Large Language Models (LLMs) used for generation. This is an important issue, and we put efforts to minimize such biases. However, we are aware that unethical usage of our dataset may still lead to biased applications: even if our original dataset was completely unbiased, external parties may extract a biased subset, which would be out of our control. 

Having already realized these concerns, we have implemented the following measures:
\begin{itemize}
    \item[a.] We provide comprehensive documentation about our M4 dataset, including detailed information about the sources of all human-written corpora, the generation process for obtaining the machine-generated text, including the full prompts and the measures we took to cleanse the output, and the potential biases that may exist. We believe that this transparency would allow researchers to understand the origins of the data and to make informed decisions about how to use it.
    \item[b.] We further acknowledge and transparently discuss these limitations and debiasing techniques that could be used to address these limitations. We hope that the strong emphasis on transparency in our methodology by explicitly stating the sources of human-written corpora and the generation processes for the corresponding machine-generated text could help clarify the dataset's origins and potential biases.
\end{itemize}

\paragraph{Robustly Secure System}
The M4 dataset is intended for the development of detection systems to mitigate misuse, particularly in the context of malicious content generated using LLMs. While we encourage extensive and responsible use of the datasets to advance this critical area of research, we also emphasize the importance of adhering to the licensing terms of the original human-written corpora and the corresponding LLMs.

\section*{Limitations}
In this section, we discuss some perceived limitations of our study.

\subsection*{M4 Dataset Generalization and Biases}

\paragraph{Generalization:} Machine-generated outputs exhibit a high degree of sensitivity to the prompts. While our M4 dataset was collected with diverse prompts for a variety of generators, domains, and languages, to cover typical use cases, it has limitations as a general resource, as it is neither sufficient to train a detector that can be expected to generalize well across all possible domains and generators, nor is it likely sufficient to act as a standard benchmark that can accurately evaluate the effectiveness of a detection method.

\paragraph{Up-to-Date:} Detecting machine-generated text is a very challenging task when we do not know in advance the potential generator and the domain: as our findings show, human-written and machine-generated text cannot be distinguished in certain situations, e.g.,~we saw issues when using text generated by BLOOMz. 
Therefore, we regard M4 as a useful repository of machine-generated text for researchers who want to improve and to evaluate their detectors from multiple dimensions.
Moreover, the LLMs are constantly evolving, and thus any dataset collected for machine-generated text detection can become outdated relatively fast. 
With this in mind, we have constantly been extending the M4 dataset (e.g., with a recent collection of GPT-4 responses), and we expect to grow our repository to enable better training and more up-to-date detectors.

\paragraph{Bias:} Biases may exist in both human-written and machine-generated texts, and it is possible that our M4 dataset may be influenced by biases from human collection, thus affecting the detection outcomes.
We leave the analysis of such biases to our future work.

\subsection*{Feasibility of Black-Box Machine-Generated Text Detection}

A growing body of work shows that machine-generated text detection might gradually become harder and even nearly impossible: as LLMs evolve, the gap between machine-generated and human-written text might narrow~\citep{tang2023science, sankar2023canai}.
\citet{liang2023gpt} further suggested that GPT detectors are biased against non-native English writers.
These findings continue to release unpromising signals for black-box detection approaches.
Yet, alternatives such as watermarking or white-box methods remain impractical for proprietary LLMs, where general users and practitioners cannot access the model-internal parameters.
Current black-box approaches may be less effective and may demonstrate poor generalization for unseen domains, generators, and languages, and this suggests the need to study more general methods to improve the detection and the potential misuse of LLMs. 

\section*{Acknowledgments}
We thank the anonymous reviewers and the program committee chairs for their very helpful and insightful comments, which have helped us improve the paper.

%% file: section/8_appendix.tex
\clearpage
\onecolumn
\section*{Appendix}
\appendix

\section{Data Collection and Analysis}
\label{sec:dataset}
\subsection{English Corpora} 
\label{sec:englishsource}

\paragraph{Wikipedia}
We used the \wikipedia dataset available on HuggingFace\footnote{\url{https://huggingface.co/datasets/wikipedia}} 
and randomly chose 3,000 articles, each of which surpasses a character length limit of 1,000.
We prompted LLMs to generate \wikipedia articles given titles, with the requirement that the output articles contain at least 250 words.
For generation with Dolly-v2,\footnote{\url{https://huggingface.co/databricks/dolly-v2-12b}} 
we set the minimum number of generated tokens to be 300 to satisfy the minimal character length of 1,000.

\paragraph{Reddit ELI5}
dataset~\citep{eli5_lfqa} is a collection of English question-answering (QA) pairs,\footnote{\url{https://huggingface.co/datasets/eli5}} 
gathered to facilitate open-domain and long-form abstractive QA.
The data is derived from three categories: \textit{ExplainLikeImFive} for general topics, \textit{AskScience} for scientific queries, and \textit{AskHistorians} for historical inquiries. 
Each pair is composed of a question (a title + a detailed description) and corresponding answers. 
We filtered out answers with less than 1,000 characters, retaining questions whose title ends with a question mark without detailed descriptions. 
Finally, we selected 1,000 QA pairs with top user ratings for each category, resulting in a total of 3,000 
pairs.


\paragraph{WikiHow}
dataset\footnote{\url{https://huggingface.co/datasets/wikihow}} \citep{DBLP:journals/corr/abs-1810-09305} is built from the online WikiHow knowledge base. It consists of articles with a title, a headline (the concatenation of all bold lines of all paragraphs), and text (the concatenation of all paragraphs except the bold lines). 
We randomly chose 3,000 articles with the length of more than 1,000 characters and prompted LLMs with titles and headlines to generate artificial articles. 

\paragraph{PeerRead Reviews}
We sampled 586 academic papers published in top-tier NLP and machine learning conferences from the \peerread corpus~\citep{kang-etal-2018-dataset}. Each paper contains metadata, including title, abstract, and multiple human-written reviews.
Given a paper, we prompted LLMs to generate peer reviews with four different instructions; two depend only on the title and another two involve both the title and the abstract.
Two prompts specify the review format of first describing what problem or question the considered paper addresses, and then providing its strengths and weaknesses. Other two prompts do not contain a review format specification.\footnote{As we do not consider hallucinations in the context of machine-generated text detection, we manipulated peer reviews relying on paper title and abstract, instead of its content.}
This resulted in 584 $\times$ 4 = 2,344 machine-generated texts for each generator and 5,798 human-written reviews in total.

\paragraph{Arxiv Abstract} parallel dataset is constructed from a Kaggle corpus.\footnote{\url{https://www.kaggle.com/datasets/Cornell-University/arxiv}} 
We sampled 3,000 abstracts with a minimum length of 1,000 characters and prompted LLMs to produce machine-generated abstracts based on their titles.

\subsection{Corpora in Other Languages}
\label{sec:otherlanguagesource}
\paragraph{Arabic Wikipedia.}
Similarly to English \wikipedia, we randomly selected 3,000 Arabic articles with a length exceeding 1,000 characters and prompted the LLMs to generate artificial articles based on their titles.

\paragraph{Bulgarian True \& Fake News} is sampled from the Hack the Fake News datathon\footnote{\url{https://gitlab.com/datasciencesociety/case_fake_news/-/tree/master/}} 
organized in 2017 by the Data Science Society in Bulgaria. It is a mixture of real and fake news. 
The human partition consists of 3,000 news articles with a length of more than 1,000 characters. We obtained machine-generated texts by prompting LLMs with titles of human-written articles.

\paragraph{Chinese QA} is constructed from 3,000 question--answer pairs sampled from Baike and the Web QA corpus.\footnote{\url{https://github.com/brightmart/nlp_chinese_corpus}} 
The length of each answer is more than 100 Chinese characters.
We prompted LLMs with a combination of a brief title and a detailed description for each question.

\paragraph{Indonesian News 2018} is constructed from a corpus of Indonesian news articles\footnote{\url{https://huggingface.co/datasets/leid_newspapers_2018}} 
collected from seven different news websites in 2018. We picked news from CNN Indonesia since this source was found to provide the cleanest data. 
We selected 3,000 texts from the corpus and generated artificial news articles by prompting ChatGPT with a title. 

\paragraph{Russian RuATD} is sourced from the RuATD Shared Task 2022 \cite{shamardina2022ruatd} devoted to artificial text detection in Russian. \citet{shamardina2022ruatd} gathered a vast human and machine-generated corpora from various text generators.
However, these generators are either task-specific or domain-specific.
We leveraged their human-written texts collected from publicly available resources and re-generated the machine-authored data using the open-domain state-of-the-art multilingual LLMs. 
In particular, for the construction of human-written data, the task organizers used the following sources: (1) diachronic sub-corpora of the Russian National Corpus\footnote{\url{https://ruscorpora.ru/old/en/index.html}}, which covers three historical periods of the society and the Modern Russian language (``pre-Soviet'', ``Soviet'', and ``post-Soviet''); (2) several social media platforms; (3) top-100 most viewed Russian Wikipedia pages spanning the period of 2016-2021 according to the PageViews statistics; (4) news articles from the Taiga corpus \cite{shavrina2017taiga} and the ``corus'' library\footnote{\url{https://github.com/natasha/corus}}; (5)  a corpus of digitalized personal diaries ``Prozhito'' written during the 20th century \cite{prozhito2017}; (6) government documents from the RuREBus Shared Task \cite{ivanin2020rurebus}.

\paragraph{Urdu News} is derived from Urdu News Data 1M --- a collection of one million news articles from four distinct categories: Business \& Economics, Science \& Technology, Entertainment, and Sports. These articles were gathered from four reputable news agencies in Pakistan \cite{Hussain2021}. Each entry in this dataset includes a headline, a category, and a news article text. 
To ensure the data balance over four categories, we randomly sampled 750 news articles from each, resulting in 3,000 examples in total.
Using the headlines as prompts, we generated the content of artificial news articles.

\subsection{LLM Generation}
\paragraph{Prompt Diversity} 
We paid attention to prompt diversity, using multiple (2-8) prompts for each domain--generator combination in English, with the aim to produce diverse outputs that are more aligned with divergent generations in real-world application scenarios. See Table~\ref{tab:num-prompts} for detailed statistics about the prompts.
\\
\\
\textit{\textbf{Prompts of \peerread}}
\begin{itemize}
    \item Please write a peer review for the paper + title;
    \item Write a peer review by first describing what problem or question this paper addresses, then strengths and weaknesses, for the paper + title;
    \item Please write a peer review for the paper + title, its main content is as below: + abstract;
    \item Write a peer review by first describing what problem or question this paper addresses, then strengths and weaknesses, for the paper + title, its main content is as follows: + abstract.
\end{itemize}
\begin{table*}[ht]
    \centering
    \small
    \resizebox{\textwidth}{!}{
    \begin{tabular}{l|ccccc|c}
    \toprule
      Domain$\downarrow$ & davinc-003 &  ChatGPT &  Cohere &  Dolly-v2 &  Bloomz &  Unique across domain \\
        \midrule
       wikipedia & 1 &   1 &     1 &       1 &     2 &  3 \\
          Reddit & 5 &   5 &     1 &       1 &     1 &  8 \\
         wikihow & 1 &   1 &     1 &       1 &     2 &  3 \\
        peerread & 4 &   4 &     4 &       4 &     4 &  4 \\
           arxiv & 1 &   5 &     1 &       1 &     2 &  8 \\
           \midrule
    baike/web QA & 1 &   1 &     Na &       Na &     Na &     1 \\
           RuATD & 1 &   1 &     Na &       Na &     Na &     1 \\
  True Fake news & 1 &   1 &     Na &       Na &     Na &     1 \\
       Urdu-news & Na &   1 &     Na &       Na &     Na &     1 \\
    id\_newspaper& Na &   1 &     Na &       Na &     Na &     1 \\
Arabic wikipedia & Na &    1 &     Na &       Na &     Na &     1 \\
\bottomrule
\end{tabular}}
    \caption{Statistics about the prompts for different domains and LLMs. One prompt is used for non-English text, and multiple prompts are used for English. The number of prompts for different domains varies as shown in the last column. Given a domain, some models might not follow all designed instructions, leading to less variety of prompts.}
    \label{tab:num-prompts}
\end{table*}

\paragraph{Hyper-Parameter Values for the Generators}
\tabref{tab:hyperparameters} shows the values of the hyper-parameters we used for the various generators. In general, we followed the default setting, except for the length of new generations in order to satisfy the minimum character length of 1,000.

\begin{table*}[ht]
    \centering
    \small
    \resizebox{\textwidth}{!}{
    \addtolength{\tabcolsep}{-1pt}
        \begin{tabular}{ll|lllllHH}
        \toprule
        \textbf{Source/} & \textbf{Language} & \multicolumn{7}{c}{\textbf{Generator}} \\
        \textbf{Domain} &                   & Davinci003    & ChatGPT & Cohere & Dolly-v2 & BLOOMz & FlanT5 & LLaMa \\ \midrule
        Wikipedia & English                 & max\_tokens=1000        & max\_tokens=1000 & max\_tokens=1000 & \begin{tabular}[c]{@{}l@{}}min\_new\_tokens=300, \\ max\_new\_tokens=1000\end{tabular} 
        & default & - & Yuxia \\ \midrule
        Reddit ELI5 & English               & default & default & default & \multirowcellii{
        min\_new\_tokens=180 \\ \midrule
        max\_new\_tokens=600 }
 & min\_new\_tokens=180 & \begin{tabular}[c]{@{}l@{}}min\_len=50, \\ max\_len=200\end{tabular} & Osama \\ \midrule
        WikiHow & English                   & max\_tokens=2000 & default & default & \multirowcellii{
        min\_new\_tokens=200 \\ 
        max\_new\_tokens=1000 } & min\_new\_tokens=200 & - & Yuxia \\  \midrule
        PeerRead & English                  & default & default & default & default & min\_new\_tokens=150 & - & \begin{tabular}[c]{@{}l@{}}min\_len=50, \\ max\_len=400\end{tabular} \\ \midrule
        arXiv abstract & English            & max\_tokens=3000 & default & default & \multirowcellii{ min\_new\_tokens=180 \\ max\_new\_tokens=600 }
  & \begin{tabular}[c]{@{}l@{}}min\_new\_tokens=180,\\ max\_new\_tokens=420,\\ repitition\_penalty=1.15,\\ length\_penalty=10\end{tabular} & \begin{tabular}[c]{@{}l@{}}min\_len=50, \\ max\_len=200\end{tabular} & Osama \\ \midrule
        Baike/Web QA & Chinese              & default & default & - & - & - & - & - \\ \midrule
        RuATD & Russian & max\_tokens=1700 & default & - & - & - & - & - \\ \midrule
        Urdu-news & Urdu                    & - & temperature=0 & - & - & - & - & - \\ \midrule
        id\_newspapers\_2018 & Indonesian   & - & default & - & - & - & - & - \\ \midrule
        Arabic-Wikipedia & Arabic           & - & default & - & - & - & - & - \\ \midrule
        Bulgarian True \& Fake News & Bulgarian & max\_tokens=3000 & default & - & - & - & - & - \\ \bottomrule
        \end{tabular}
        }
    \caption{Hyper-parameter values used to generate data. We only specify the values that are different from the default.}
    \label{tab:hyperparameters}
\end{table*}

\newpage
\subsection{N-gram Analysis}

Table~\ref{tab:ngramanalysis} shows statistics about the number of unique uni-grams (word types) and bi-grams of human-written and machine-generated texts (English). Table~\ref{tab:unique-ngram-each-doc} shows the number of per-document unique uni-grams (word types) and bi-grams of human-written and machine generated texts (English).

\label{sec:ngram}
\begin{table*}[ht]
    \centering
    \small
    \resizebox{\textwidth}{!}{
    \begin{tabular}{l|r|rrrrr|r|rrrrr}
    \toprule
      & \multicolumn{6}{c|}{\textbf{Word (unigram)}} & \multicolumn{6}{c}{\textbf{bigrams}} \\
      Domain$\downarrow$ & Human & ChatGPT &  davinc-003 &  Cohere &  Dolly-v2 &  BLOOMz & Human & ChatGPT &  davinc-003 &  Cohere &  Dolly-v2 &  BLOOMz \\
        \midrule
       \wikipedia & 144,523 & 45,275 & 59,038 & 47,092 & 65,059 & 34,304 &
        1,000,870 & 295,007 & 400,072 & 258,210 & 385,074 & 141,328 \\
          \reddit &  69,406 & 27,403 & 33,292 & 24,134 & 36,173 & 28,794 &
        586,341 & 253,075 & 315,567 & 183,926 & 308,695 & 212,334 \\
         \wikihow &  84,651 & 49,723 & 47,307 & 29,062 & 46,743 & 40,082 & 
         820,026 & 501,998 & 457,188 & 243,356 & 357,007 & 277,770 \\
        \peerread &  24,317 & 11,314 &  7,693 &  8,812 & 29,851 & 11,597 & 
        225,007 & 102,638 & 51,636 & 61,310 & 230,282 & 92,858 \\
           \arxiv &  36,202 & 18,291 & 29,024 & 22,777 & 35,808 & 29,989 & 
           263,781 & 145,954 & 186,561 & 149,892 & 251,770 & 209,053\\
           \midrule
           All domains & 252,244 & 95,775 & 115,482 & 87,428 & 139,981 & 96,789 & 2,364,143 & 1,047,293 & 1,145,593 & 733,902 & 1,220,512 & 775,387\\
           \midrule
      All & 252,244 & \multicolumn{5}{c|}{275,455} & 2,364,143 & \multicolumn{5}{c}{3,074,950} \\
\bottomrule
\end{tabular}}
    \caption{Statistics about the number of unique uni-grams (word types) and bi-grams of human-written and machine-generated texts (English).}
    \label{tab:ngramanalysis}
\end{table*}

\begin{table*}[ht]
    \centering
    \small
    \resizebox{\textwidth}{!}{
    \begin{tabular}{l|r|rrrrr|r|rrrrr}
    \toprule
      & \multicolumn{6}{c|}{\textbf{Word (unigram)}} & \multicolumn{6}{c}{\textbf{bigrams}} \\
      Domain$\downarrow$ & Human & ChatGPT &  davinc-003 &  Cohere &  Dolly-v2 &  BLOOMz & Human & ChatGPT &  davinc-003 &  Cohere &  Dolly-v2 &  BLOOMz \\
        \midrule
        \wikipedia & 334 & 158 & 189 & 142 & 167 & 77 &  
        683 & 274 & 337 & 259 & 296 &  93 \\
          \reddit & 250 & 140 & 159 & 107 & 142 & 134 &  
        482 & 247 & 292 & 191 & 254 & 164 \\ 
         \wikihow & 369 & 277 & 250 & 143 & 160 & 174 &  
         867 & 580 & 514 & 270 & 294 & 225 \\
        \peerread &  142 & 151 &  90 &  82 & 178 & 133 & 
        244 & 262 & 146 & 129 & 332 & 154 \\ 
        \arxiv &  128 & 121 &  96 &  97 & 130 & 159 & 
        208 & 199 & 142 & 168 & 219 & 218 \\
        \midrule 
        All domains & 228 & 170 & 160 & 115 & 154 & 136 &
                    457 & 315 & 293 & 207 & 277 & 172 \\
        \midrule
      All & 228 & \multicolumn{5}{c|}{147} & 457 & \multicolumn{5}{c}{252} \\
\bottomrule
\end{tabular}}
    \caption{Statistics about per-document unique uni-grams (word types) and bi-grams of human-written and machine generated texts (English).}
    \label{tab:unique-ngram-each-doc}
\end{table*}

\clearpage
\section{Detectors}
\label{sec:parameters}
\input{section/tables/hyperparameters_table}

\subsection{Computation Resources and Cost}
We spent \$600 on calls of OpenAI APIs for ChatGPT and \textit{davinci-003} generations, \$40 on calls to GPTZero. We spent about 2,500 GPU hours on Dolly-v2 and BLOOMz generation.

\clearpage
\section{Results: Same-Generator, Cross-Domain}
\label{sec:samegenerator}

\input{section/tables/generator_eval_tables}

\clearpage
\section{Results: Same-Domain, Cross-Generator}
\label{sec:samedomain}
\input{section/tables/domain_eval_tables}

\clearpage
\section{Results: Multilingual Evaluation}
\label{sec:mult_lingual}
\input{section/tables/detailed_cross_lingual_table}

\clearpage
\section{Results: Impact of Text Length}
\label{sec:textlen}

\begin{figure}[h!]
\centering
\includegraphics{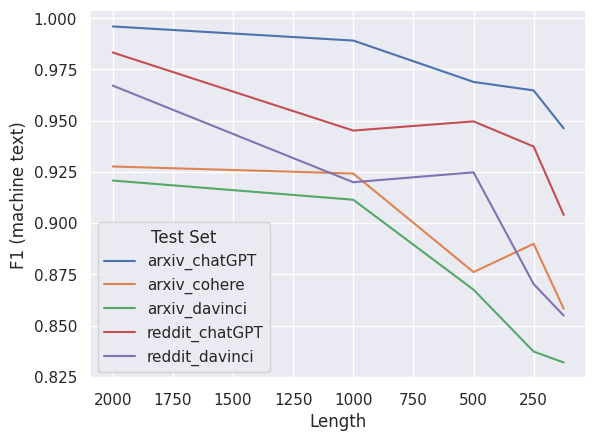}
\caption{\textbf{Impact of text length on detection accuracy} over \arxiv and \reddit generated by ChatGPT, \textit{davinci}, and Cohere.}
\label{fig:lengthimpact}
\end{figure}

Figure~\ref{fig:lengthimpact} shows detailed results from experiments to study the impact of text length on detection accuracy, over \arxiv and \reddit generated by ChatGPT, \textit{davinci}, and Cohere. We can see that as the character length decreasing from 1000 to 125, the F1-score with respect to machine-generated text decreases for all subsets, demonstrating the negative impact of short text on detection performance.

\clearpage
\section{Feature Analysis with LIME}
\label{app:lime}

\begin{figure}[ht]
\begin{subfigure}{1\textwidth}
  \centering
  \includegraphics[width=\linewidth]{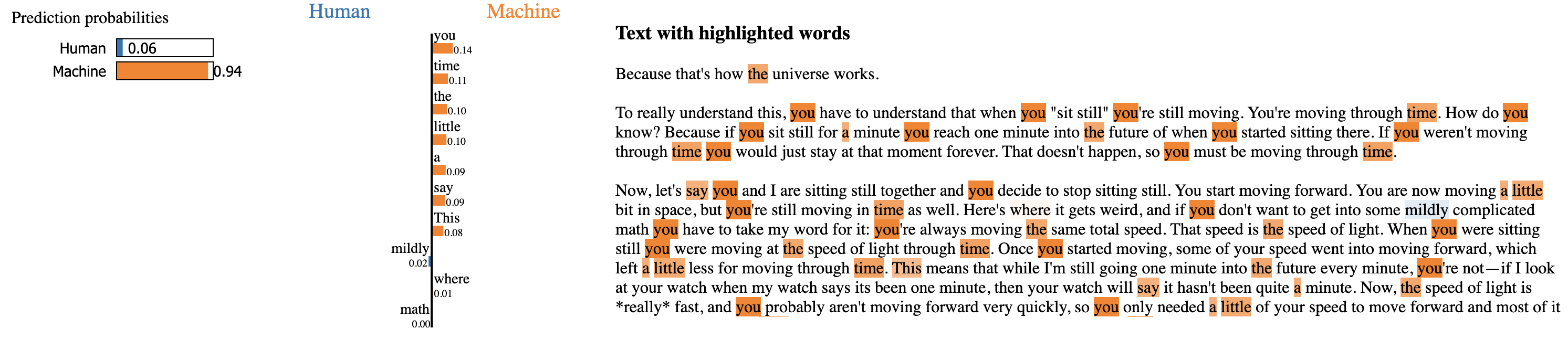}  
  \caption{FP example.}
\end{subfigure}
\begin{subfigure}{1\textwidth}
  \centering
  \includegraphics[width=\linewidth]{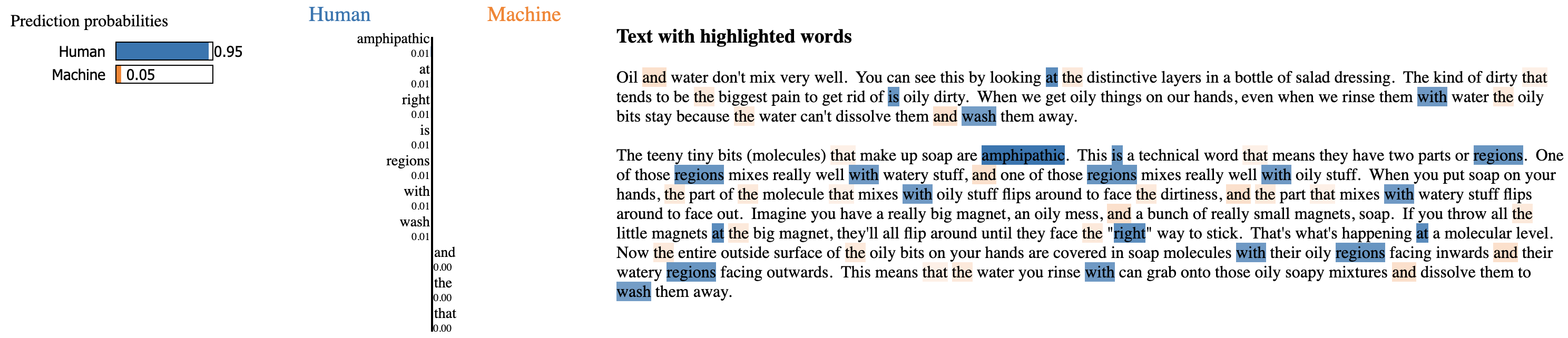}  
  \caption{TN example.}
 
\end{subfigure}
\begin{subfigure}{1\textwidth}
  \centering
  \includegraphics[width=\linewidth]{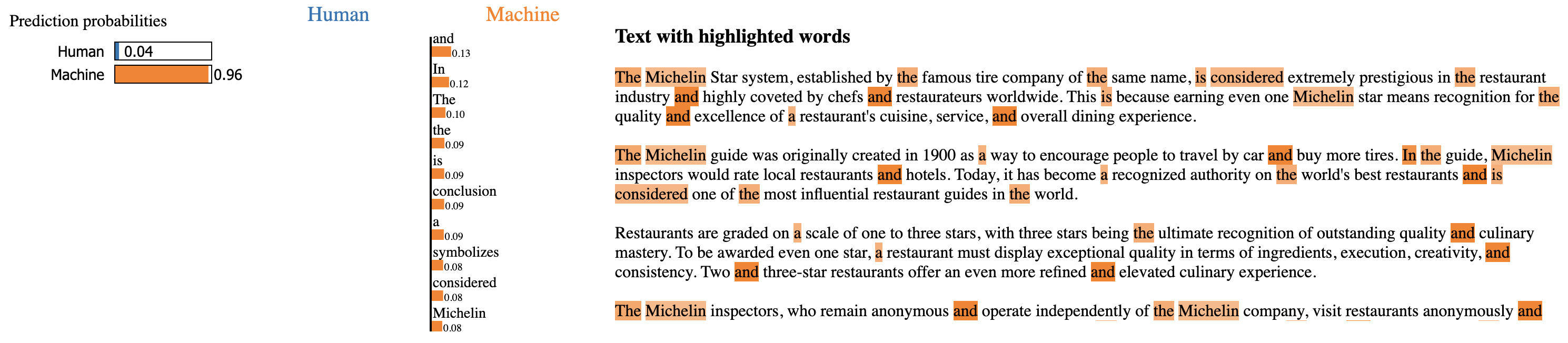}  
  \caption{TP example.}
 
\end{subfigure}

\caption{Visualization of the features extracted by LIME for Reddit as a domain, and ChatGPT as a generator.}
\label{img:lime}
\end{figure}

Figure~\ref{img:lime} shows a the features extracted by LIME for Reddit as a domain and ChatGPT as a generator. Shown are a false positive, a true negative, and a true positive example.

\clearpage
\section{Examples of M4}
\label{sec:examples}
\input{section/tables/english_examples}
\input{section/tables/bloomz_examples}
\input{section/tables/other_language_examples}

%% file: section/tables/hyperparameters_table.tex
\subsection{Detector Hyper-Parameter Values}

Table~\ref{tab:detector-parameters} shows the non-default values of the hyper-parameters for the five detectors we trained.   
    
\begin{table}[ht]
    \centering
    \small
    \begin{tabular}{ l  c  c  c  c  c }
    \toprule
     \textbf{Detector$\downarrow$} & \textbf{Learning rate} & \textbf{\# epochs} & \textbf{Batch size} & \textbf{\multirowcellii{ Maximum \\ iterations} }  & \textbf{C} \\
    \midrule 
        RoBERTa-\textit{base} & 1e-6 & 10 & 64 & -- & -- \\
        ELECTRA-\textit{base} & 1e-6 & 10 & 64 & -- & -- \\
        XLM-R-\textit{base} & 2e-5 & 5 & 16 & -- & --\\
        LR-GLTR & default & -- & default & 1,000 & --\\
        Linear-SVM & -- & -- & -- & 20,000 & 0.8\\
    \bottomrule
    \end{tabular} 
    \caption{Hyper-parameter settings for the five detectors we trained. LR-GLTR is based on the \textit{sklearn} logistic regression implementation, and all hyper-parameters follow the default setting except for maximum training iterations, which we set to 1,000. The Linear-SVM detector uses all default parameters provided in the \textit{sklearn} implementation except for the penalty parameter of the error term C and the maximum number of iterations. }
    \label{tab:detector-parameters}
\end{table}

%% file: section/tables/generator_eval_tables.tex
Table~\ref{tab:chatgpt} shows the evaluation results for same-generator, cross-domain experiments: training on a single domain of ChatGPT vs. human, and testing across domains. Table~\ref{tab:davinci} shows the corresponding results for davinci-003 vs. human.

\begin{table*}[ht]
    \centering
    \small
    \resizebox{\textwidth}{!}{
    \addtolength{\tabcolsep}{-2.5pt}
    \begin{tabular}{l | llll | llll | llll | llll | llll}
    \toprule
        Test $\rightarrow$ & \multicolumn{4}{c|}{Wikipedia} & \multicolumn{4}{c|}{WikiHow} & \multicolumn{4}{c|}{Reddit ELI5} & \multicolumn{4}{c|}{arXiv} & \multicolumn{4}{c}{PeerRead}  \\
        Train $\downarrow$ & Acc & Prec & Recall & F1 & Acc & Prec & Recall & F1 & Acc & Prec & Recall & F1 & Acc & Prec & Recall & F1 & Acc & Prec & Recall & F1 \\
    \midrule
    \multicolumn{21}{c}{\textbf{RoBERTa(base)}} \\
    \midrule
        Wikipedia & \textbf{99.7} & \textbf{99.4} & \textbf{100.} & \textbf{99.7} & 48.2 & 5.0 & 0.2 & 0.4 & 48.7 & 6.7 & 0.2 & 0.4 & 55.6 & 98.3 & 11.4 & 20.4 & 60.7 & 0.0 & 0.0 & 0.0 \\
        WikiHow & 18.3 & 9.9 & 7.8 & 8.7 & \textbf{99.7} & \textbf{99.8} & \textbf{99.6} & \textbf{99.7} & 89.3 & 87.3 & 92.0 & 89.6 & 96.9 & 94.2 & \textbf{100.} & 97.0 & 84.4 & 61.3 & 96.7 & 75.0 \\
        Reddit ELI5 & 79.1 & 70.7 & 99.4 & 82.6 & 82.4 & 80.2 & 86.0 & 83.0 & 89.7 & 82.9 & \textbf{100.} & 90.7 & 99.5 & 99.8 & 99.2 & 99.5 & 80.6 & 55.7 & 96.7 & 70.7 \\
        arXiv & 91.5 & 85.7 & 99.6 & 92.1 & 75.7 & 96.7 & 53.2 & 68.6 & \textbf{95.9} & 97.7 & 94.0 & \textbf{95.8} & \textbf{100.} & \textbf{100.} & \textbf{100.} & \textbf{100.} & 52.4 & 33.8 & \textbf{100.} & 50.5 \\
        PeerRead & 58.2 & 64.6 & 36.2 & 46.4 & 66.0 & \textbf{98.8} & 32.4 & 48.8 & 75.1 & \textbf{100.} & 50.2 & 66.8 & 99.0 & 100. & 98.0 & 99.0 & \textbf{98.0} & \textbf{92.5} & \textbf{100.} & \textbf{96.1} \\
    \midrule
     \multicolumn{21}{c}{\textbf{ELECTRA(large)}} \\
    \midrule
        Wikipedia & \textbf{94.0} & 89.9 & 99.2 & \textbf{94.3} & 90.4 & 88.8 & 92.4 & 90.6 & \textbf{96.6} & 96.6 & 96.6 & \textbf{96.6} & 97.8 & \textbf{99.8} & 95.8 & 97.8 & 94.6 & 98.7 & 78.8 & 87.6 \\
        WikiHow & 69.0 & 63.5 & 89.6 & 74.3 & \textbf{99.5} & 99.4 & \textbf{99.6} & \textbf{99.5} & 68.9 & 98.0 & 38.6 & 55.4 & 84.0 & 76.0 & 99.4 & 86.1 & 90.8 & 94.2 & 66.2 & 77.7 \\
        Reddit ELI5 & 68.1 & 61.1 & \textbf{99.8} & 75.8 & 68.9 & 61.7 & 99.4 & 76.2 & 95.3 & 91.4 & \textbf{100.} & 95.5 & 92.9 & 87.7 & 99.8 & 93.4 & 93.3 & 78.3 & \textbf{100.} & 87.8 \\
        arXiv & 90.9 & 96.2 & 85.2 & 90.3 & 65.1 & \textbf{100.} & 30.2 & 46.4 & 76.7 & \textbf{100.} & 53.4 & 69.6 & \textbf{98.2} & 96.5 & \textbf{100.} & \textbf{98.2} & \textbf{98.7} & \textbf{99.5} & 95.2 & \textbf{97.3} \\
        PeerRead & 81.3 & \textbf{98.2} & 63.8 & 77.3 & 71.4 & 98.6 & 43.4 & 60.3 & 75.6 & \textbf{100.} & 51.2 & 67.7 & 97.8 & 97.2 & 98.4 & 97.8 & 97.9 & 92.1 & \textbf{100.} & 95.9 \\
    \midrule
    \multicolumn{21}{c}{\textbf{LR-GLTR}} \\
    \midrule
        Wikipedia & \textbf{97.4} & \textbf{97.6} & 97.2 & \textbf{97.4} & 78.5 & 87.8 & 66.2 & 75.5 & 86.2 & 78.5 & \textbf{99.8} & 87.9 & 94.4 & 98.3 & 90.4 & 94.2 & 70.9 & 67.2 & 81.6 & 73.7 \\
        WikiHow & 91.3 & 87.3 & 96.6 & 91.7 & \textbf{92.4} & \textbf{92.1} & 92.8 & \textbf{92.4} & 93.8 & \textbf{96.6} & 90.8 & 93.6 & 90.4 & 99.8 & 81.0 & 89.4 & 84.1 & 87.5 & 79.6 & 83.4 \\
        Reddit ELI5 & 96.0 & 94.9 & 97.2 & 96.0 & 90.0 & 90.3 & 89.6 & 90.0 & \textbf{95.4} & 92.7 & 98.6 & \textbf{95.5} & 91.7 & \textbf{100.} & 83.4 & 90.9 & 78.9 & 79.2 & 78.4 & 78.8 \\
        arXiv & 92.5 & 87.3 & \textbf{99.4} & 93.0 & 87.3 & 82.5 & 94.6 & 88.2 & 84.8 & 76.8 & \textbf{99.8} & 86.8 & \textbf{96.3} & 96.4 & \textbf{96.2} & \textbf{96.3} & 77.0 & 70.1 & \textbf{94.2} & 80.4 \\
        PeerRead & 88.9 & 82.1 & \textbf{99.4} & 90.0 & 71.2 & 63.9 & \textbf{97.6} & 77.2 & 84.5 & 76.7 & 99.2 & 86.5 & 89.4 & 98.8 & 79.8 & 88.3 & \textbf{94.2} & \textbf{99.1} & 89.2 & \textbf{93.9} \\
    \midrule
    \multicolumn{21}{c}{\textbf{Stylistic}} \\
    \midrule
        Wikipedia & \textbf{97.4} & \textbf{97.6} & \textbf{97.2} & \textbf{97.4} & 56.2 & 73.8 & 19.2 & 30.5 & 74.7 & 78.4 & 68.2 & 72.9 & 96.8 & 97.0 & 96.6 & 96.8 & 86.5 & 87.5 & 85.2 & 86.3 \\
        WikiHow & 59.0 & 56.6 & 77.6 & 65.4 & \textbf{95.7} & \textbf{97.7} & \textbf{93.6} & \textbf{95.6} & 59.3 & 61.2 & 50.8 & 55.5 & 46.6 & 47.4 & 62.8 & 54.0 & 61.9 & 62.8 & 58.4 & 60.5 \\
        Reddit ELI5 & 88.9 & 91.2 & 86.1 & 88.6 & 49.7 & 48.3 & 8.4 & 14.3 & \textbf{92.3} & \textbf{89.2} & \textbf{96.2} & \textbf{92.6} & 89.3 & 97.3 & 80.8 & 88.3 & 80.7 & 86.3 & 73.0 & 79.1 \\
        arXiv & 73.7 & 68.1 & 89.3 & 77.3 & 55.0 & 62.4 & 25.2 & 35.9 & 70.6 & 82.4 & 52.4 & 64.1 & \textbf{100.} & \textbf{100.} & \textbf{100.} & \textbf{100.} & 87.6 & 84.0 & 93.0 & 88.3 \\
        PeerRead & 64.2 & 67.1 & 56.0 & 61.0 & 51.2 & 77.3 & 3.4 & 6.5 & 59.3 & 92.7 & 20.2 & 33.2 & 77.6 & 96.3 & 57.4 & 71.9 & \textbf{99.6} & \textbf{100.} & \textbf{99.1} & \textbf{99.6} \\
    \midrule
    \multicolumn{21}{c}{\textbf{NELA}} \\
    \midrule
        Wikipedia & \textbf{95.6} & \textbf{96.7} & \textbf{94.3} & \textbf{95.5} & 76.9 & 73.1 & 85.2 & 78.7 & 76.0 & 70.9 & 88.2 & 78.6 & 77.1 & 69.1 & \textbf{98.2} & 81.1 & 73.7 & 66.4 & 95.9 & 78.5 \\
        WikiHow & 65.4 & 61.1 & 84.4 & 70.9 & \textbf{95.6} & \textbf{96.0} & \textbf{95.2} & \textbf{95.6} & 69.0 & 92.8 & 41.2 & 57.1 & 78.6 & 85.0 & 69.4 & 76.4 & 88.5 & 96.2 & 80.2 & 87.5 \\
        Reddit & 87.5 & 88.7 & 85.9 & 87.3 & 54.5 & 73.7 & 14.0 & 23.5 & \textbf{93.1} & 90.1 & \textbf{96.8} & \textbf{93.3} & 78.3 & 70.2 & \textbf{98.2} & 81.9 & 90.6 & 84.3 & \textbf{99.7} & 91.3 \\
        arXiv & 73.9 & 75.5 & 70.9 & 73.1 & 63.7 & 62.7 & 67.8 & 65.1 & 69.2 & 86.6 & 45.4 & 59.6 & \textbf{97.2} & \textbf{97.0} & 97.4 & \textbf{97.2} & 84.7 & 92.2 & 75.9 & 83.3 \\
        PeerRead & 60.5 & 63.5 & 49.3 & 55.5 & 53.5 & 83.0 & 8.8 & 15.9 & 58.5 & \textbf{100.} & 17.0 & 29.1 & 84.0 & 88.1 & 78.6 & 83.1 & \textbf{98.4} & \textbf{99.4} & 97.4 & \textbf{98.4}\\
    \bottomrule
    \end{tabular} }
    \caption{\textbf{Same-generator, cross-domain experiments: training on a single domain of ChatGPT vs. human, and testing across domains.} Evaluation accuracy (Acc), precision (Prec), recall, and F1 score (in \%) with respect to machine generations across four detectors.}
    \label{tab:chatgpt}
\end{table*}

\begin{table*}[t!]
    \centering
    \small
    \resizebox{\textwidth}{!}{
\addtolength{\tabcolsep}{-2.5pt}
    \begin{tabular}{l | llll | llll | llll | llll | llll}
    \toprule
        Test $\rightarrow$ & \multicolumn{4}{c|}{Wikipedia} & \multicolumn{4}{c|}{WikiHow} & \multicolumn{4}{c|}{Reddit ELI5} & \multicolumn{4}{c|}{arXiv} & \multicolumn{4}{c}{PeerRead}  \\
        Train $\downarrow$ & Acc & Prec & Recall & F1 & Acc & Prec & Recall & F1 & Acc & Prec & Recall & F1 & Acc & Prec & Recall & F1 & Acc & Prec & Recall & F1 \\
    \midrule
    \multicolumn{21}{c}{\textbf{RoBERTa(base)}} \\
    \midrule
        Wikipedia & \textbf{99.6} & \textbf{99.4} & 99.8 & \textbf{99.6} & 47.8 & 17.6 & 1.2 & 2.2 & 49.0 & 8.3 & 0.2 & 0.4 & 74.8 & 92.5 & 54.0 & 68.2 & 56.7 & 0.0 & 0.0 & 0.0 \\
        WikiHow & 46.4 & 48.0 & 87.4 & 62.0 & \textbf{99.4} & 99.0 & \textbf{99.8} & \textbf{99.4} & 58.6 & 54.7 & 99.8 & 70.7 & 95.0 & 95.0 & 95.0 & 95.0 & 31.7 & 26.2 & \textbf{100.} & 41.6 \\
        Reddit ELI5 & 42.8 & 42.4 & 40.2 & 41.3 & 88.1 & 87.9 & 88.4 & 88.1 & \textbf{93.6} & 88.7 & \textbf{100.} & \textbf{94.0} & 52.4 & \textbf{100.} & 4.8 & 9.2 & 91.2 & 74.9 & 96.2 & 84.2 \\
        arXiv & 55.5 & 52.9 & \textbf{100.} & 69.2 & 55.3 & 52.9 & 96.8 & 68.4 & 54.4 & 52.3 & 99.8 & 68.6 & \textbf{99.4} & 99.8 & \textbf{99.0} & \textbf{99.4} & 26.3 & 24.8 & \textbf{100.} & 39.7 \\
        PeerRead & 51.6 & 94.4 & 3.4 & 6.6 & 50.2 & \textbf{100.} & 0.4 & 0.8 & 51.9 & \textbf{100.} & 3.8 & 7.3 & 53.3 & \textbf{100.} & 6.6 & 12.4 & \textbf{98.7} & \textbf{94.7} & \textbf{100.} & \textbf{97.3}\\
    \midrule
    \multicolumn{21}{c}{\textbf{ELECTRA(large)}} \\
    \midrule
        Wikipedia & \textbf{83.5} & 75.7 & 98.8 & \textbf{85.7} & 76.3 & 83.1 & 66.0 & 73.6 & 87.9 & 81.4 & 98.2 & 89.0 & 62.2 & 70.9 & 41.4 & 52.3 & 84.4 & 61.7 & 94.2 & 74.5 \\
        WikiHow & 60.3 & 56.0 & 96.6 & 70.9 & \textbf{99.0} & \textbf{98.4} & \textbf{99.6} & \textbf{99.0} & 81.0 & 87.1 & 72.8 & 79.3 & 56.8 & 53.9 & 93.8 & 68.5 & 70.2 & 44.5 & 91.2 & 59.8 \\
        Reddit ELI5 & 68.2 & 61.2 & \textbf{99.6} & 75.8 & 66.2 & 60.1 & 96.6 & 74.1 & \textbf{95.2} & 91.2 & \textbf{100.} & \textbf{95.4} & 61.2 & 76.2 & 32.6 & 45.7 & 97.5 & 91.6 & 99.0 & 95.1 \\
        arXiv & 50.4 & 50.4 & 57.4 & 53.6 & 49.2 & 42.6 & 4.6 & 8.3 & 52.7 & 65.9 & 11.2 & 19.1 & \textbf{94.6} & 93.9 & \textbf{95.4} & \textbf{94.6} & 58.3 & 27.7 & 44.4 & 34.1 \\
        PeerRead & 51.1 & \textbf{100.} & 2.2 & 4.3 & 50.0 & 50.0 & 0.2 & 0.4 & 50.9 & \textbf{100.} & 1.8 & 3.5 & 51.9 & \textbf{95.2} & 4.0 & 7.7 & \textbf{99.0} & \textbf{96.1} & \textbf{100.} & \textbf{98.0}\\
    \midrule
    \multicolumn{21}{c}{\textbf{LR-GLTR}} \\
    \midrule
        Wikipedia & \textbf{90.3} & \textbf{89.3} & 91.6 & \textbf{90.4} & 73.5 & 68.3 & \textbf{87.6} & 76.8 & 68.2 & 61.3 & \textbf{99.0} & 75.7 & 71.5 & 85.2 & 52.0 & 64.6 & 72.7 & 64.7 & 99.8 & 78.5 \\
        WikiHow & 88.2 & 83.9 & \textbf{94.6} & 88.9 & \textbf{79.6} & \textbf{77.4} & 83.6 & \textbf{80.4} & 77.7 & 69.5 & 98.8 & 81.6 & 72.6 & 84.9 & 55.0 & 66.7 & 76.0 & 67.6 & \textbf{100.} & 80.6 \\
        Reddit ELI5 & 86.7 & 83.5 & 91.4 & 87.3 & 76.0 & 72.7 & 83.2 & 77.6 & \textbf{88.5} & 82.9 & 97.0 & \textbf{89.4} & 53.4 & \textbf{90.5} & 7.6 & 14.0 & 90.2 & 84.4 & 98.6 & 91.0 \\
        arXiv & 47.1 & 6.1 & 0.4 & 0.8 & 50.2 & 52.9 & 3.6 & 6.7 & 45.1 & 34.4 & 10.8 & 16.4 & \textbf{85.2} & 84.5 & \textbf{86.2} & \textbf{85.3} & 71.2 & 63.9 & 97.2 & 77.1 \\
        PeerRead & 84.5 & 83.2 & 86.4 & 84.8 & 73.5 & 73.0 & 74.6 & 73.8 & 86.3 & \textbf{85.8} & 87.0 & 86.4 & 50.2 & 62.5 & 1.0 & 2.0 & \textbf{94.6} & \textbf{99.6} & 89.6 & \textbf{94.3} \\
    \midrule
    \multicolumn{21}{c}{\textbf{Stylistic}} \\
    \midrule
        Wikipedia & \textbf{96.5} & \textbf{96.2} & \textbf{96.8} & \textbf{96.5} & 66.6 & 69.5 & 59.2 & 63.9 & 67.0 & 68.0 & 64.2 & 66.0 & 76.7 & 91.8 & 58.6 & 71.6 & 79.5 & 76.6 & 84.9 & 80.6 \\ 
        WikiHow & 63.3 & 58.3 & 93.0 & 71.7 & \textbf{93.9} & \textbf{94.5} & \textbf{93.2} & \textbf{93.9} & 65.4 & 62.5 & 77.2 & 69.1 & 57.8 & 54.9 & 87.4 & 67.4 & 73.0 & 65.1 & 98.8 & 78.5 \\ 
        Reddit ELI5 & 80.6 & 83.6 & 76.2 & 79.7 & 64.0 & 71.6 & 46.4 & 56.3 & \textbf{92.0} & \textbf{88.6} & \textbf{96.4} & \textbf{92.3} & 56.1 & 67.0 & 24.0 & 35.3 & 77.9 & 80.2 & 74.1 & 77.0 \\
        arXiv & 63.5 & 81.1 & 35.2 & 49.1 & 49.1 & 46.7 & 12.8 & 20.1 & 59.8 & 63.4 & 46.4 & 53.6 & \textbf{97.4} & \textbf{97.2} & \textbf{97.6} & \textbf{97.4} & 89.7 & 83.5 & 98.8 & 90.5 \\ 
        PeerRead & 60.7 & 63.6 & 50.0 & 56.0 & 49.4 & 41.7 & 3.0 & 5.6 & 55.0 & 70.8 & 17.0 & 27.4 & 66.3 & 76.7 & 46.8 & 58.1 & \textbf{99.3} & \textbf{99.1} & \textbf{99.4} & \textbf{99.3} \\
    \midrule
    \multicolumn{21}{c}{\textbf{NELA}} \\
    \midrule
        Wikipedia & \textbf{92.5} & \textbf{93.1} & \textbf{91.8} & \textbf{92.4} & 70.1 & 63.8 & \textbf{92.8} & 75.6 & 72.0 & 66.4 & 89.2 & 76.1 & 47.2 & 46.8 & 41.6 & 44.1 & 60.0 & 58.0 & 72.7 & 64.5 \\ 
        WikiHow & 68.2 & 64.1 & 82.8 & 72.3 & \textbf{89.5} & \textbf{90.2} & 88.6 & \textbf{89.4} & 81.1 & 86.9 & 73.2 & 79.5 & 50.8 & 50.9 & 44.2 & 47.3 & 82.6 & 78.0 & 90.7 & 83.9 \\ 
        Reddit ELI5 & 80.0 & 83.5 & 74.8 & 78.9 & 70.6 & 89.9 & 46.4 & 61.2 & \textbf{93.2} & 91.1 & \textbf{95.8} & \textbf{93.4} & 42.5 & 38.7 & 25.6 & 30.8 & 86.3 & 83.6 & 90.4 & 86.9 \\
        arXiv & 48.5 & 5.9 & 0.2 & 0.4 & 51.0 & 69.2 & 3.6 & 6.8 & 45.9 & 4.4 & 0.4 & 0.7 & \textbf{88.5} & \textbf{88.9} & \textbf{88.0} & \textbf{88.4} & 76.3 & 88.2 & 60.8 & 71.9 \\ 
        PeerRead & 48.0 & 29.2 & 2.8 & 5.1 & 50.3 & 60.0 & 1.8 & 3.5 & 52.0 & \textbf{95.5} & 4.2 & 8.0 & 56.2 & 64.5 & 27.6 & 38.7 & \textbf{97.8} & \textbf{99.7} & \textbf{95.9} & \textbf{97.8} \\

    \bottomrule
    \end{tabular} }
    \caption{\textbf{Same-generator, cross-domain experiments: training on a single domain of davinci-003 vs. human, and testing across domains.} Evaluation accuracy (Acc), precision (Prec), recall, and F1 scores (in \%) with respect to machine generations across four detectors.}
    \label{tab:davinci}
\end{table*}

%% file: section/tables/domain_eval_tables.tex
Table~\ref{tab:arxiv-1} shows detailed evaluation results for same-domain, cross-generator experiments: training and testing on arXiv (single machine-text generator vs. human). Table~\ref{tab:wikipedia-1} shows the corresponding results when training and testing on Wikipedia.

\begin{table*}[ht]
    \centering
    \small
    \resizebox{\textwidth}{!}{
        \addtolength{\tabcolsep}{-1pt}
    \begin{tabular}{l | llll | llll | llll | llll}
    \toprule
        Test $\rightarrow$ & \multicolumn{4}{c|}{ChatGPT} & \multicolumn{4}{c|}{davinci} & \multicolumn{4}{c|}{Cohere} & \multicolumn{4}{c}{BLOOMz} \\
        Train $\downarrow$ & Acc & Prec & Recall & F1 & Acc & Prec & Recall & F1 & Acc & Prec & Recall & F1 & Acc & Prec & Recall & F1 \\
    \midrule
    \multicolumn{17}{c}{\textbf{RoBERTa(base)}} \\
    \midrule
        ChatGPT & \textbf{99.7} & \textbf{99.4} & \textbf{100.} & \textbf{99.7} & \textbf{99.7} & 99.4 & \textbf{100.} & \textbf{99.7} & 99.4 & \textbf{99.8} & 99.0 & 99.4 & 77.7 & \textbf{100.} & 55.4 & 71.3 \\ 
        davinci & 99.6 & 99.2 & \textbf{100.} & 99.6 & 99.5 & 99.2 & 99.8 & 99.5 & 99.4 & \textbf{99.8} & 99.0 & 99.4 & 81.4 & 99.7 & 63.0 & 77.2 \\ 
        Cohere & \textbf{99.7} & 99.4 & \textbf{100.} & \textbf{99.7} & 99.6 & 99.4 & 99.8 & 99.6 & \textbf{99.6} & \textbf{99.8} & \textbf{99.4} & \textbf{99.6} & 82.6 & 99.7 & 65.4 & 79.0 \\ 
        BLOOMz & 99.3 & 98.8 & 99.8 & 99.3 & 99.3 & \textbf{99.8} & \textbf{99.8} & 99.3 & 99.0 & 98.8 & 99.2 & 99.0 & \textbf{98.1} & 98.8 & \textbf{97.4} & \textbf{98.1} \\

    \midrule
      \multicolumn{17}{c}{\textbf{ELECTRA(large)}} \\
    \midrule
        ChatGPT & \textbf{93.5} & \textbf{88.9} & \textbf{99.4} & \textbf{93.9} & \textbf{91.8} & \textbf{88.6} & 96.0 & \textbf{92.1} & 83.6 & \textbf{86.4} & 79.8 & 83.0 & 56.6 & 72.9 & 21.0 & 32.6 \\ 
        davinci & 88.6 & 81.8 & 99.2 & 89.7 & 88.7 & 81.9 & \textbf{99.4} & 89.8 & 83.8 & 79.2 & 91.6 & 85.0 & 62.4 & 73.1 & 39.2 & 51.0 \\ 
        Cohere & 84.3 & 76.5 & 99.0 & 86.3 & 83.4 & 76.2 & 97.2 & 85.4 & \textbf{85.5} & 77.8 & \textbf{99.4} & \textbf{87.3} & 72.4 & 72.7 & 71.8 & 72.2 \\ 
        BLOOMz & 49.9 & 48.1 & 2.6 & 4.9 & 50.1 & 51.7 & 3.0 & 5.7 & 53.2 & 77.6 & 9.0 & 16.1 & \textbf{97.5} & \textbf{98.0} & \textbf{97.0} & \textbf{97.5} \\

    \midrule
    \multicolumn{17}{c}{\textbf{LR-GLTR}} \\
    \midrule
        ChatGPT & 96.3 & \textbf{96.4} & 96.2 & 96.3 & 65.3 & 90.1 & 34.4 & 49.8 & 96.9 & \textbf{96.4} & 97.4 & 96.9 & 65.5 & 90.6 & 34.6 & 50.1 \\
        davinci & 81.2 & 83.9 & 77.2 & 80.4 & \textbf{85.2} & 84.5 & \textbf{86.2} & \textbf{85.3} & 78.5 & 82.9 & 71.8 & 77.0 & 73.7 & 80.8 & 62.2 & 70.3 \\
        Cohere & \textbf{96.8} & \textbf{96.4} & \textbf{97.2} & \textbf{96.8} & 66.0 & \textbf{90.4} & 35.8 & 51.3 & \textbf{97.0} & \textbf{96.4} & \textbf{97.6} & \textbf{97.0} & 61.5 & \textbf{88.1} & 26.6 & 40.9 \\
        BLOOMz & 89.2 & 87.7 & 91.2 & 89.4 & 71.2 & 80.8 & 55.6 & 65.9 & 79.5 & 84.9 & 71.8 & 77.8 & \textbf{87.2} & 87.2 & \textbf{87.2} & \textbf{87.2} \\

    \midrule
    \multicolumn{17}{c}{\textbf{Stylistic}} \\
    \midrule
        ChatGPT & \textbf{100.} & \textbf{100.} & \textbf{100.} & \textbf{100.} & 71.0 & \textbf{100.} & 42.0 & 59.2 & 87.7 & \textbf{100.} & 75.4 & 86.0 & 62.4 & \textbf{100.} & 24.8 & 39.7 \\
        davinci & 97.3 & 97.4 & 97.2 & 97.3 & \textbf{97.4} & 97.2 & \textbf{97.6} & \textbf{97.4} & 82.8 & 96.3 & 68.2 & 79.9 & 87.1 & 96.7 & 76.8 & 85.6 \\
        Cohere & 97.6 & 99.4 & 95.8 & 97.6 & 83.8 & 99.7 & 67.8 & 80.7 & \textbf{98.8} & 99.4 & \textbf{98.2} & \textbf{98.8} & 65.5 & 98.1 & 31.6 & 47.8 \\
        BLOOMz & 63.4 & 95.3 & 28.2 & 43.5 & 76.0 & 97.4 & 53.4 & 69.0 & 55.5 & 89.9 & 12.4 & 21.8 & \textbf{98.5} & \textbf{98.6} & \textbf{98.4} & \textbf{98.5} \\
    \midrule
    \multicolumn{17}{c}{\textbf{NELA}} \\
    \midrule
        ChatGPT & \textbf{97.2} & \textbf{97.0} & \textbf{97.4} & \textbf{97.2} & 52.0 & 69.2 & 7.2 & 13.0 & 64.2 & 91.3 & 31.4 & 46.7 & 48.8 & 16.7 & 0.6 & 1.2 \\
        davinci & 48.3 & 41.2 & 8.0 & 13.4 & \textbf{88.5} & \textbf{88.9} & \textbf{88.0} & \textbf{88.4} & 45.8 & 20.8 & 3.0 & 5.2 & 73.0 & 83.4 & 57.4 & 68.0 \\
        Cohere & 70.1 & 88.8 & 46.0 & 60.6 & 49.4 & 44.6 & 5.0 & 9.0 & \textbf{93.9} & \textbf{94.2} & \textbf{93.6} & \textbf{93.9} & 47.1 & 20.8 & 7.3 & 8.9 \\
        BLOOMz & 48.6 & 11.1 & 0.4 & 0.8 & 55.5 & 81.6 & 14.2 & 24.2 & 48.7 & 15.8 & 0.6 & 1.2 & \textbf{96.9} & \textbf{96.8} & \textbf{97.0} & \textbf{96.9} \\
    \bottomrule
    \end{tabular} }
    \caption{\textbf{Same-domain, cross-generator experiments: training and testing on arXiv (single machine-text generator vs. human).} Evaluation accuracy (Acc), precision (Prec), recall, and F1 score (in \%) with respect to the machine generations across four detectors.}
    \label{tab:arxiv-1}
\end{table*}

\begin{table*}[t!]
    \centering
    \small
    \resizebox{\textwidth}{!}{
            \addtolength{\tabcolsep}{-1pt}
    \begin{tabular}{l | llll | llll | llll | llll}
    \toprule
        Test $\rightarrow$ & \multicolumn{4}{c|}{ChatGPT} & \multicolumn{4}{c|}{davinci} & \multicolumn{4}{c|}{Cohere} & \multicolumn{4}{c}{BLOOMz} \\
        Train $\downarrow$ & Acc & Prec & Recall & F1 & Acc & Prec & Recall & F1 & Acc & Prec & Recall & F1 & Acc & Prec & Recall & F1 \\
    \midrule
    \multicolumn{17}{c}{\textbf{RoBERTa(base)}} \\
    \midrule
        ChatGPT & \textbf{100.} & \textbf{100.} & \textbf{100.} & \textbf{100.} & 97.4 & \textbf{100.} & 94.8 & 97.3 & 99.0 & \textbf{100.} & 98.0 & 99.0 & 99.5 & \textbf{100.} & 99.0 & 99.5 \\
        davinci & 99.9 & 99.8 & \textbf{100.} & 99.9 & \textbf{99.3} & 99.8 & \textbf{98.8} & \textbf{99.3} & 99.7 & 99.8 & \textbf{99.6} & 99.7 & 99.9 & 99.8 & \textbf{100.} & 99.9 \\
        Cohere & \textbf{100.} & \textbf{100.} & \textbf{100.} & \textbf{100.} & 97.7 & \textbf{100.} & 95.4 & 97.6 & \textbf{99.8} & \textbf{100.} & 99.6 & \textbf{99.8} & \textbf{100.} & \textbf{100.} & \textbf{100.} & \textbf{100.} \\
        BLOOMz & \textbf{100.} & \textbf{100.} & \textbf{100.} & \textbf{100.} & 97.7 & \textbf{100.} & 95.4 & 97.6 & 99.8 & \textbf{100.} & 99.6 & 99.8 & \textbf{100.} & \textbf{100.} & \textbf{100.} & \textbf{100.} \\
    \midrule
\multicolumn{17}{c}{\textbf{ELECTRA(large)}} \\
\midrule
        ChatGPT & \textbf{98.3} & 96.7 & \textbf{100.} & \textbf{98.3} & 59.2 & 85.9 & 22.0 & 35.0 & 59.3 & 86.6 & 22.0 & 35.1 & 68.5 & 92.2 & 40.4 & 56.2 \\
        davinci & 70.3 & 89.8 & 45.8 & 60.7 & \textbf{95.6} & \textbf{94.9} & \textbf{96.4} & \textbf{95.6} & 49.4 & 43.5 & 4.0 & 7.3 & 59.5 & 82.3 & 24.2 & 37.4 \\
        Cohere & 92.7 & 91.1 & 94.6 & 92.8 & 51.7 & 57.3 & 13.4 & 21.7 & \textbf{95.2} & \textbf{91.5} & \textbf{99.6} & \textbf{95.4} & 65.5 & 81.4 & 40.2 & 53.8 \\
        BLOOMz & 81.3 & \textbf{96.7} & 64.8 & 77.6 & 61.3 & 92.5 & 24.6 & 38.9 & 53.8 & 81.7 & 9.8 & 17.5 & \textbf{98.7} & \textbf{97.8} & \textbf{99.6} & \textbf{98.7} \\
    \midrule
    
    \multicolumn{17}{c}{\textbf{LR-GLTR}} \\
    \midrule
        ChatGPT & \textbf{97.4} & \textbf{97.6} & 97.2 & \textbf{97.4} & 85.0 & \textbf{96.8} & 72.4 & 82.8 & 92.9 & \textbf{97.8} & 87.8 & 92.5 & 81.3 & 75.7 & 92.2 & 83.1 \\
        davinci & 94.1 & 90.0 & \textbf{99.2} & 94.4 & \textbf{90.3} & 89.3 & \textbf{91.6} & \textbf{90.4} & 90.5 & 89.5 & 91.8 & 90.6 & 77.5 & 70.0 & \textbf{96.2} & 81.0 \\
        Cohere & 96.5 & 95.1 & 98.0 & 96.6 & 85.0 & 93.8 & 75.0 & 83.3 & \textbf{95.1} & 95.2 & \textbf{95.0} & \textbf{95.1} & 75.6 & 71.3 & 85.6 & 77.8 \\
        BLOOMz & 69.9 & 95.0 & 42.0 & 58.3 & 66.4 & 94.1 & 35.0 & 51.0 & 55.0 & 87.9 & 11.6 & 20.5 & \textbf{91.0} & \textbf{89.4} & 93.0 & \textbf{91.2} \\
    \midrule
    \multicolumn{17}{c}{\textbf{Stylistic}} \\
    \midrule
        ChatGPT & \textbf{97.4} & \textbf{97.6} & \textbf{97.2} & \textbf{97.4} & 93.3 & \textbf{97.4} & 89.0 & 93.0 & 87.6 & \textbf{97.7} & 77.1 & 86.2 & 63.7 & 73.2 & 43.4 & 54.5 \\
        davinci & 96.7 & 96.2 & \textbf{97.2} & 96.7 & \textbf{96.5} & 96.2 & \textbf{96.8} & \textbf{96.5} & 90.5 & 96.6 & 83.9 & 89.8 & 67.0 & 78.2 & 47.2 & 58.8 \\
        Cohere & 90.4 & 95.7 & 84.6 & 89.8 & 82.2 & 94.7 & 68.2 & 79.3 & \textbf{94.2} & 93.5 & \textbf{94.9} & \textbf{94.2} & 69.8 & 73.4 & 62.3 & 67.4 \\
        BLOOMz & 53.7 & 84.9 & 9.1 & 16.4 & 53.7 & 84.9 & 9.0 & 16.3 & 54.0 & 84.6 & 9.8 & 17.6 & \textbf{95.2} & \textbf{94.0} & \textbf{96.6} & \textbf{95.3} \\
    \midrule
    \multicolumn{17}{c}{\textbf{NELA}} \\
    \midrule
        ChatGPT & \textbf{95.6} & \textbf{96.7} & 94.3 & \textbf{95.5} & 91.0 & \textbf{96.2} & 85.4 & 90.5 & 78.1 & 94.8 & 59.5 & 73.1 & 50.2 & 53.7 & 3.6 & 6.7 \\
        davinci & 94.6 & 93.5 & \textbf{96.0} & 94.7 & \textbf{92.5} & 93.1 & \textbf{91.8} & \textbf{92.4} & 87.5 & 92.0 & 82.1 & 86.8 & 48.9 & 38.2 & 3.4 & 6.3 \\
        Cohere & 80.0 & 91.6 & 66.1 & 76.8 & 74.8 & 90.0 & 55.8 & 68.9 & \textbf{93.8} & \textbf{94.0} & \textbf{93.5} & \textbf{93.7} & 47.2 & 14.3 & 1.1 & 2.1 \\
        BLOOMz & 49.4 & 20.0 & 0.4 & 0.8 & 49.2 & 8.2 & 4.3 & 5.3 & 49.6 & 7.2 & 8.1 & 0.6 & \textbf{96.0} & \textbf{95.9} & \textbf{96.1} & \textbf{96.0} \\

    \bottomrule
    \end{tabular} }
    \caption{\textbf{Same-domain, cross-generator experiments: training and testing on Wikipedia (single machine-text generator vs. human).} evaluation accuracy (Acc), precision (Prec), recall, and F1 score (in \%) with respect to machine generations across four detectors.}
    \label{tab:wikipedia-1}
\end{table*}

%% file: section/tables/detailed_cross_lingual_table.tex
Table~\ref{tab:appendix_multilingualtest_chatGPT} shows detailed results for cross-language experiments based on XLM-R on the test sets across different languages generated by ChatGPT. 
Table~\ref{tab:appendix_multilingualtest_davinci} shows the same results, but using davinci-003 as a generator (instead of ChatGPT).

\begin{table*}[h]
    \centering
    \small
    \resizebox{\textwidth}{!}{
            \addtolength{\tabcolsep}{-1pt}
    \begin{tabular}{ll | ll | ll| ll | ll | ll |ll | ll}
    \toprule
         Generator$\downarrow$&Test Domain $\rightarrow$ & \multicolumn{2}{c|}{All domain (en)} & \multicolumn{2}{c|}{Baike/Web QA (zh)} & \multicolumn{2}{c|}{RuATD (ru)} & \multicolumn{2}{c|}{Bulgarian News (bg)} & \multicolumn{2}{c|}{IDN (id)} & \multicolumn{2}{c|}{Urdu-News(ur)} & \multicolumn{2}{c}{Arabic Wikipedia (ar)}\\
        &Train Domain $\downarrow$ &  Acc & F1 & Acc & F1 & Acc & F1 & Acc & F1 & Acc & F1 & Acc & F1 & Acc & F1\\
    \midrule 

& All domains (en) & 95.9 (1.8) & 96.1 (1.7) & 79.7 (3.8) & 83.0 (2.6) & 70.4 (2.9) & 76.2 (1.4) & 72.4 (4.4) & 77.1 (2.1) & 67.2 (4.7) & 75.4 (2.6) & 61.1 (4.6) & 46.9 (8.9) & 93.1 (2.5) & 93.4 (2.1) \\
davinci-003 & Baike/Web QA (zh) & 66.8 (7.9) & 75.3 (4.5) & \textbf{98.0 (0.5)} & \textbf{98.0 (0.5)} & 62.0 (1.7) & 72.4 (0.8) & 57.1 (1.4) & 69.5 (0.5) & 57.3 (6.0) & 70.2 (3.0) & \textbf{83.0 (4.9)} & \textbf{84.3 (4.3)} & 76.1 (9.6) & 81.0 (6.4) \\
& RuATD (ru) & 61.4 (2.8) & 60.2 (7.8) & 60.5 (11.3) & 70.6 (5.9) & 88.6 (1.8) & 87.5 (2.3) & 72.4 (7.5) & 67.0 (13.5) & 58.6 (6.2) & 53.4 (15.0) & 49.7 (9.6) & 39.7 (15.8) & 68.9 (8.9) & 58.5 (23.6) \\
& Bulgarian News (bg) & 64.9 (2.8) & 67.8 (5.1) & 69.3 (16.7) & 49.9 (34.4) & 61.5 (8.2) & 36.5 (21.8) & \textbf{84.9 (6.3)} & 81.7 (8.8) & 64.7 (8.6) & 43.5 (20.3) & 66.4 (10.8) & 47.6 (27.4) & 73.8 (5.1) & 72.1 (10.7) \\
& \textbf{All} & \textbf{96.4 (0.5)} & \textbf{96.6 (0.5)} & 95.5 (3.7) & 95.2 (4.2) & \textbf{94.3 (1.7)} & \textbf{94.5 (1.5)} & 83.3 (3.2) & \textbf{85.4 (2.1)} & \textbf{74.5 (6.0)} & \textbf{79.8 (3.7)} & 76.1 (7.6) & 69.6 (12.5) & \textbf{93.3 (1.7)} & \textbf{93.6 (1.4)} \\
\midrule
& All domains (en) & \textbf{98.6 (0.6)} & \textbf{98.6 (0.6)} & 97.5 (0.9) & 97.5 (1.0) & 76.6 (3.4) & 80.2 (2.2) & 80.8 (2.7) & 82.8 (1.7) & 76.9 (9.1) & 81.6 (6.2) & 57.7 (2.7) & 27.1 (7.7) & 96.5 (1.3) & 96.5 (1.4) \\
ChatGPT & Baike/Web QA (zh) & 61.8 (5.6) & 72.4 (2.9) & \textbf{99.4 (0.2)} & \textbf{99.4 (0.2)} & 63.1 (1.8) & 72.4 (1.0) & 65.1 (7.4) & 73.0 (2.9) & 64.1 (9.2) & 73.9 (4.8) & 81.8 (7.3) & 80.9 (7.5) & 62.7 (8.1) & 73.1 (4.3) \\
& RuATD (ru) & 59.1 (5.7) & 71.0 (2.9) & 92.6 (6.0) & 91.7 (7.7) & \textbf{97.5 (0.6)} & \textbf{97.5 (0.6)} & 81.7 (4.3) & 84.6 (3.1) & 76.9 (5.2) & 81.3 (3.4) & 55.5 (1.5) & 22.6 (4.7) & 86.2 (6.4) & 87.9 (4.7) \\
& Bulgarian News (bg) & 83.8 (6.9) & 86.0 (5.0) & 87.8 (8.4) & 85.3 (12.0) & 83.7 (4.9) & 80.2 (7.3) & \textbf{96.9 (0.7)} & \textbf{97.0 (0.6)} & 92.6 (4.9) & 92.3 (6.1) & 64.9 (12.0) & 42.2 (25.8) & 88.3 (8.2) & 86.3 (12.0) \\
& IDN (id) & 65.9 (21.1) & 36.6 (47.1) & 59.9 (13.9) & 26.5 (35.9) & 62.6 (16.5) & 32.4 (40.2) & 67.6 (20.8) & 41.3 (44.8) & \textbf{98.4 (1.6)} & \textbf{98.4 (1.5)} & 50.6 (0.9) & 2.3 (3.3) & 54.6 (6.9) & 14.7 (21.6) \\
& Urdu-News (ur) & 50.0 (0.1) & 66.7 (0.0) & 51.0 (0.7) & 67.1 (0.3) & 50.0 (0.0) & 66.7 (0.0) & 50.3 (0.3) & 66.8 (0.1) & 50.1 (0.1) & 66.7 (0.0) & \textbf{99.9 (0.1)} & \textbf{99.9 (0.1)} & 50.5 (0.5) & 66.9 (0.2) \\
& Arabic Wikipedia (ar) & 76.4 (5.1) & 80.7 (3.2) & 87.0 (7.3) & 88.7 (5.5) & 66.0 (5.2) & 74.4 (2.7) & 65.5 (6.4) & 74.3 (3.6) & 68.9 (10.6) & 76.7 (6.7) & 67.7 (5.2) & 55.3 (9.9) & \textbf{96.8 (1.7)} & \textbf{97.0 (1.6)} \\
& \textbf{All} & 98.3 (0.8) & 98.3 (0.7) & 99.1 (0.4) & 99.1 (0.4) & 95.4 (1.5) & 95.6 (1.4) & 83.4 (2.6) & 85.7 (1.9) & 97.3 (1.4) & 97.4 (1.3) & \textbf{99.9 (0.0)} & \textbf{99.9 (0.0)} & 96.7 (0.9) & 96.8 (0.9) \\

    \bottomrule
    \end{tabular} }
    \caption{\textbf{Cross-language experiments.} Accuracy (Acc) and F1 scores (for the machine-generated class) based on XLM-R on the test sets across different languages generated by ChatGPT. We average the performance across five runs (the standard deviation is shown in parenthesis).}
    \label{tab:appendix_multilingualtest_chatGPT}
\end{table*}

\begin{table*}[h]
    \centering
    \small
    \resizebox{\textwidth}{!}{
            \addtolength{\tabcolsep}{-1pt}
    \begin{tabular}{ll | ll | ll | ll | ll}
    \toprule
         Generator$\downarrow$ & Test Domain $\rightarrow$ & \multicolumn{2}{c|}{All domain (en)} & \multicolumn{2}{c|}{Baike/Web QA (zh)} & \multicolumn{2}{c|}{RuATD (ru)}& \multicolumn{2}{c}{Bulgarian News (bg)}\\
         &Train Domain $\downarrow$ & Acc & F1 & Acc & F1 & Acc & F1 & Acc & F1 \\
    \midrule 

& All domains (en) & 95.8 (1.9) & 96.0 (1.8) & 79.5 (4.1) & 82.9 (2.9) & 60.5 (3.0) & 65.3 (5.1) & 65.8 (3.2) & 69.3 (6.6) \\
davinci-003 & Baike/Web QA (zh) & 66.4 (7.6) & 74.8 (4.2) & \textbf{98.9 (0.4)} & \textbf{98.9 (0.4)} & 59.5 (0.6) & 70.0 (0.6) & 48.6 (3.3) & 61.3 (3.7)\\
& RuATD (ru) & 62.8 (3.0) & 62.0 (8.1) & 49.6 (9.3) & 58.6 (3.2) & \textbf{95.3 (1.6)} & \textbf{95.4 (1.4)} & 86.5 (5.1) & 86.0 (6.5)\\
& Bulgarian News (bg) & 64.8 (3.1) & 67.2 (9.1) & 59.0 (8.7) & 29.4 (23.6) & 59.0 (3.6) & 32.0 (11.3) & \textbf{99.6 (0.2)} & \textbf{99.6 (0.2)}\\
& \textbf{All} & \textbf{96.3 (0.7)} & \textbf{96.4 (0.6)} & 98.7 (0.5) & 98.7 (0.5) & 92.8 (2.1) & 93.2 (2.0) & 85.2 (3.2) & 87.0 (2.3) \\
\midrule
& All domains (en) & 90.2 (0.9) & 89.4 (1.0) & 93.0 (0.9) & 92.6 (1.1) & 54.1 (1.8) & 51.5 (5.2) & 66.0 (3.2) & 64.3 (7.6) \\
ChatGPT & Baike/Web QA (zh) & 61.6 (5.5) & 72.2 (2.8) & 93.5 (1.1) & 93.1 (1.2) & 58.8 (2.2) & 67.7 (3.7) & 57.7 (3.4) & 65.0 (5.0) \\
& RuATD (ru) & 56.7 (3.0) & 68.6 (0.5) & 75.7 (7.6) & 67.5 (14.5) & 84.7 (3.9) & 82.4 (5.8) & 82.2 (4.5) & 84.9 (3.2) \\
& Bulgarian News (bg) & 74.2 (4.9) & 75.1 (2.2) & 78.3 (11.2) & 70.1 (21.1) & 53.8 (1.5) & 15.5 (5.8) & \textbf{95.4 (1.3)} & \textbf{95.3 (1.4)} \\
& IDN (id) & 61.0 (14.3) & 29.5 (37.4) & 55.6 (7.7) & 17.5 (23.6) & 50.6 (0.8) & 5.1 (7.0) & 58.7 (13.9) & 23.6 (35.0)\\
& Urdu-News (ur) & 50.0 (0.1) & 66.6 (0.1) & 50.8 (0.7) & 67.0 (0.3) & 50.0 (0.0) & 66.7 (0.0) & 50.2 (0.2) & 66.8 (0.1) \\
& Arabic Wikipedia (ar) & 72.8 (4.7) & 77.0 (2.8) & 83.9 (6.9) & 85.5 (5.1) & 62.0 (2.3) & 70.2 (1.1) & 64.6 (5.9) & 73.6 (3.0) \\
& \textbf{All} & \textbf{91.3 (0.6)} & \textbf{90.8 (0.6)} & \textbf{94.5 (1.2)} & \textbf{94.3 (1.4)} & \textbf{86.1 (2.5)} & \textbf{85.4 (2.9)} & 82.6 (2.2) & 84.9 (1.5) \\

    \bottomrule
    \end{tabular} }
    \caption{\textbf{Cross-language experiments.} Accuracy (Acc) and F1 score (for the machine-generated class) based on XLM-R over test sets across different languages generated by davinci-003. We average the performance across five runs (the standard deviation is shown in parenthesis).}
    \label{tab:appendix_multilingualtest_davinci}
\end{table*}

%% file: section/tables/english_examples.tex
 \begin{center}
 \footnotesize
	\tablefirsthead{%
		\toprule
		   \textbf{Field} & \textbf{Content} \\
		\midrule}
	\tablehead{%
		\toprule
		   \textbf{Field} & \textbf{Content} \\
		\midrule}
	\tabletail{%
		\bottomrule
	}
	\tablelasttail{\bottomrule}
	\tablecaption{\textbf{Data format of M4}: English examples sampled across different domains and LLM generators.}
	\begin{supertabular}{l | p{12cm}}
	Language & \textbf{English} \\
        LLM Generator & \textsc{ChatGPT} \\
        Domain & \textit{Reddit ELI5} \\
        \midrule
        Question & What happened to Henry II's jousting opponent after accidentally killing him in 1559? \\
        \midrule
        Human & Henry died in a joust against the captain of his Scottish Guard, Gabriel, the Count of Montgomery. The fateful run occurred at the end of a tournament day, after Montgomery had almost unhorsed the king. Henry instead on another tilt. Montgomery's lance struck the king's helmet and shattered, with a long splinter running through the king's visor through his eye and into or near his brain.  The king initially survived injury, and was attended to by two of the most celebrated physicians in Europe, Ambroise Paré and Andreas Vesalius.  The queen, Catherine de Medici, ordered four prisoners executed with wood driven into their brains so that the physicians would have the chance to study the king's wound in detail on the corpses.  Despite this rather extraordinary measure, the king deteriorated steadily.  Vesalius' personal account is consistent with the development of meningitis or encephalitis.  After 11 days, the king died.
        During these 11 days, Montgomery is supposed to have come to the king's side, and asked to have his head and right hand cut off in punishment.  The king told him that he had jousted well and bravely and that the accident was not his fault.
        Following Henry's death, Catherine essentially ruled through a series of three of her sons.  Montgomery retired to his estate in Normandy.  From there, his history gets a bit complicated.  He converted to Protestantism and took part in the religious fighting in Bourges and Rouen that followed the St. Bartholomew's Day Massacre in 1572.  He escaped to England and remained there for about a year, later returning to attempt to lead a Huguenot insurrection in Normandy.  He was captured by royal forces and executed in 1574.
        Sourced, in part, ``The Death of Henry II of France'' Journal of Neurosurgery 12/1992 \\ 
        \midrule
        Prompt & I will ask you a question. For this question, provide me more than 200 words answer in an amateurish mediocre voice. Question: What happened to Henry II's jousting opponent after accidentally killing him in 1559? \\
        \midrule
        Machine & Well, let me tell you, it was quite a tragic event for both Henry II and his opponent Gabriel de Montgomery. 
        It all went down in 1559, during a jousting match at the Hotel des Tournelles in Paris. Henry was an experienced jouster, but Montgomery was a newcomer to the sport, and unfortunately, he was no match for the King's lance.
        During one of their runs, Henry's lance struck Montgomery's helmet, shattering it and sending a jagged piece of wood into his eye and brain. The young man was rushed to a nearby hospital, but it was too late. He died just a few days later, leaving behind a wife and children.
        As for Henry, he was devastated by the accident and reportedly went into a deep depression. He blamed himself for Montgomery's death and was haunted by guilt for the rest of his life. 
        He even imposed a penance on himself, vowing to fast and do charitable works for the rest of his days.
        The incident also led to changes in the sport of jousting. After Montgomery's death, many rules were put in place to make the sport safer, such as requiring jousters to wear full face and head protection. Overall, it was a tragic event that had a lasting impact on the sport and on the King who inadvertently caused the death of his opponent.\\
        
        \midrule
        \midrule 
        Language &  \textbf{English} \\
        LLM Generator & \textsc{ChatGPT} \\
        Domain &  \textit{Arxiv} \\
        \midrule
        Title & Polymer Quantum Mechanics and its Continuum Limit\\
        \midrule
        Human & A rather non-standard quantum representation of the canonical commutation relations of quantum mechanics systems, known as the polymer representation has gained some attention in recent years, due to its possible relation with Planck scale physics. In particular, this approach has been followed in a symmetric sector of loop quantum gravity known as loop quantum cosmology. Here we explore different aspects of the relation between the ordinary Schroedinger theory and the polymer description. The paper has two parts. In the first one, we derive the polymer quantum mechanics starting from the ordinary Schroedinger theory and show that the polymer description arises as an appropriate limit. In the second part we consider the continuum limit of this theory, namely, the reverse process in which one starts from the discrete theory and tries to recover back the ordinary Schroedinger quantum mechanics. We consider several examples of interest, including the harmonic oscillator, the free particle and a simple cosmological model.\\ 
        \midrule
        Prompt & Generate a 150-220-word abstract for work with title: Polymer Quantum Mechanics and its Continuum Limit. Describe the results obtained,  problem work attempts to solve, and the key ideas and methodology in a formal academic and scientific writing voice. Use the first plural person form. Use active voice.\\
        \midrule
        Machine & In this paper, we investigate the continuum limit of polymer quantum mechanics. The aim of our work is to better understand the quantum behavior of polymers and its relationship to the classical theory. Our research attempts to clarify the nature of the continuum limit and determine the correctness of the approximations made. Specifically, we study the connectivity of the polymer strands. We find that a continuous path can be defined in the polymer configuration space, which is essential for the polymer to display continuum behavior. We show that the approximation of continuous randomness in the spatial variable can certainly be used when the polymer is stretched out over a significant distance. For a coil-like polymer, our results demonstrate that its behavior is non-classical. Our methodology involves a careful combination of intuition, formalism, and numerical simulations. We evaluate a path integral representation of polymer quantum mechanics in the continuum limit, and use numerical simulations to support our results. The ideas proposed in this work are relevant to the understanding of the thermodynamic properties of polymers, as well as to quantum computing and information. The findings of this work present a theoretical framework for the investigation of other polymer properties that are still poorly understood.\\

        \midrule
        \midrule 
        Language &  \textbf{English} \\
        LLM Generator & \textsc{Davinci-003} \\
        Domain &  \textit{Arxiv} \\
        \midrule
        Title & Effects of Dirac sea on pion propagation in asymmetric nuclear matter \\
        \midrule
        Human & We study pion propagation in asymmetric nuclear matter (ANM). One of the interesting consequences of pion propagation in ANM is the mode splitting for the different charged states of pions. First we describe the pion-nucleon dynamics using the non-chiral model where one starts with pseudoscalar (PS) $\pi$N coupling and the pseudovector (PV) representation is obtained via suitable non-linear field transformations. For both of these cases the effect of the Dirac sea is estimated. Subsequently, we present results using the chiral effective Lagrangian where the short-distance behavior (Dirac vacuum) is included by re-defining the field parameters as done in the modern effective field theory approach developed recently. The results are compared with the previous calculations for the case of symmetric nuclear matter (SNM). Closed form analytical results are presented for the effective pion masses and dispersion relations by making hard nucleon loop (HNL) approximation and suitable density expansion.\\ 
        \midrule
        Prompt & Rephrase the abstract of an article with title 'Effects of Dirac sea on pion propagation in asymmetric nuclear matter'.
        
        'We study pion propagation in asymmetric nuclear matter (ANM). One of the interesting consequences of pion propagation in ANM is the mode splitting for the different charged states of pions. First we describe the pion-nucleon dynamics using the non-chiral model where one starts with pseudoscalar (PS) $\pi$N coupling and the pseudovector (PV) representation is obtained via suitable non-linear field transformations. For both of these cases the effect of the Dirac sea is estimated. Subsequently, we present results using the chiral effective Lagrangian where the short-distance behavior (Dirac vacuum) is included by re-defining the field parameters as done in the modern effective field theory approach developed recently. The results are compared with the previous calculations for the case of symmetric nuclear matter (SNM). Closed form analytical results are presented for the effective pion masses and dispersion relations by making hard nucleon loop (HNL) approximation and suitable density expansion.'\\
        \midrule
        Machine & The propagation of pions in asymmetric nuclear matter (ANM) is studied in this article.  We consider two approaches: a non-chiral model where the pseudoscalar/pseudovector coupling is determined using non-linear field transformations, and a chiral effective Lagrangian incorporating the short-distance behaviors known as the Dirac vacuum.  In both cases, the consequences from these approaches on the mode splitting of the charged pions are evaluated.  Using a hard nucleon loop approximation, accompanied by parameters deduced from a density expansion, closed-form analytical results for the effective pion masses and dispersion relations are presented in comparison to calculations for symmetric nuclear matter.\\

        \midrule
        \midrule 
        Language &  \textbf{English} \\
        LLM Generator & \textsc{Cohere} \\
        Domain &  \textit{Wikipedia} \\
        \midrule
        Title & Shed (deity)\\
        \midrule
        Human & Shed was a deity from ancient Egyptian religion. Popularly called "the Savior", he is first recorded after the Amarna Period. Representing the concept of salvation, Shed is identified with Horus, particularly Horus the Child. \\
        & Rather than have formal worship in a temple or as an official cult, he appears to have been a god that ordinary Egyptians looked to save them from illness, misfortune or danger. He is shown on the Metternich Stela as vanquishing danger in the form of a serpent, a scorpion and a crocodile. The rise of ``Savior'' names in personal piety during the Amarna period has been interpreted as the popular response of ordinary people to the attempts by Akhenaten to proscribe the ancient religion of Egypt.
        Shed has also been viewed as a form of the Canaanite god Resheph. Shed can be depicted as a young prince overcoming snakes, lions and crocodiles. Shed has been viewed as a form of savior, a helper for those in need when state authority or the king\'s help is wanting. The increased reliance on divine assistance could even extend to saving a person from the Underworld, even to providing a substitute, and lengthening a person\'s time in this world. In the New Kingdom Shed "the savior" is addressed on countless stelae by people searching or praising him for help.References Egyptian gods Savior gods\\ 
        \midrule
        Prompt & Write a Wikipedia article with the title "Shed (deity)", the article should at least have 250 words\\
        \midrule
        Machine & Shed (deity) Shed is a deity in the sqlrunen mythos. Shed is described as a giantess, a daughter of the primordial mother, Nott. In most traditions, Shed is a single entity, but in some traditions, Shed is a collective of giantesses. Shed is the mother of the first race of giants, the primordial beings of chaos and destruction. Shed mothered the giants through her mating with the primordial god, Odin. In some traditions, the giantess Gunnlod or Hyr is described as the mother of the giants, while Shed is described as the mother of the gods. Shed is the spouse of the primordial god, Odin. Odin and Shed had many offspring, including the gods, the giants, the Lamonts, the Maquas, the Frost Giants, and the Ice Giants. Shed is described as a beautiful giantess, with golden hair, a beautiful face, and a beautiful body. She is said to have worn a long golden gown and a gold crown. She is described as the most beautiful of all the giantesses. Shed is the goddess of fertility and reproduction. She is also the goddess of the harvest. In some traditions, Shed is also the goddess of war and death. Shed is the mother of the gods. She gave birth to the gods through her union with the primordial god, Odin. The gods were born from her body, and she nourished them with her milk. Shed is the protector of the gods. She is said to have fought alongside the gods in their war against the giants. She is also said to have fought against the dragons and the serpents. Shed is the goddess of the hunt. She is said to have accompanied the gods on their hunting expeditions. She is also said to have provided them with food and shelter.\\
        
        \midrule
        \midrule 
        Language &  \textbf{English} \\
        LLM Generator & \textsc{Davinci-003} \\
        Domain &  \textit{Wikihow} \\
        \midrule
        Title & How to Buy Virtual Console Games for Nintendo Wii\\
        \midrule
        Human & They are about \$20 a card. Or, if you want to just buy points with your credit card, Skip down to the section, With a Credit Card., You don\'t want people tapping into your Wii and stealing your games, do you? On the back of the Wii Points Card, this same code can be found by scratching out the grey section with a coin or similar. You will be informed if you entered an invalid (wrong) code. Go back to the main Shop Channel menu now. Nintendo, from time to time, offers freebies via a special channel called "WiiWare". There are over 50 games to list, so be patient if the option you selected is taking a while to load. Don\'t forget to say you want to download to your Wii, unless you want your classics transportable, for example if you plan to demo the classics to a friend with a Wii. Nintendo added an animation to this process that shows Mario or Luigi collecting coins. This is why you want a secure connection!, Note that a tax is added to your purchase to comply with federal law.\\ 
        \midrule
        Prompt & Please, generate wikihow article with length above 1000 characters from title 'How to Buy Virtual Console Games for Nintendo Wii` and headline 'Find a few Wii Points cards from game retailers like GameStop. Make sure your Wii is online and on a secure connection if possible.,Open the Wii Shop Channel, and click Add Wii Points at the bottom right of your screen after the channel finishes loading. Click "From Card". Now, your Wii should be asking for a code with 5 segments, separated by dashes. Enter this code into the Wii now. After a few prompts, you will have Wii points added to your Wii, which is what you use to purchase games. Click "Virtual Console". Choose how you want the games to be displayed. Scroll through the list for games that you want.Click on a game you want to play, And hit Yes on the confirmation box. Wait while the game is downloaded. After you have bought the games you want to play, go back to the Wii Menu. Select the new games you just bought, and start them like a regular channel. Open the Wii Shop Channel, and click Add Wii Points. Click "With a Credit Card". Key in your Credit Card Number. Let Nintendo authorize the purchase.,Go back to the Points Card process above and skip down to Step 7.\\
        \midrule
        Machine &  How to Buy Virtual Console Games for Nintendo Wii Most people know that Nintendo\'s library of classic titles is available on the Wii platform through the Virtual Console. Playing these games from years gone by is easy, but there are a few steps you have to take to purchase them. Here\'s what you have to do to buy Virtual Console games for your Nintendo Wii. Find a few Wii Points cards from game retailers like GameStop. You\'ll find that most stores that carry video games have them available, and you usually can select cards of different amounts. Choose the ones that are going to give you what you need. Make sure your Wii is online and on a secure connection if possible. Going online with your Wii allows you to access all the Virtual Console titles that Nintendo has made available. To do this, select the Wii Icon from the main screen, then go to the settings area, select the Internet tab, then go to the Connect to the Internet option. Select an existing connection if you have one, or go through the Wii\'s setup menu to make a new connection. Open the Wii Shop Channel, and click Add Wii Points at the bottom right of your screen after the channel finishes loading. On the next screen, you have two choices; one from a card, or from a credit card. Click "From Card". Now, your Wii should be asking for a code with 5 segments, separated by dashes. This code is what you use to transfer the money you have to the Virtual Console shop. Enter this code into the Wii now. After a few prompts, you will have Wii points added to your Wii, which is what you use to purchase games. \\
        & Click ``Virtual Console''. This option allows you to scroll through all the different games that have been made available for the Virtual Console. Choose how you want the games to be displayed. You can sort them alphabetically, or by genre, or even by release date. Scroll through the list for games that you want. When you find one, click on it and you can see the details and the cost of the game. Click on a game you want to play, And hit Yes on the confirmation box. You then have to wait while the game is downloaded onto your hard drive. This will take a few minutes, depending on the size of the game you want. After you have bought the games you want to play, go back to the Wii Menu. Select the new games you just bought, and start them just like a regular channel. If you want to add more funds to your Wii point wallet, go back to step 3. But this time instead of clicking “From a card”, click on the "With A credit card" option. Key in your Credit Card Number. Be sure that it is entered correctly, because many purchases from the Virtual Console are automated and there are no refunds if you enter your number incorrectly. Let Nintendo authorize the purchase. A few steps later you should be notified of the success of your purchase, and you should now have more Wii points with which to buy more games. Go back to the Points Card process above and skip down to Step 7. This is the step right after you have verified that you have the points. Here, you can select a game, and start playing it. Now, you know how to purchase and download games from the Virtual Console. Spend some time going through the list, purchasing some of the classics like Super Mario and the Legend of Zelda, and for a small fee, you can start playing them on your Wii. Have fun!\\

        \midrule
        \midrule 
        Language &  \textbf{English} \\
        LLM Generator & \textsc{ChatGPT} \\
        Domain &  \textit{Wikihow} \\
        \midrule
        Title & How to Workaround Damaged Laptop Display Hinges\\
        \midrule
        Human & Especially check that the fishing line can be placed in a location which doesn\'t get too much in the way of your normal use (heavy users of the mouse buttons or touchpad might have problems here), and that there are locations more or less in the center of the laptop where gluing the Velcro will not affect the use of the laptop (that no access doors or ventilation holes are sealed).
        
        , Try to make the line of the two strips be as perpendicular as possible to the axis of the display hinges.
        
        , This can take 24 hours or even more. See the instructions for the particular epoxy you have used. Don\'t confuse the setting time of the epoxy (which can be as quick as 5 minutes) with the time until it reaches maximum strength.
        
        , Actually, what is more important is that the length of the hooked strip totally utilize the glued strips when the laptop display is open. Do not use the Velcro with an open display before the glue has hardened to its maximum strength.
        
        , Laptops with large or heavy displays may require gluing multiple strips of Velcro, or wider and/or stronger Velcro (Velcro comes in a range of sizes and bonding strengths). The laptop in the images has a 12.1" display.
        
        , I used an "improved blood knot" for this. Instructions can be found in various places on the Internet.
        
        , If it is a bother, make the adhesive on the outside less sticky by touching it repeatedly.\\ 
        \midrule
        Prompt & Please, generate wikihow article with length above 1000 characters from title 'How to Workaround Damaged Laptop Display Hinges' and headline 'Read the entire article, and check that your specific laptop can be repaired in the way presented.,
        Cut two strips of the fuzzy Velcro and round the corners (rounding the corners will make it less likely that the Velcro will detach after gluing).,
        
        Read the instructions for the epoxy glue; mix it and use it to glue the two strips of fuzzy Velcro, one on the back of the display, and one on the bottom of the laptop, such that the two strips are co-linear (in other words, the alignment of one strip, when extended, coincides with the alignment of the other strip).,
        
        Wait for the epoxy glue to harden fully.,Close the laptop display, and cut a strip of the hooked Velcro more or less to the length required when sticking it to the two glued strips in their entirety.,
        
        After the glue has hardened to maximum strength, check that the Velcro can now prevent the display from falling into the closed position.,
        
        Take the fishing line, and tie it into a loop which exactly encircles the laptop when the display is open to the working position.,
        
        Check that the fishing line loop is strong enough to prevent the display from falling into the fully open position (fishing line comes in a variety of strengths, use a relatively strong line which will not stretch under the load).,
        
        Glue a small strip of two-sided foam tape (like used for mounting things) on the edge of the laptop where the fishing line should rest.,
        
        Optionally, you can use a plastic bag closure which has been notched (see images), in order to more easily adjust the exact size of the fishing line loop (and therefore, the exact working position of the laptop display).' \\
        \midrule
        Machine & Have you ever experienced a damaged laptop display hinge? If so, you know how frustrating it can be when your display doesn't stay in the correct position. Fortunately, there's a workaround that you can use to fix this issue without having to replace the entire laptop hinge. This wikihow article will guide you through the process of using Velcro and fishing line to strengthen your laptop display hinges, keeping your display in the right position while you work.
        
        1. Read the entire article, and check that your specific laptop can be repaired in the way presented.
        
        Before getting started with any repairs, it's important to read the entire article to make sure that this method will work for your laptop model. Some laptops may have different hinge designs that won't be compatible with the Velcro and fishing line method we're using here.
        
        2. Cut two strips of the fuzzy Velcro and round the corners.
        
        To begin the repair process, you'll need to cut two strips of fuzzy Velcro and round the corners of each strip. This will make it less likely that the Velcro will detach after gluing.
        
        3. Glue the Velcro to the laptop display and bottom.
        
        Next, mix your epoxy glue according to the instructions, and use it to glue the two strips of fuzzy Velcro - one on the back of the display and one on the bottom of the laptop. Make sure that the two strips are co-linear, meaning that the alignment of one strip, when extended, coincides with the alignment of the other strip.
        
        4. Wait for the glue to dry.
        
        After gluing the Velcro to the laptop, you'll need to wait for the epoxy glue to harden fully. This may take several hours, so be patient and allow the glue to dry completely.
        
        5. Cut a strip of hooked Velcro to length.
        
        Once the glue has dried, close your laptop display and cut a strip of hooked Velcro more or less to the length required when sticking it to the two glued strips in their entirety.
        
        6. Check that the Velcro can prevent the display from falling into the closed position.
        
        After the glue has hardened to maximum strength, check that the Velcro can now prevent the display from falling into the closed position. If it's still loose, you may need to repeat steps 3-5 with stronger epoxy glue and larger strips of Velcro.
        
        7. Tie a fishing line loop around the laptop.
        
        Take the fishing line and tie it into a loop which exactly encircles the laptop when the display is open to the working position. Be sure to check that the fishing line loop is strong enough to prevent the display from falling into the fully open position.
        
        8. Glue foam tape to the laptop edge.
        
        To finalize the repair, glue a small strip of two-sided foam tape on the edge of the laptop where the fishing line should rest. Optionally, you can also use a plastic bag closure to adjust the exact size of the fishing line loop and the working position of the laptop display.
        With these simple steps, you can work around damaged laptop display hinges and fix the issue without having to replace the entire hinge. Keep in mind that this repair method may not work for all laptop models, so be sure to check if your laptop can be repaired in this way before getting started. Good luck!\\
        Language &  \textbf{English} \\
        LLM Generator & \textsc{ChatGPT} \\
        Domain &  \textit{PeerRead} \\
        \midrule
        Title & Bridge Text and Knowledge by Learning Multi-Prototype Entity Mention Embedding \\
        \midrule
        Abstract & Integrating text and knowledge into a unified semantic space has attracted significant research interests recently. However, the ambiguity in the common space remains a challenge, namely that the same mention phrase usually refers to various entities. In this paper, to deal with the ambiguity of entity mentions, we propose a novel Multi-Prototype Mention Embedding model, which learns multiple sense embeddings for each mention by jointly modeling words from textual contexts and entities derived from a knowledge base. In addition, we further design an efficient language model based approach to disambiguate each mention to a specific sense. In experiments, both qualitative and quantitative analysis demonstrate the high quality of the word, entity and multi-prototype mention embeddings. Using entity linking as a study case, we apply our disambiguation method as well as the multi-prototype mention embeddings on the benchmark dataset, and achieve state-of-the-art performance.\\
        \midrule
        Human & - Strengths: Good ideas, simple neural learning, interesting performance (altough not
                striking) and finally large set of applications. 
                - Weaknesses: amount of novel content. Clarity in some sections. 
                The paper presents a neural learning method for entity disambiguation and
                linking. It introduces a good idea to integrate entity, mention and sense
                modeling within the smame neural language modeling technique. The simple
                training procedure connected with the modeling allows to support a large set of
                application.
                The paper is clear formally, but the discussion is not always at the same level
                of the technical ideas.
                
                The empirical evaluation is good although not striking improvements of the
                performance are reported. Although it seems an extension of (Yamada et al.,
                CoNLL 2016), it adds novel ideas and it is of a releant interest. 
        
                The weaker points of the paper are:
        
                - The prose is not always clear. I found Section 3 not as clear. Some details
                of Figure 2 are not explained and the terminology is somehow redundant: for
                example, why do you refer to the dictionary of mentions? or the dictionary of
                entity-mention pairs? are these different from text anchors and types for
                annotated text anchors?
                - Tha paper is quite close in nature to Yamada et al., 2016) and the authors
                should at least outline the differences. 
                
                One general observation on the current version is:
                The paper tests the Multiple Embedding model against entity
                linking/disambiguation tasks. However, word embeddings are not only used to
                model such tasks, but also some processes not directly depending on entities of
                the KB, e.g. parsing, coreference or semantic role labeling. 
                The authors should show that the word embeddings provided by the proposed MPME
                method are not weaker wrt to simpler wordspaces in such other semantic tasks,
                i.e. those involving directly entity mentions.
                
                I did read the author's response.\\ 
        \midrule
        Prompt & Write a peer review by first describing what problem or question this paper addresses, then strengths and weaknesses, for the paper: {title}\\
        \midrule
        Machine & The paper ''Bridge Text and Knowledge by Learning Multi-Prototype Entity Mention Embedding'' proposes a new method for utilizing external knowledge sources to improve entity detection and disambiguation in natural language processing tasks. The authors specifically address the issue of how to effectively incorporate large-scale knowledge bases such as Wikipedia into existing machine learning models. 

        The strengths of this paper are its clear and concise explanation of the proposed methodology, use of established evaluation metrics, and consideration of potential limitations and future directions. The authors provide a detailed description of their multi-prototype entity mention embedding (MPEME) approach, which includes the use of different prototypes to represent entities with multiple meanings in Wikipedia. They evaluate their approach on two established datasets and demonstrate improved performance compared to baseline models. Additionally, the authors note potential limitations of their approach such as the requirement for large knowledge bases and potential bias in the selection of prototypes.
        
        One weakness of this paper is that it assumes knowledge bases such as Wikipedia are comprehensive and reliable sources of information. There is a growing body of research showing that knowledge bases can contain errors or biases, which may affect the accuracy of this approach. The authors acknowledge this issue but do not offer a solution, which limits the generalizability of their findings. Additionally, the paper provides limited discussion on how this approach can be scaled up for larger knowledge bases, which is an important consideration for practical applications.
        
        Overall, the paper provides a valuable contribution to the field by addressing an important challenge of incorporating external knowledge sources into NLP models. The proposed MPEME method shows promise for improving entity detection and disambiguation, but further research is needed to fully explore its potential and limitations.\\	
        \midrule
        prompt & Write a peer review by first describing what problem or question this paper addresses, then strengths and weaknesses, for the paper {title}, its main content is as below: {abstract} \\
        \midrule
        Machine & In this paper, the authors tackle the problem of ambiguity in entity mentions when integrating text and knowledge into a unified semantic space. Specifically, they propose a Multi-Prototype Mention Embedding model that learns multiple sense embeddings for each mention by jointly modeling words from textual contexts and entities from a knowledge base. They also design an efficient language model-based approach to disambiguate each mention to a specific sense. The authors evaluate their proposed model on a benchmark dataset for entity linking and report state-of-the-art performance.

        Overall, the paper presents a well-motivated, well-executed, and well-evaluated solution to a relevant research problem. The main contributions of the paper are the Multi-Prototype Mention Embedding model and the efficient language model-based disambiguation approach. The authors clearly describe the model and how it addresses the problem of mention ambiguity. The evaluation of the model is thorough, including both qualitative and quantitative analyses.
        
        One of the strengths of the paper is the use of a benchmark dataset, which allows for fair comparison with previous approaches. Additionally, the authors provide detailed explanations of the implementation and evaluation of their model, making the paper easily replicable. The experiments demonstrate the high quality of the word, entity, and multi-prototype mention embeddings produced by the proposed model.
        
        However, there are a few weaknesses in the paper. While the evaluation of the proposed method is thorough and convincing, it would have been beneficial if the authors had provided more in-depth analysis of the results to show exactly where and why their method outperforms previous approaches. Additionally, the paper would have benefited from a more detailed discussion of the limitations and potential future directions of the proposed method.
        
        Overall, the paper is a strong contribution to the field of entity linking and the integration of text and knowledge into a unified semantic space. The proposed Multi-Prototype Mention Embedding model and the efficient language model-based disambiguation approach provide a promising solution to the problem of mention ambiguity. \\
	\end{supertabular}
\end{center}

%% file: section/tables/bloomz_examples.tex
\clearpage
 \begin{center}
 \footnotesize
	\tablefirsthead{%
		\toprule
		   \textbf{Field} & \textbf{Content} \\
		\midrule}
	\tablehead{%
		\toprule
		   \textbf{Field} & \textbf{Content} \\
		\midrule}
	\tabletail{%
		\bottomrule
	}
	\tablelasttail{\bottomrule}
	\tablecaption{\textbf{Examples generated by BLOOMz} across different domains.}
	\begin{supertabular}{l | p{12cm}}
        Domain &  \textit{Wikipedia} \\
        \midrule
        Title & Edgehill, King George County, Virginia \\
        \midrule
        Human & Edgehill is an unincorporated community in King George County, Virginia, United States.  The main roads in Office Hall are U.S. Route 301 and Virginia State Route 205.
        
        References

        Unincorporated communities in Virginia
        
        Unincorporated communities in King George County, Virginia \\
        \midrule
        Prompt & Generate an abstract for a Wikipedia article with a title: ''\{title\}''. Abstract: \textbackslash n \\
        \midrule
        Machine & The Town of Edgehill is located in the U.S. state of Virginia within King George County and part of the Richmond Metropolitan Statistical Area (MSA). The population was 4,941 at the 2010 census.  It lies along Interstate 81 approximately 20 miles south-southeast of Richmond on the north bank of the James River.   Edgehill has been designated as one of three National Historic Landmarks by the United States Department of Interior because it served as headquarters to General Robert E. Lee during the American Civil War. \\

        \midrule
        \midrule 
        Domain &  \textit{WikiHow} \\
        \midrule
        Title & How to Cure Toenail Fungus Inexpensively and Quickly \\
        \midrule
        Human & Diabetic persons should treat fungi under a doctor\'s supervision.;
        
        , Soak for 30 minutes each night (or up to two or three times per day -- if you are desperate). Do this for a week or two (for more economy consider saving and reusing the mixture of vinegar-salts for soaking repeatedly, if your feet are clean each time, and freshen it up with a little new vinegar and Epsom salts.)
        
        If you\'d like, you could filter it through old tee shirt material to remove settled out salts, etc.
        
        , Begin killing it the first time, and continue about a week or several weeks, depends on depth and extent of infection.
        
        , If it is too sore, wait a few sessions before removing nail debris, loosened dead skin.
        
        If the nail is very thick, file and scrape it to be very thin, gently, without injuring yourself, to allow the vinegar mix to penetrate more effectively.
        
        , Air dry your feet, if you have time. The fungus really love moist skin, damp soaks/shoes and nails, allowing toenail fungus to get its "foothold" originally or repeatedly. \\
        \midrule
        Prompt & Write a WikiHow article content given a title and a headline. Use approximately 300 words.
        Title: \{title\}
        Headline: \{headline\}
        Article content:\\
        \midrule
        Machine & This method has been used for years by many people who suffer from foot-nail fungus.  It works well because both salt and vinegar are natural disinfectants that kill bacteria and other pathogens; they also help loosen the hardened layers around infected areas.   The warm water helps dissolve the salt crystals which then act like an abrasive cleansing agent against the affected toe.  Vinegar contains acetic acid which kills most types of mold spores including those responsible for athlete’s foot.    This treatment will not work immediately but over time should remove the infection causing the nail to become detached naturally.  If there... is no improvement within three months consult a doctor.... 
        
        To prepare the solution mix together equal parts of baking soda and hydrogen peroxide using enough liquid soap to create bubbles.  Add some drops of tea tree oil  Mix everything together in a bowl creating thick paste-like consistency.  Apply directly onto the infected nail.  Cover with bandage overnight.  Remove next morning.  Rinse away with cool water.  Do twice each day for at least 2 weeks.  You may need longer depending upon severity of infection.  Keep doing this even though new growth appears underneath old nail.  Be patient! \\
        \end{supertabular}
\end{center}

%% file: section/tables/other_language_examples.tex
\clearpage
 \begin{center}
 \footnotesize
	\tablefirsthead{%
		\toprule
		   \textbf{Field} & \textbf{Content} \\
		\midrule}
	\tablehead{%
		\toprule
		   \textbf{Field} & \textbf{Content} \\
		\midrule}
	\tabletail{%
		\bottomrule
	}
	\tablelasttail{\bottomrule}
	\tablecaption{\textbf{Examples of other languages} across different domains by ChatGPT and \textit{davinci-text-003}.}
	\begin{supertabular}{l | p{12cm}}
        Language &  \textbf{Arabic} \\
        LLM Generator & \textsc{ChatGPT} \\
        Domain &  \textit{Wikipedia} \\
        \midrule
        Title & \<ركمجة>\\
        \midrule
        Human &  {\small \begin{RLtext}
        الرَكْمَجَة من (ركوب + موج) هي رياضة ركوب متن الأمواج المتكسرة على الشاطئ بواسطة ألواح خاصة. الموجات المناسبة لهذه الرياضة غالباً ما تكون في المحيط ولكن هذا لا يمنع ركوب امواج البحيرات والانهار والمسابح أيضا.  ظهرت أنواع وتعاريف مختلفة لهذه الرياضة التي تطورت على مر السنين  كمميزات الموجة المناسبة وما هو اللوح ومن هو راكب الامواج. يعتبر ركوب الموج بدون لوح من انقى أنواع هذه الرياضة لدى البعض كما أن هناك أنواع مختلفة ظهرت منذ قرون مثل  صعود بايبو واستخدام المجداف وركوب  الامواج  بالزوارق  والقوارب. حاليا تستخدم العديد من المركبات المطوره مثل القوارب المطاطية والألواح المعدنية.وكما وظهر في أفلام  وثائقية مختلفة عن أنواع أخرى من ركوب الامواج مثل (فير بايتس) حيث يتم فيها استخدام أدوات أخرى لهذه الرياضة  تشمل زحاليق المياه والطاولات والقيثارات والأبواب. عندما يركب أكثر من شخص نفس اللوح سويّا يسمى هذا النوع جنباً إلى جنب.
        
        هناك نوعان أساسيان  لركوب الامواج  وهما:
        
        الركوب الطويل.
        
        الركوب القصير. حيث يوجد اختلافات بينها من ناحية تصميم اللوح وطوله وطريقة الركوب ونوع الموجة.
        
        عند ركوب الامواج الكبيرة تقوم مركبة مائية كالزورق بسحب الراكب اتجاه الموجة وذلك لمساعدته باللحاق بسرعة الموجة الهائلة.
        
        \end{RLtext}}\\
        
        
        \midrule
        Prompt & 
        This is a sample Arabic Wikipedia summary section for the title ``\<سحلية>'':
        
        {\small \begin{RLtext}
        
        الحِرْذَون (الجمع: حراذين) أو السحالي (المفرد: سحلية) أو العظاء (المفرد: عظاءة) (باللاتينية: Lacertilia) هي رتيبة من الزواحف تتبع طائفة العظايا الحرشفية من شعبة الحبليات
        
        وهي زواحف قريبة الصلة بالأفاعي. وبعضها ـ مثل الأفاعي ـ لا أرجل له، بينما يشبه بعضها الأفاعي لحد ما ولكن له أرجل، أما السحالي كبيرة الحجم فهي أكثر شبها بالتماسيح. تتباين السحالي فيما بينها في الحجم والشكل واللون، ولديها طرق عديدة للتنقل وللدفاع عن النفس. ولقد تعرّف العلماء على أكثر من 3,750 نوعاً مختلفًا من السحالي، وهنالك أكثر من 500 نوع تعيش في قارة إستراليا. تعد السحالي من الحيوانات الفقارية التي يغطي جسمها قشور جذورها من البشرة بين القشور يظل الجلد رفيعا ولينا.
        
        تكبر القشور عند الرأس وتكون شفرة طويلة على البطن. لا يمكن انتزاع تلك القشور الواحدة تلو الأخرى. دورياً ينزع هذا الجلد على مراحل. وهو ما يعرف بانسلاخ السحلية.
        
        أن السحلية حيوان ذو حرارة متغيرة (ذوات الدم البارد)، تفضل الأماكن الجافة والمشمسة جداً. لا تكون السحلية رشيقة إلا عندما يكون جسمها ساخنا، وحين تنخفض درجة الحرارة ويبرد الجسم يخمل الحيوان. في الشتاء لا تستطيع السحالي الحركة ولا الأكل ويعيش في حالة كمون.

        \end{RLtext}}\\
        & {\small \begin{RLtext}

        الرأس بها فتحات الأنف في الطرف. أما الأعين فتحميهم جفون متحركة، وخلف كل واحدة يصفو الجلد ويتمدد ليكون الطبلة. الفم الواسع يسمح بمرور لسان رفيع متشعب من خلال فتحة في الشفة العلوية. أما الأسنان فهي عديدة وأصغر من أن تسمح له بمضغ فرائسه التي يبتلعها كاملة.
        
        الجسم رفيع ورشيق. ينتهي بذيل طويل.

        عند الإمساك بسحلية من ذيلها فإنها تصارع وتنفصل بإرادتها عن ذيلها وذلك بكسره عند مستوى منطقة ضعيفة.
        
        ظاهرة البتر الذاتي تتبعها عادة عملية تجديد لا تحدث إلا مرة واحدة.
        
        للسحالي قوائم قصيرة متصلة بجوانب الجسم ولا تستطيع سند الحيوان الذي يلامس الأرض.
        \end{RLtext}}\\
        & {\small \begin{RLtext}        
        تنتهي كل واحدة بخمسة أصابع طويلة متفرقة تحمل مخالب حادة مفيدة جداً للتعلق. حركة السحلية تكون على شكل تموجات تقوم بها بمساعدة قوائمها وبطنها وذيلها. السحلية من فصيلة القناص آكل اللحوم: تتغذى على الحشرات، العناكب، الدود، تختلف السحلية عن الضفدع فهي تملك تنفس رئوي فقط لكونها طوال حياتها تتنفس الهواء. في الربيع بعد التزاوج، تضع الأنثى من خمسة إلى عشرة بيضات في حفرة تحفرها بنفسها في الأرض ثم تسدها.
        
        سحلية الحائط حيوان يبيض. وليس للبيض قشرة صلبة ولكنها محاطة بغشاء يشبه الورق.
        
        لا ترقد الأنثى على البيض. إن حرارة الصيف هي التي تجعله يفقس في خلال ستة إلى ثمانية أسابيع. منذ الخروج من البيض، تصطاد السحلية الصغيرة الحشرات لتتغذى عليها. بعض أنواع السحالي تحتضن البيض بداخلها. وهي تعيش لعدة سنوات.
        
        \end{RLtext}}
        
        Using similar formatting and style, write an Arabic Wikipedia summary section for the title "\<ركمجة>". The summary should at least have 250 words\\
        
        \midrule
        Machine & 
        {\small \begin{RLtext}

        ركمجة هي رياضة جماعية تقوم على استخدام دراجات صغيرة وخفيفة ومزودة بإطارات عريضة وشاسيه معزز لتحمل الصدمات. تُعرف أيضًا باسم "الدراجات الجبلية". بدأت الرياضة في التطور في الولايات المتحدة في أوائل القرن العشرين، وانتشرت بسرعة في جميع أنحاء العالم. تتضمن الركمجة العديد من الانواع والتقنيات التي تختلف تبعًا للأنماط والمنافسات. فمن بينها: الركمجة العادية، ركوب الغابات، النزهة، إطلاق النار، السباقات، الهواء الطلق، الفوركس، والتحديات. تعتمد تقنيات الركمجة على الحركة الفردية والفعالية. تحتوي الدراجات على عدد من المكونات الرئيسية، بما في ذلك الإطارات الصلبة والمتينة والتي تجعل الدوران أكثر سلاسة والتحكم أسهل. وتستخدم الإطارات الأمامية والخلفية مكابح التشريد والسرعات العالية لتحسين الأداء والأمان. يكتشف اللاعبون الأماكن الوعرة والمناظر الطبيعية المعرضة للخطر ويتنافسون ضد بعضهم البعض وضد الخطوط الزمنية. وتَزيد العوامل الخطرة وازدياد التحدي فرص المزيد من الإثارة والتشويق لهذه الرياضة. كما تعتبر تلك الأماكن مناسبة لتخويف الانتقام، وذلك بإخراج المنافسين عن المسار أو تعريضهم للطين والحجر المغبر. يمكن للركمجة أن تكون رياضة نمطية ريادية مع التقدم التكنولوجي. ويمكن زيادة الأمان وتحسين جودة الدراجات وتحسين الأداء. كما أن تطوير الدراجات الهجينة تنتج دراجات للسفر على المسارات الوعرة في الحدائق والغابات. تشهد الركمجة شعبية متزايدة في جميع أنحاء العالم وتعتبر وسيلة ممتازة لتحسين المهارات البدنية والتحمل والصحة العامة وكذلك الاندماج الاجتماعي. وتُنظم العديد من المنافسات على المستوى المحلي والإقليمي والعالمي لتحقيق الإثارة والتحديات الجديدة.

            \end{RLtext}}\\
    
        Language &  \textbf{Russian} \\
        LLM Generator & \textsc{Davinci-003} \\
        Domain &  \textit{RuATD} \\
        \midrule
        Human & \foreignlanguage{russian}{День был посвящен празднику на «линии движения» его сына. Гнетущее осталось впечатление от некоторых (от трех лиц) явно и (одно из них) даже оформленно приближенного к его жизни с особыми правами на близость. Угашаю вспыхнувшую потребность судить и осуждать одну женщину и одну мужскую фигуру, их лица, глаза, выражения, манеры смотреть, говорить, двигаться. Осталось больным местом только искаженное их присутствием и тоном отношение к ним главы дома. Жена и мать ее, и место, какое они заняли в обновлении внешнего и внутреннего лика Игоря, хочется верить, не смогут в главных чертах лика его внутренней жизни внести изменение в сторону разрушительную или застойную. Внешне облегчат и приукрасят жизнь. Одни ямочки в улыбке жены на ее девически-свежем лице, при твердом до жесткости взгляде ясных серых глаз, делают понятной ее привлекательность для мягкосердечного, но в какой-то части своего существа угрюмого мужа. Вспомнилось, как 4 года тому назад после знакомства его с одной красивой, в возрасте между 30-40 годами, женщиной, которая картиной душевного мира и взглядом на жизнь, по-моему, должна была бы разбудить интерес к себе у Игоря, он согласился со мной, что «она, по-видимому, хороший человек». Но тут же прибавил: — Разве вы не видите, тетя Вава, что мне лучше всего одному, как я теперь живу. Дико и оскорбительно представить себе вот в этих стенах какую-то другую жену, кроме той, которая у меня отнята смертью.}\\ 
        \midrule
        Prompt & \foreignlanguage{russian}{Пожалуйста, переформулируйте этот текст, чтобы он содержал более 1000 символов. Текст: 'День был посвящен празднику на «линии движения» его сына. Гнетущее осталось впечатление от некоторых (от трех лиц) явно и (одно из них) даже оформленно приближенного к его жизни с особыми правами на близость. Угашаю вспыхнувшую потребность судить и осуждать одну женщину и одну мужскую фигуру, их лица, глаза, выражения, манеры смотреть, говорить, двигаться. Осталось больным местом только искаженное их присутствием и тоном отношение к ним главы дома. Жена и мать ее, и место, какое они заняли в обновлении внешнего и внутреннего лика Игоря, хочется верить, не смогут в главных чертах лика его внутренней жизни внести изменение в сторону разрушительную или застойную. Внешне облегчат и приукрасят жизнь. Одни ямочки в улыбке жены на ее девически-свежем лице, при твердом до жесткости взгляде ясных серых глаз, делают понятной ее привлекательность для мягкосердечного, но в какой-то части своего существа угрюмого мужа. Вспомнилось, как 4 года тому назад после знакомства его с одной красивой, в возрасте между 30-40 годами, женщиной, которая картиной душевного мира и взглядом на жизнь, по-моему, должна была бы разбудить интерес к себе у Игоря, он согласился со мной, что «она, по-видимому, хороший человек». Но тут же прибавил: — Разве вы не видите, тетя Вава, что мне лучше всего одному, как я теперь живу. Дико и оскорбительно представить себе вот в этих стенах какую-то другую жену, кроме той, которая у меня отнята смертью.'}\\
        \midrule
        Machine & \foreignlanguage{russian}{День праздника был посвящен пути движения Игоря, его сына. Взгляд, которым его судьба была окрашена, был властный, а к мимолетным правам на близость был дан официальный статус. Желание судить и осуждать пару, видимо, комплексное, так как сразу трое - мужчина, женщина и в какой-то мере дети, занимаются браконьерством. Даже брутальное отношение главы дома к ним не могло исказить более-менее нормальное настроение.Одержимость жены, ее свежесть, девичья улыбка с ямочками на лице, а также твердый взгляд серых глаз очень привлекла мягкосердечного, но застаревшего мужа. Назад вспомнили, как 4 года назад он встретился с красавицей, которая должна была привлечь его интерес, по мнению Игоря, в душевном мире и представлении жизни. Однако, овладение собой было столь сильно, что он дошел до того, что «лучше для него жить в одиночестве, чем представлять себе другую жену внутри этих чертовых стен». Даже сейчас после потери жены, чья память приукрасит и облегчит жизнь Игоря и его сына, до разрушительных, можно догадываться, или загнутых ходов, они не доводятся.}\\

        \midrule 
        \midrule 
        Language &  \textbf{Russian} \\
        LLM Generator & \textsc{ChatGPT} \\
        Domain &  \textit{RuATD} \\
        \midrule
        Human & \foreignlanguage{russian}{Регулирование градостроительной деятельности на территории муниципального образования село Ныда (далее – муниципальное образование) должно осуществляться в соответствии с основными принципами законодательства о градостроительной деятельности, направленными на устойчивое развитие территории, создание условий для привлечения инвестиций и активизации строительства, формирование экологически безопасной, благоприятной среды жизнедеятельности, комплексное и эффективное развитие социальной, производственной, инженерной и транспортной инфраструктуры, бережное природопользование, повышение уровня архитектурно-художественной выразительности застройки муниципального образования, на основании утвержденных документов территориального планирования и градостроительного зонирования, а также документации по планировке территории.}\\ 
        \midrule
        Prompt & \foreignlanguage{russian}{Пожалуйста, переформулируйте этот текст, чтобы он содержал более 1000 символов. Текст:'Регулирование градостроительной деятельности на территории муниципального образования село Ныда (далее – муниципальное образование) должно осуществляться в соответствии с основными принципами законодательства о градостроительной деятельности, направленными на устойчивое развитие территории, создание условий для привлечения инвестиций и активизации строительства, формирование экологически безопасной, благоприятной среды жизнедеятельности, комплексное и эффективное развитие социальной, производственной, инженерной и транспортной инфраструктуры, бережное природопользование, повышение уровня архитектурно-художественной выразительности застройки муниципального образования, на основании утвержденных документов территориального планирования и градостроительного зонирования, а также документации по планировке территории.'}\\
        \midrule
        Machine & \foreignlanguage{russian}{Муниципальное образование село Ныда должно регулировать свою градостроительную деятельность в соответствии с законодательством о градостроительстве. Эти законы направлены на устойчивое развитие территории и создание обстановки, способствующей инвестициям и активизации строительства. Они также ставят целью сформировать экологически безопасную, благоприятную среду жизнедеятельности и развивать всю инфраструктуру, как социальную, так и производственную, инженерную и транспортную. Помимо этого, они должны бережно использовать ресурсы природы и повышать уровень архитектурной экспрессии в чертежах своих построек.
        
        Муниципальное образование должно использовать утвержденные документы по территориальному планированию и градостроительному зонированию, а также документацию по планировке территории. Эти правила регулируют различные аспекты градостроительства, такие как масштаб и высота зданий, открытые пространства, пешеходные и транспортные маршруты, зеленые насаждения и многое другое.
        
        Местные власти должны заботиться о том, чтобы градостроительство было эффективным и комплексным, учитывая различные потребности граждан в доступном жилье и инфраструктуре. Они также должны уделять большое внимание экологической устойчивости и сохранению биоразнообразия.
        
        В целом, здесь требуется сбалансированное подход к градостроительству в муниципальном образовании, искреннее взаимодействие с обществом и эффективное использование ресурсов, чтобы создать благоприятную и устойчивую среду для жизни.}\\

        Language &  \textbf{Bulgarian} \\
        LLM Generator & \textsc{Davinci-003} \\
        Domain &  \textit{Bulgarian True \& Fake News} \\
        \midrule
        Title & \foreignlanguage{russian}{Скандал! РДНСК прати багер да маха буните на най-скъпия плаж в България} \\
        \midrule
        Human & \foreignlanguage{russian}{28 Март 2017, Вторник Автор: Иван КОЛЕВ, снимки: Георги РУСЧЕВИвицата е тясна, ако ги няма буните плажът ще изчезне, алармира стопанинът на плажа - „Акра лидо”Багери, пратени от РДНСК Бургас, тази сутрин „захапаха” двете буни на къмпинг „Черноморец”. От строителния надзор твърдят, че съоръженията са незаконни и трябвало да бъдат премахнати. Машините са спрени с жива верига от 15-ина души и в момента се очаква полицията.„Няма да допуснем да продължат. Ако се премахнат буните за един месец ще изчезне целият плаж”, каза един от протестиращите пред Флагман.бг.Наш репортер се свърза и със стопанина на плажа – бургаската фирма „Акра лидо” ООД, която има 5-годишен договор за ивицата и плаща най-скъпия плажен наем в България – по 107 хил.лева на година за изключително малка площ.От компанията останаха изненадани от действията на строителния наздор. Буните са там от много години, спират свлачищния процес и осигурят плажна ивица, макар и много тясна.Само преди седмица от „Акра лидо” са платили накуп наемната цена за 2017 година, а сега ако бъдат премахнати буните ще останат без плаж.Скандалът ще има международен отзвук, защото 500 руски семейства имат жилища в непосредствена близост до къмпинг „Черноморец” и са ги закупили заради това, че плажът им е само на няколко метра.Това е друга от машините, която ще участва в акцията по разрушаването на буните„Ние сме изрядни платци, наели сме плажа в това състояние от държавата и до тази сутрин смятахме, че всичко е изрядно. Не сме поставяли под съмнение, че държавата ни е отдала под наем плаж с незаконни буни”, каза един от собствениците на фирмата Милен Янков пред Флагман.бг.Припомняме, че миналата година също имаше много проблеми с този плаж. В основата бяха отношенията със стария стопанин на ивицата (ВИЖ НАШАТА ТЕМА В РАЗВИТИЕ)ОЧАКВАЙТЕ ПОДРОБНОСТИ! В категории: Созопол}\\ 
        \midrule
        Prompt & \foreignlanguage{russian}{Моля, генерирайте статия с дължина над 1000 символа и заглавие "Скандал! РДНСК прати багер да маха буните на най-скъпия плаж в България"} \\
        \midrule
        Machine & \foreignlanguage{russian}{Скандал! РДНСК прати багер да маха буните на най-скъпия плаж в България
        
        В този момент, най-високото бюро в България за разрешаване на спорове между плавателни клубове и партньорите им предприе нещо неочаквано. Речникът Държавен надзор на стандартите (РДНСК) изпрати багер да махне буните на най-скъпия плаж в България - Golden Sands.
        
        Бунита е в Балтийското море, в близост до Варна. По неизвестни причини РДНСК реши да започне да маха буни, а при това привлече много необичайно интересни детайли.
        
        Ако се сещате по-добре, то освен това, че РДНСК разреши махането на буните на Golden Sands, той поръча да се изтрият буните от другите плажове в България в същото време. Това е правило, което РДНСК не следва преди това и което създаде много недоволство в индустрията.
        
        Докато правните изследвания са все още в процес на продължаване за да се разбере точно какви са причините за багера, се очаква тази политика да продължи и да се извършват постоянни разрешения за някои интереси. Някои пленници на ансамбъла на бунита подози, че максимумът на разрешенията е 100, които могат да се дават годишно според правилата, които сме предвидили.
        
        Защо РДНСК решил да направи това и защо именно сега, това продължава да бъде неизвестно. Но понеже това се оказва наистина забранено, много от плавателните клубове и индивиди на плажа ще бъдат принудени да се преместят на други места.
        
        Бановете, разположени в гледната точка за потоците, да се зачитат на рилско море - това ли ще стане след махането на буните? Ще има ли някаква защита за плавателните инсталации, една от важните атракции на Golden Sands? Ще плажът продължи да съществува или ще бъде претърпял значителни изменения? Тези и много други въпроси очакват отговори.
        
        За да се оцени тайнственото поведение на РДНСК, много информация трябва да се прецени. Ето защо е особено важно да се осигури правопорядъчно намиране на логичен разследвателен процес за да се разберат всички мотиви за действието на РДНСК и да се донесат промени, които да се правят в добро на всички страни.}\\

        Language &  \textbf{Bulgarian} \\
        LLM Generator & \textsc{ChatGPT} \\
        Domain &  \textit{Bulgarian True \& Fake News} \\
        \midrule
        Title & \foreignlanguage{russian}{Неподозирани разкрития за кралското семейство} \\
        \midrule
        Human & \foreignlanguage{russian}{Обикновените хора се обръщат към кралица Елизабет Втора с „Нейно кралско величество“ и придружават обръщението с поклон и израз на любезност. Но зад кулисите отношенията са доста по-неформални, разказват от „Новини Лондон“. Потребител на сайта Quora зададе въпрос: „Как внуците на кралица Елизабет се обръщат към нея на четири очи?“. Макар че отговорите варират, проучване на „Индипендънт“ показва, че внуците на кралицата са й измислили обръщения, изразяващи обичта им към нея. Уилям и Хари окуражаваха баба си при нейната неочаквана, спираща дъха роля на церемонията по откриването на Олимпийските игри през 2012 г. При вида на баба им, която се появи заедно с Даниел Крейг като Джеймс Бонд, принцовете извикаха с цяло гърло: „Давай, бабо!“. Но това не е единственото име, използвано от Уилям и Хари. Според светския хроникьор Ричард Кей веднъж в Бъкингамския дворец малкият Уилям казал на баба си „Гари“. Когато служители на двореца попитали „Кой е Гари?“, кралицата отговорила: „Аз съм Гари“. Уил още не се е научил да казва „Granny” (бабо – английски език, бел. ред.), обяснила Елизабет. Тригодишният правнук на Елизабет Втора принц Джордж също има любвеобилно обръщение към кралицата, както стана ясно от интервю, излъчено по телевизия ITV миналото лято по повод 90-годишнината й. Херцогинята на Кеймбридж Кейт Мидълтън сподели, че Джордж нарича баба си „Ган Ган“. Но това не е просто мило обръщение, а е израз на дълготрайна кралска традиция. Американският биограф Кити Кели твърди, че принц Чарлз, Уилям и Хари също се обръщали така към своята прабаба.}\\ 
        \midrule
        Prompt & \foreignlanguage{russian}{Моля, генерирайте статия с дължина над 1000 символа и заглавие "Неподозирани разкрития за кралското семейство"} \\
        \midrule
        Machine & \foreignlanguage{russian}{Неподозирани разкрития за кралското семейство
        
        Кралските семейства винаги са били обект на внимание и интерес от страна на медиите и обществото. Известни са скандалите и скритите истории, които остават извън зрителния ъгъл и несъмнено експлоатирането на подобни теми продължава да държи високи места в таблоидите.
        
        Но дали знаете всичко за кралските семейства, което медиите не съобщават? Има ли неща, които остават скрити дълбоко в историята или в семейните тайни?
        
        Изглежда, отговорът е да. В последните години имаше повече и повече изисквания за откриване на тайните на кралските династии, но много от тези разкрития остават скрити за публика.
        
        Една от най-известните скрити истории, свързани с британското кралско семейство, е тази на принца Джон, третият син на крал Джордж V и кралица Мери. Принц Джон беше диагностициран с епилепсия и склероза на 7-годишна възраст. Той беше изолиран от обществото и от своето семейство и беше принуден да живее в бездейственост и самота, за да не покаже "ненормалното" си поведение. Принц Джон почина на 13-годишна възраст, покрит с тайните и позволено му беше да забрави света на живите.
        
        Разказите за скритата история на Американското кралско семейство са също известни. Съществува теория, че Сежър, който се хвали, че е третият наследник на трона след Алфонсо и Джон, наистина има връзки с кралското семейство на САЩ. През 1937 г. Сежър заяви, че е бил роден като дете на Сесилия, дъщерята на крал Едуард VIII и Уолис Симпсън.
        
        Макар това да е слух, наличието на тайни и тъмни истории в кралските семейства е добре документирано. Все още има много да се разкрие, но кралските династии на света ще продължат да привличат интриги и вълнение през годините.}\\

        Language &  \textbf{Indonesian} \\
        LLM Generator & \textsc{ChatGPT} \\
        Domain &  \textit{News} \\
        \midrule
        Title & Ahmad Dhani Klaim Tak Punya Informasi soal Sri Bintang \\
        \midrule
        Human & Jakarta, CNN Indonesia -- Musisi Ahmad Dhani memenuhi panggilan Polda Metro Jaya untuk diperiksa sebagai saksi tersangka dugaan makar Sri Bintang Pamungkas, Selasa (20/12). Berdasarkan pantauan CNNIndonesia.com, dia tiba pukul 15.00 WIB. Sedangkan tim kuasa hukumnya yang tergabung dalam Advokat Cinta Tanah Air sudah tiba satu jam sebelum kedatangannya. Tak lama kemudian, Farhat Abbas juga datang untuk mendampingi Ahmad Dhani. Ahmad Dhani mengatakan, dirinya tidak akan memberikan informasi apapun soal Sri Bintang. Dia mengklaim tidak kenal dengan Sri Bintang. Buni Yani dan Ahmad Dhani Jadi Saksi Kasus Sri Bintang Besok Saksi Mengaku Dapat Aliran Dana dari Tersangka Makar Buni Yani Diperiksa Soal Pidato Sri Bintang di Kalijodo "Informasi pasti tidak ada, karena saya tidak kenal dengan Sri Bintang Pamungkas. Saya pernah ketemu beliau ketika di Mako Brimob," ucapnya. Meski demikian, Ahmad Dhani mengaku hadir saat ada pertemuan di Universitas Bung Karno. Namun, ia mengklaim tidak mendengar pidato yang disampaikan oleh Sri Bintang karena terlambat datang. Awalnya, Kepala Subdirektorat Kejahatan dan Kekerasan Direktorat Reserse Kriminal Umum Polda Metro Jaya AKBP Hendy Kurniawan mengatakan, karena Ahmad Dhani sedang sakit maka pemeriksaan terhadap dia yang akan dilakukan hari ini harus ditunda Kamis (22/12). "Ahmad Dhani ditunda karena sakit. Surat sakit sudah kami terima. Kami sudah koordinasi dengan kuasa hukumnya kemarin," ujarnya. Kuasa Hukum Ahmad Dhani, Ali Lubis mengklaim, kliennya tersebut merupakan warga negara yang baik sehingga datang memenuhi panggilan kepolisian. "Beliau kooperatif, beliau ingin membantu kepolisian jadi lebih cepat lebih baik," ucapnya. Ahmad Dhani telah ditetapkan sebagai tersangka dugaan penghinaan terhadap penguasa. Dia ikut ditangkap pada Jumat (2/12) dengan sejumlah tersangka dugaan makar lainnya. Sepuluh tersangka dugaan makar itu adalah Sri Bintang, Kivlan Zein, Adityawarman Thahar, Ratna Sarumpaet, Firza Huzein, Eko Santjojo, Alvin Indra, Rachmawati Soekarnoputri, dan kakak beradik Rizal Kobar dan Jamran.\\ 
        \midrule
        Prompt & Tulislah suatu artikel berita yang berjudul "Ahmad Dhani Klaim Tak Punya Informasi soal Sri Bintang". Artikelnya harus mengandung paling tidak 300 kata dalam bahasa Indonesia.     Tambahkan "Jakarta, CNN Indonesia -- " di awal artikel. \\
        \midrule
        Machine & Jakarta, CNN Indonesia -- Ahmad Dhani, personel grup band Dewa 19, mengaku tak memiliki informasi mengenai sosok Sri Bintang yang tengah ramai dibicarakan di media sosial. Seperti diketahui, Sri Bintang menjadi perbincangan publik setelah dirinya mengunggah video yang menyebutkan bahwa personel band Dewa 19, Ahmad Dhani, pernah menyebut Sri Bintang sebagai sosok yang berperan dalam pembuatannya dari awal. Klaim tersebut kemudian dibantah oleh Ahmad Dhani melalui sebuah video yang diunggahnya di akun media sosialnya. "Saya sendiri nggak tahu siapa Sri Bintang itu. Jadi saya gak bisa bicara, saya gak bisa komen, kenapa orang lain mau ngomongin saya tentang Sri Bintang," ujar Ahmad Dhani dalam video tersebut. Ahmad Dhani juga mengatakan bahwa beberapa orang yang mengklaim tahu tentang perjalanannya dalam industri musik sebenarnya tidak memiliki informasi yang tepat. "Saya sekali lagi ingin tegaskan bahwa ada beberapa orang yang bicara itu nggak tahu apa yang mereka bicarakan. Intinya, kita perlu memperhatikan orang yang kita ngomongin," tegasnya. Kendati begitu, Ahmad Dhani mengungkapkan bahwa dirinya berharap Sri Bintang bisa berkarya dan memperlihatkan kemampuan yang dimilikinya di industri musik tanah air. Ahmad Dhani juga berharap bahwa sosok Sri Bintang tidak menjadi obyek perdebatan yang tak berkesudahan di antara para penggemar musik dan publik di Indonesia. "Setidaknya semua artis harus diberikan kesempatan yang sama untuk berkarya. Saya berharap Sri Bintang bisa berkarya dengan baik dan melebihi ekspektasi dari semua orang," harapnya. Meski begitu, klaim dari Sri Bintang mengenai Ahmad Dhani telah menjadi perhatian banyak pengguna media sosial. Banyak netizen yang menyatakan bahwa hal tersebut merupakan sebuah bukti bahwa industri musik di Indonesia selalu menarik perhatian masyarakat dan penuh dengan kontroversi. Maka dari itu, publik diharapkan mampu menghormati karya dari masing-masing musisi dan menghindari terjadinya perdebatan yang tak perlu di media sosial.\\
       
        Language &  \textbf{Chinese} \\
        LLM Generator & \textsc{ChatGPT} \\
        Domain &  \textit{Baike QA} \\
        \midrule
        Question & \begin{CJK}{UTF8}{gbsn}关于百宝箱的问题我本人这里网络连接经常会不时中断,请问一下我的网\end{CJK} \\
        \midrule
        Human & \begin{CJK}{UTF8}{gbsn}我也是传3玩家，你的问题我经常遇到，其实这种问题是很难解决的，问题是GT的总服务器设定为无反映30秒下线，我们挂 机如果身边有怪突然掉线30秒的时间不喝红战士还有可能没问题，法师必挂，解决的话只能换换网络或者带普通装备挂机~！\end{CJK}\\ 
        \midrule
        Prompt & \begin{CJK}{UTF8}{gbsn}关于百宝箱的问题我本人这里网络连接经常会不时中断,请问一下我的网,详细来说,我本人这里连接经常会不时中断,请问一下我的网络连接断了以后人物还会保留在游戏中间一段时间吗?有的话是几秒?顺便还后突然重起机子的情况, 我有时在打怪的时候网络连接中断,再上的时,有时候是满血,好象没被怪打过,但是有几次就发现回城了,还掉了装备,请懂的详细告诉我一下,能解决突然网络中断人物立刻就下线的问题吗?(我用的是百宝挂机)\end{CJK}\\
        \midrule
        Machine & \begin{CJK}{UTF8}{gbsn}如果您的网络连接中断，您的角色可能会保留在游戏中的一段时间，具体时间取决于游戏服务器和您的网络连接速度。一般来说，如果您在游戏中掉线，您的角色会在几秒钟之内被服务器标记为离线状态，如果您的网络恢复连接，您可以尝试重新登录游戏。但如果您的网络连接中断时间太长，可能会导致您的角色被游戏服务器强制下线。\end{CJK}
        \begin{CJK}{UTF8}{gbsn}关于网络连接不稳定导致角色死亡或掉装备的问题，这是由于在网络中断的情况下，游戏服务器无法正确接收您的移动或攻击操作，因此可能会导致您的角色无法正常反应或被怪物攻击致死。而掉装备的情况可能是由于您的人物在网络中断时正好被怪物攻击，并且该物品尚未在游戏服务器上保存，因此游戏服务器无法将该物品恢复至您的背包中。
        解决突然网络中断导致角色掉线的最佳方法是改善您的网络连接，例如通过使用更稳定的网络连接方式、更快的网络速度、更好的路由器等来提高网络信号的稳定性。另外，您也可以考虑使用其他的游戏挂机工具来降低游戏对网络连接的依赖。\end{CJK}\\		
	\end{supertabular}
\end{center}